\documentclass[runningheads]{llncs}
\usepackage{graphicx}
\usepackage{amsmath,amssymb} % define this before the line numbering.
\usepackage{color}

\usepackage{ownstyles}

\newif\ifcomments%

\commentsfalse
\ifcomments%
\newcommand{\comments}[1]{#1}
\else
\newcommand{\comments}[1]{}
\fi
\usepackage{xcolor}
\usepackage{subcaption} % for subfigures
\usepackage{hyperref}

\usepackage{amsmath,amsfonts,bm}

\def\eqref#1{equation~\ref{#1}}
\def\1{\bm{1}}

\def\vc{{\bm{c}}}

\def\vx{{\bm{x}}}
\def\vy{{\bm{y}}}
\def\vz{{\bm{z}}}

\DeclareMathAlphabet{\mathsfit}{\encodingdefault}{\sfdefault}{m}{sl}
\SetMathAlphabet{\mathsfit}{bold}{\encodingdefault}{\sfdefault}{bx}{n}

\def\gN{{\mathcal{N}}}

\def\gZ{{\mathcal{Z}}}

\def\sR{{\mathbb{R}}}

\newcommand{\E}{\mathbb{E}}

\newcommand{\R}{\mathbb{R}}

\newcommand{\eg}{e.g.\xspace}
\newcommand{\ie}{i.e.\xspace}
\newcommand{\class}[1]{\ensuremath{\mathsf{#1}\xspace}}
\newcommand{\layer}[1]{\ensuremath{\mathsf{#1}\xspace}}

\definecolor{ao}{rgb}{0.0, 0.5, 0.0}
\newcommand{\qi}{\comments{\textcolor{red}{Qi: }\textcolor{red}}}
\newcommand{\subsec}[1]{\noindent{\textbf{#1~~}}}

\begin{document}
\pagestyle{headings}
\mainmatter

% \def\ACCV20SubNumber{717}  % Insert your submission number here

%===========================================================
% \title{Improving sample diversity of a pre-trained, class-conditional GAN by changing its class embeddings
% } % Replace with your title

\title{A cost-effective method for improving and re-purposing large, pre-trained GANs by fine-tuning their class-embeddings}
% \title{\qi{Improving sample diversity of a pre-trained, class-conditional GAN by finding multiple class embeddings}}
\titlerunning{A cost-effective method for improving and re-purposing BigGANs}

\author{Qi Li\inst{1} \and
Long Mai\inst{2} \and
Michael A. Alcorn \index{Alcorn, Michael} \inst{1} \and
Anh Nguyen\inst{1} 
% Anh Nguyen\inst{1}\orcidID{2222--3333-4444-5555. if there's orcidID}
}

\authorrunning{Q. Li et al.}
\institute{Auburn University, Auburn, AL 36849, USA \\
\email{\{qzl0019, alcorma\}@auburn.edu, anh.ng8@gmail.com} \and
Adobe Research, San Jose, CA 95110, USA \email{malong@adobe.com}
% \institute{Department of Computer Science and Software Engineering, Auburn University, Auburn AL 36849, USA \\
% \email{\{qzl0019, alcorma\}@auburn.com, anh.ng8@gmail.com} \and
% Adobe Inc., San Jose CA 95110, USA \email{malong@adobe.com}
%\\
% \url{http://www.springer.com/gp/computer-science/lncs}
}
\maketitle

%===========================================================
\begin{abstract}
Large, pre-trained generative models have been increasingly popular and useful to both the research and wider communities.
Specifically, BigGANs \cite{brock2019large}---a class-conditional Generative Adversarial Networks trained on ImageNet---achieved excellent, state-of-the-art capability in generating realistic photos.
However, fine-tuning or training BigGANs from scratch is \emph{practically impossible} for most researchers and engineers because (1) GAN training is often unstable and suffering from mode-collapse \cite{arjovsky2017towards,brock2019large}; and (2) the training requires a significant amount of computation, 256 Google TPUs for 2 days or 8 $\times$ V100 GPUs for 15 days.
Importantly, many pre-trained generative models both in NLP and image domains were found to contain biases that are harmful to the society \cite{AIWeekly70:online,sheng2019woman}.
Thus, we need computationally-feasible methods for modifying and re-purposing these huge, pre-trained models for downstream tasks.
In this paper, we propose a cost-effective optimization method for improving and re-purposing BigGANs by fine-tuning only the class-embedding layer.
We show the effectiveness of our model-editing approach in three tasks: (1) significantly improving the realism and diversity of samples of complete mode-collapse classes; (2) re-purposing ImageNet BigGANs for generating images for Places365; and (3) de-biasing or improving the sample diversity for selected ImageNet classes.

% From GPT-2 \cite{radford2019language}, RoBERTa \cite{liu2020roberta} to BigGAN \cite{brock2019large}, large pre-trained machine learning (ML) models have been increasingly popular and useful to the wider community.

% Mode collapse is a well-known issue with Generative Adversarial Networks (GANs) posing a big challenge to the research community.
% We propose a simple solution to mode collapse \ie \qi{improving the sample diversity of a pre-trained class-conditional GAN by finding multiple class embeddings for each class.}
% improving the sample diversity of a pre-trained class-conditional GAN by modifying only its class embeddings.

% To keep the samples in correct classes while the embeddings change in the direction of maximizing sample diversity, we also move the embeddings in the direction of maximizing the log probability outputs of an auxiliary classifier pre-trained on the same dataset.

% \qi{To increase the samples diversity for one class while the embeddings change in the direction of maximizing sample diversity, we also move the embeddings in the direction of maximizing the log probability outputs of an auxiliary classifier pre-trained on the same dataset.}

% We improved the sample diversity of state-of-the-art ImageNet BigGANs at both $128\times128$ and $256\times256$ resolutions.
% By replacing the embeddings, we can also synthesize plausible images for Places365 using a BigGAN pre-trained on ImageNet, revealing---for the first time---the surprising expressivity of the BigGAN class embedding space.
\end{abstract}

%===========================================================
\section{Introduction}

% 1. Large, Pre-trained models are useful
% 2. But they are hard to trained / fine-tuned and may have biases issues
% 3. For example, in the image domain, BigGANs have taken the field by storm. 
% But they have complete mode-collapse and ARE takes 35 days to train
% 4. How to train/edit BigGANs?
% 5. In this paper, we propose AM method.

% From GPT-2 \cite{radford2019language}, RoBERTa \cite{liu2020roberta} to BigGAN \cite{brock2019large}, large pre-trained machine learning (ML) models have been increasingly popular and useful to the wider community.

From GPT-2 \cite{radford2019language} to BigGAN \cite{brock2019large}, large, pre-trained generative models have been increasingly popular and useful to both the research and wider communities.
Interestingly, these pre-trained models have remarkably high utility but near-zero re-trainability.
That is, GPT-2 or BigGANs were all trained on extremely large-scale computational infrastructure, which is not available to the rest of the community.
In practice, training or fine-tuning such models is impossible to most researchers and engineers.
Importantly, pre-trained generative models in both text and image domains were found to capture undesired, hidden biases that may be harmful to the society \cite{AIWeekly70:online,sheng2019woman}.
Therefore, the community needs techniques for fine-tuning and re-purposing pre-trained generative models.

%%%%%%% BigGAN and its problems

\begin{figure*}
	\centering
	{	
		\begin{flushleft}
			\hspace{0.9cm} (A) ImageNet
			\hspace{1.8cm} (B) BigGAN \cite{brock2019large}
			\hspace{1.8cm} (C) AM (ours)
		\end{flushleft}
	}
	\vspace{-0.3cm}
	\begin{subfigure}[b]{1.0\linewidth}
		\centering
		\includegraphics[width=1.0\linewidth]{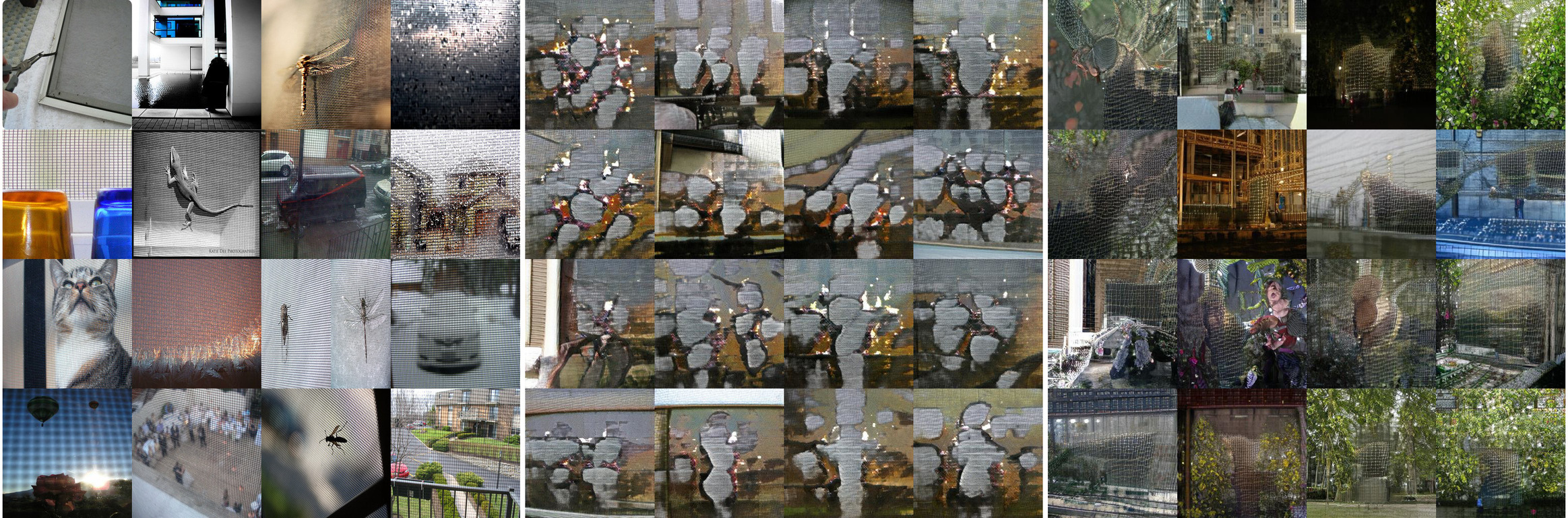}
		\caption{Samples from the \class{window~screen} class (904).}
	\end{subfigure}
	\begin{subfigure}[b]{1.0\linewidth}
		\centering
		\includegraphics[width=1.0\linewidth]{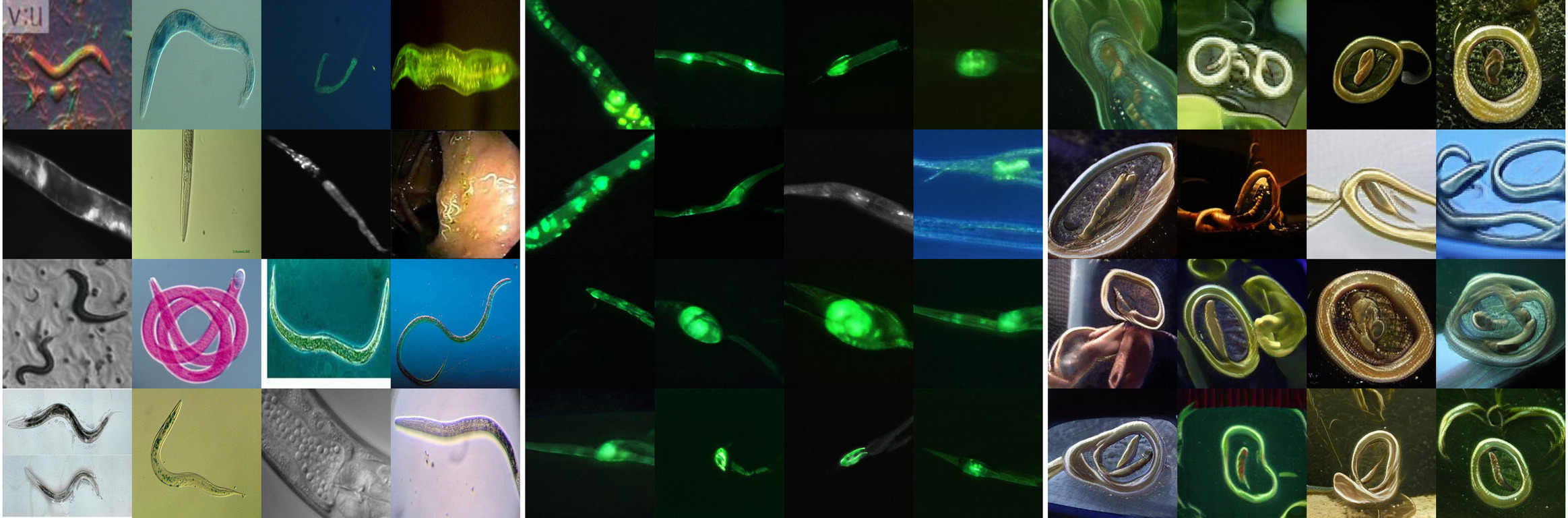}
		\caption{Samples from the \class{nematode} class (111).}
	\end{subfigure}
% 	\begin{subfigure}[b]{1.0\linewidth}
% 		\centering
% 		\includegraphics[width=1.0\linewidth]{images/256_imagenet_biggan_biggan-am_815_final_show.jpg}
% 		\caption{Samples from the \class{spider~web} class (815).}
% 	\end{subfigure}
	
	\caption{For some classes, $256\times256$ BigGAN samples (B) have poor realism and diversity (\ie samples are biased towards one type of data) while the real ImageNet images (A) are diverse.
% 	A comparison between the $256\times256$ samples from the ImageNet training set (A), the original BigGAN model (B), and our AM method (C).
	AM samples (C) are of higher diversity than the original BigGAN samples (B). 
	}
	\label{fig:repair_mode_collapse}
\end{figure*}

The class-conditional BigGAN \cite{brock2019large} has reached an unprecedented state-of-the-art image quality and diversity on ImageNet by using large networks and batch sizes.
However, fine-tuning or training BigGANs from scratch is impractical for most researchers and engineers due to two main reasons.
First, Generative Adversarial Networks (GANs) training is notoriously unstable and subject to mode-collapse \cite{arjovsky2017towards,brock2019large}
\ie the generated distribution does not capture all modes of the true distribution \cite{arjovsky2017towards}.
Consistent with \cite{ravuri2019seeing}, we observed that BigGAN samples from a set of $\sim$50 classes exhibit substantially lower diversity than samples from other classes do.
For example, BigGAN samples from the \class{window~ screen} class are rubbish examples \ie noisy patterns that are not recognizable to humans (Fig.~\ref{fig:repair_mode_collapse}a).
Similarly, \class{nematode} samples are heavily biased towards green worms on black, but the training data includes worms of a variety of colors and backgrounds (Fig.~\ref{fig:repair_mode_collapse}b).
% Similar mode-collapse classes were found in other BigGAN models \cite{ravuri2019seeing}.

Second, re-training BigGANs requires significantly expensive computation---the original $256\times 256$ model took 48 hours of training on 256 Google Cloud TPUs.
On more modest hardware of 8 $\times$ V100 GPUs, the training is estimated to take more than 2 weeks \cite{brock2019pytorch} but has not been found to match the published results in \cite{brock2019large}.
Importantly, re-training or finetuning BigGANs were found to still cause a set of classes to collapse as observed in a BigGAN-deep model \cite{ravuri2019seeing} (in addition to BigGAN models) released by \cite{brock2019large}.

% For example, generated images for \class{daisy} mostly show white flowers on green grass, but the training data includes images of daisies with a variety of colors and backgrounds (Fig.~\ref{fig:teaser}).
% Furthermore, samples for \class{window~screen} not only have low diversity but also poor realism (see Fig.~\ref{fig:imagenet_vs_bad_biggan} for more low-diversity BigGAN samples).

%%%%%%% Our solution

\begin{figure*}[h]
	\centering
	{	
		\begin{flushleft}
			\hspace{0.5cm} (A) BigGAN  \cite{brock2019large}
			\hspace{0.9cm} (B) Modifying class embeddings
			\hspace{0.8cm} (C) AM (ours)
		\end{flushleft}
	}
	\vspace{-0.3cm}
	\includegraphics[width=1.0\linewidth]{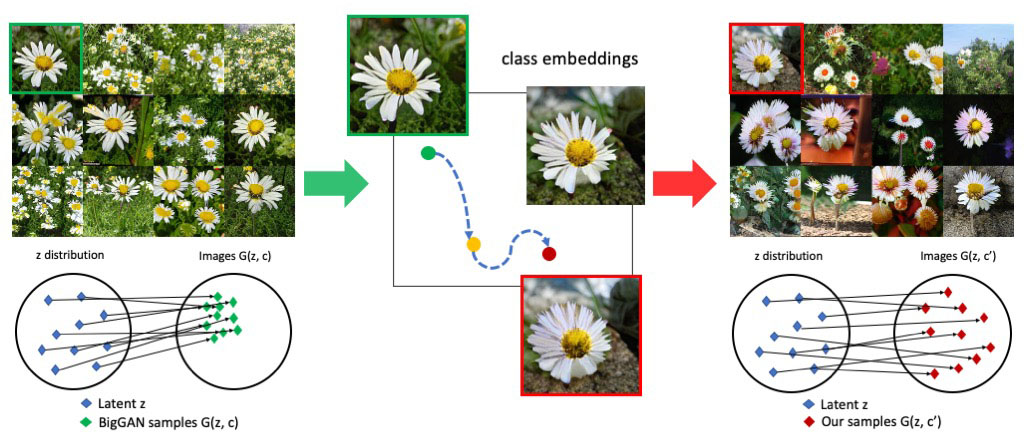}

	\caption{
		With BigGAN embeddings (A), the latent $\vz$ vectors are mapped to nearby points (green \textcolor{green}{$\blacklozenge$}) \ie similarly-looking images. 
		Our embedding optimization moves the original embedding to a new vector where the generated samples (red \textcolor{red}{$\blacklozenge$}) are more diverse. 
        Here, the updated class embedding $\vc$ changes the background of a daisy from green grass (\textcolor{green}{$\Box$}) to brown soil (\textcolor{red}{$\Box$}).
        Note that the pose of the flower (controlled by $\vz$) remain the same. Effectively, with only a change in the embedding, the latent vectors are re-mapped to more spread-out points or more diverse set of samples (C).
		}
	\label{fig:opt_process}
\end{figure*}

% Is there a simple solution to improving sample diversity of BigGANs or re-purposing it for generating images of unseen classes?
In this paper, we propose a cost-effective method for improving sample diversity of BigGANs and re-purposing it for generating images of unseen classes.
Leveraging the intuition that the BigGAN generator is already able to synthesize photo-realistic images for many ImageNet classes \cite{brock2019large}, we propose to modify \emph{only the class embeddings} while keeping the generator unchanged (Fig.~\ref{fig:opt_process}).
We demonstrate our simple yet effective approach on three different use cases:\footnote{Code for reproducibility is available at \url{https://github.com/qilimk/biggan-am}.}

\begin{enumerate}
% 	\item The BigGAN class embeddings qualitatively capture class semantics (Sec.~\ref{sec:low_diversity}). 
% 	For example, bird classes are nearby in t-SNE visualizations (Fig.~\ref{fig:biggan_tsne_zoomin}).
    \item Changing only the embeddings is surprisingly sufficient to ``recover'' diverse and plausible samples for complete mode-collapse classes \eg \class{window~screen} (Fig.~\ref{fig:repair_mode_collapse}a). 
    
    \item We can re-purpose a BigGAN, pre-trained on ImageNet, for generating images matching unseen Places365 classes (Sec.~\ref{sec:places}).
    
% 	\item On ImageNet, our method reduces the sample diversity gap (via MS-SSIM and LPIPS metrics) between the real and generated data by $\sim50\%$ (Sec.~\ref{sec:AM}).
% 	A human study found that our method produced more diverse and similarly realistic images compared to BigGAN (Sec.~\ref{sec:human}).
% 	\item Our approach improved the sample diversity for two BigGANs released by the authors---at $256\times256$ and $128\times128$ resolutions (Sec.~\ref{sec:128x128})---and some mode-collapse BigGAN snapshots at $128\times128$ (Sec.~\ref{sec:snapshots}).
    \item On ImageNet, our method improves the sample diversity by $\sim50\%$ for the pre-trained BigGANs released by the authors---at $256\times256$ and $128\times128$ resolutions by finding multiple class embeddings for each class (Sec.~\ref{sec:128x128}). 
    A human study confirmed that our method produced more diverse and similarly realistic images compared to BigGAN samples (Sec.~\ref{sec:human}).
    % We also improved some mode-collapse BigGAN snapshots at $128\times128$ (Sec.~\ref{sec:snapshots}).

% 	\item By updating only the embeddings, we can harness a BigGAN trained on ImageNet to output images matching the Places365 scene classes (Sec.~\ref{sec:places}), revealing a surprising expressiveness of the embedding space.

\end{enumerate}

\section{Methods}

\subsection{Problem formulation}
\label{sec:objective}

% \an{Editing here}

Let $G$ be a class-conditional generator, here a BigGAN pre-trained by \cite{brock2019large}, that takes a class embedding $\vc \in \sR^{128}$ and a latent vector $\vz \in \R^{140}$ as inputs and outputs an image $G(\vc, \vz) \in \R^{256\times 256\times 3}$.
% The embeddings were learned during GAN training.
We test improving BigGAN's sample diversity by only updating the embeddings (pre-trained during GAN training).

\begin{figure*}[h]
	\centering
	\includegraphics[width=0.8\linewidth]{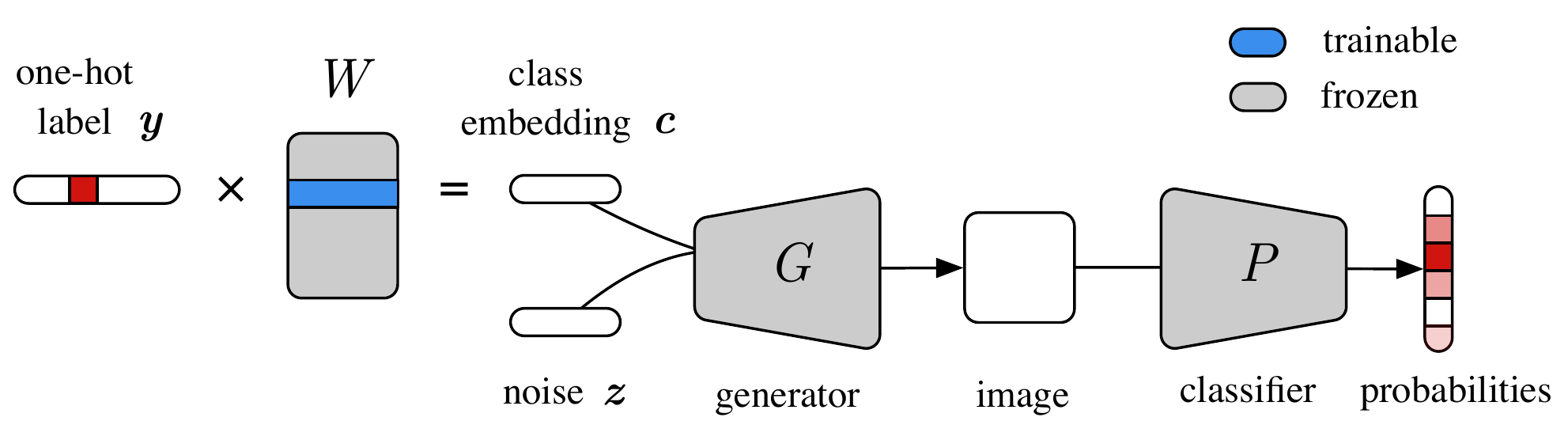}
	\caption{To improve the samples for a target class represented by a one-hot vector $\vy$, we iteratively take steps to find an embedding $\vc$ (\ie a row in the embedding matrix $W$) such that all the generated images $\{G(\vc, \vz^i)\}$, for different random noise vectors $\vz^i \sim \gN(0, I)$, would be (1) classified as the target class $\vy$; and (2) diverse \ie yielding different softmax probability distributions.
		We backpropagate through both the frozen, pre-trained generator $G$ and classifier $P$ and perform gradient descent to maximize the target-class probability of the generated samples over a batch of random latent vectors $\{\vz^i\}$.
	}
	\label{fig:concept}
\end{figure*}

\subsec{Increasing Diversity} 
Intuitively, we search for an input class embedding $\vc$ of the generator $G$ such that the set of output images $ \{ G(\vc, \vz^i) \}$ is diverse with random latent vectors $ \vz^i \sim \gN(0,I)$.
Specifically, we encourage a small change in the latent variable to yield a large change in the output image \cite{yang2019diversitysensitive} by maximizing:

\begin{equation}
\max_\vc ~ L_{\text{D}}(\vc) = \mathbb{E}_{\vz^i, \vz^j \sim \gN(0,I)}   \frac{\left \lVert \phi(G(\vc,\vz^i)) - \phi(G(\vc,\vz^j))\right \lVert}{\left \lVert \vz^i-\vz^j\right \lVert} %\right]
\label{eq:diversity}
\end{equation}

where $\phi(.)$ is a feature extractor.
In \cite{yang2019diversitysensitive}, $\phi(.)$ is an identity function to encourage pixel-wise diversity.
We also tested with $\phi(.)$ being outputs of the \layer{conv5} layer and the output \layer{softmax} layer of AlexNet.

Via hyperparameter tuning, we found that maximizing the above objective via $10$ unique pairs of ($\vz^i, \vz^j$) selected from $\gZ$ to be effective (full hyperparameter details are in Sec.~\ref{sec:implementation}).

\subsec{Activation maximization}
When a class embedding changes, it is critical to keep the generated samples to be still realistic and in the target class.
To achieve that, we also move the class embedding $\vc$ of the generator $G$ such that the output image $G(\vc, \vz)$ for any random $\vz \sim \gN(0,I)$ would cause some classifier $P$ to output a high probability for a target class $\vy$ (Fig.~\ref{fig:concept}).
Here, we let $P$ be a pre-trained ImageNet classifier \cite{krizhevsky2012imagenet} that maps an image $\vx \in \sR^{256\times 256\times 3}$ onto a softmax probability distribution over 1,000 output classes. 
%$\gY$.
Formally, we maximize the following objective given a pre-defined class $y_c$:

\begin{equation}
\label{eq:likelihood}
%\vspace{-0.3cm}
%\vc^* = \argmin_{\vc} \E_{\vz \sim \gN(0,I)} \Big[ - \text{log~}P(\vy~|~ G(\vc, \vz))  \Big]
\max_{\vc} ~ L_{\text{AM}}(\vc) =
\E_{\vz \sim \gN(0,I)}~\text{log~}P(\vy = y_c~|~ G(\vc, \vz)) 
%\newline
\end{equation}

The above objective is basically a common term in the classification objectives for class-conditional GAN discriminators \cite{odena2017conditional,brock2019large,mirza2014conditional} and also called the Activation Maximization (AM) in image synthesis using pre-trained classifiers \cite{nguyen2016synthesizing,nguyen2017plug,nguyen2019understanding,erhan2009visualizing,simonyan2013deep}.
We try to solve the above AM objective via mini-batch gradient descent.
That is, we iteratively backpropagate through both the classifier $P$ and the generator $G$ and change the embedding $\vc$ to maximize the expectation of the log probabilities over a set $\gZ$ of random latent vectors.

In sum, we encouraged the samples to be diverse but still remain in a target class $\vy$ via the full objective function below (where $\lambda$ is a hyperparameter):

\begin{align}
\max_\vc ~ L_{\text{AM-D}}(\vc) = L_{\text{AM}} + \lambda L_{\text{D}}
\label{eq:amd}
\end{align}

% \noindent where $\lambda$ is a hyperparameter to be tuned.

\subsection{Datasets and Networks}
\label{sec:dataset}

\subsec{Datasets} While the generators and classifiers were pre-trained on the full 1000-class ImageNet 2012 dataset, we evaluated our methods on a subset of 50 classes (hereafter, ImageNet-50) where we qualitatively found BigGAN samples exhibit the lowest diversity.
The selection of 50 classes were informed by two diversity metrics (see below) but decided by humans before the study.

\subsec{Generators}
We used two pre-trained ImageNet BigGAN generators \cite{brock2019large}, a $256\times256$ and a $128\times128$ model, released by the authors in PyTorch \cite{brock2019pytorch}.
For the purpose of studying diversity, all generated images in this paper were sampled from the full, non-truncated prior distribution \cite{brock2019large}.

\subsection{Evaluation metrics}

Because there is currently no single metric that is able to capture the multi-dimensional characteristics of an image set \cite{borji2019pros}, we chose a broad range of common metrics to measure sample diversity and sample realism separately.

\subsec{Diversity}
We measured intra-class diversity by randomly sampling 200 image pairs from an image set and computing the MS-SSIM \cite{odena2017conditional} and LPIPS \cite{zhang2018unreasonable} scores for each pair.
For each method, we computed a mean score across the 50 classes $\times$ 200 image pairs.

\subsec{Realism}
To measure sample realism, we used three standard metrics: Inception Score (IS) with 10 splits \cite{salimans2016improved}, Fr\'echet Inception Distance (FID) \cite{heusel2017gans}, and Inception Accuracy (IA) \cite{odena2017conditional}.
These three metrics were computed for every set of 50,000 images = 50 classes $\times$ 1000 images.
To evaluate the set of mixed samples from both BigGAN and AM embeddings, we randomly select 500 images from each and create a new set contains 1000 images per ImageNet class.

% \subsection{Generators}

% \textbf{Classifiers~} Our default image classifier is AlexNet \cite{krizhevsky2012imagenet} pre-trained on the 1000-class ImageNet 2012 dataset.
% Note that other ImageNet classifiers can also be used as shown in Sec.~\ref{sec:AM}.

% \textbf{Generators~}

\subsection{Implementation details}
\label{sec:implementation}

% Our AM method updates the original embedding following the gradients from a pre-trained classifier to maximize its log probabilities (Fig.~\ref{fig:concept}).

We found two effective strategies for implementing the AM method (described in Sec.~\ref{sec:objective}) to improve BigGAN samples: (1) searching within a small region around the original embeddings (AM-S); (2) searching within a large region around the mean embedding (AM-L).

\subsec{Hyperparameters}
For AM-S, we randomly initialized the embedding within a Gaussian ball of radius $0.1$ around the original embedding. We used a learning rate of $0.01$.
For AM-L, we randomly initialized the embedding around the mean of all 1000 embeddings and used a larger learning rate of $0.1$.
For both settings, we maximized Eq.~\ref{eq:likelihood} using the Adam optimizer and its default hyperparameters for $200$ steps.
We re-sampled a set $\gZ = \{\vz^i \}_{20}$ every $20$ steps.
Every step, we kept the embeddings within $[-0.59, 0.61]$ by clipping.
To evaluate each trial, we used the embedding from the last step and sampled 1000 images per class.
We ran 5 trials per class with different random initializations.
We used 2 to 4 $\times$ V100 GPUs for each optimization trial.

\subsec{Classifiers}
In the preliminary experiments, we tested four 1000-class-ImageNet classifiers: AlexNet \cite{krizhevsky2012imagenet}, Inception-v3 \cite{szegedy2016rethinking}, ResNet-50 \cite{he2016deep}, and a ResNet-50 \cite{engstrom2019learning} that is robust to pixel-wise noise.
By default, we resized the BigGAN output images to the appropriate input resolution of each classifier.

With Inception-v3, we achieved an FID score that is (a) substantially better than those for the other three classifiers (Table~\ref{tab:diff_classifiers}; 30.24 vs. 48.74), and (b) similar to that of the original BigGAN (30.24 vs. 31.36).
The same trends were observed with the Inception Accuracy metrics (Table~\ref{tab:diff_classifiers}).
However, we did not find any substantial qualitative differences among the samples of the four treatments.
Therefore, we chose AlexNet because of its fastest run time.

\section{Experiments and Results}

\subsection{Repairing complete mode-collapse classes of BigGANs}
\label{sec:AM}

Consistent with \cite{ravuri2019seeing}, we found that BigGAN samples for some classes, \eg \class{window} \class{screen}, contain similar, human-unrecognizable patterns (see Fig.~\ref{fig:repair_mode_collapse}a).
However, re-training BigGANs is impractical to most researchers given its significance computation requirement.

Here, we apply AM-L (see Sec.~\ref{sec:implementation}) to ``repair'' the mode-collapse \class{window~screen} embedding to generate more realistic and diverse images.
Intuitively, AM-L enables us to make a larger jump out of the local optimum than AM-S.

% describe the implementation details and results of applying our AM method (described in Sec.~\ref{sec:objective}) to ``repair'' the mode-collapse \class{window~screen} embedding to generate more realistic and diverse images.

% \qi{We observed that there are few complete-mode-collapse classes \eg \class{window~screen} whose images are rubbish (Fig.~\ref{fig:repair_mode_collapse}C) in BigGAN samples which is similar to the observation in \cite{ravuri2019seeing}. To fix this issue, we could re-train the BigGAN models and keep a better weights but the cost it significantly high (Table.~\ref{tab:cost_table}). We propose Activation Maximization to update an embedding following the gradients from an image classifier to maximize its log probabilities (Fig.~\ref{fig:concept}).
% } 

\noindent\textbf{Results~}
Interesting, by simply changing the embedding, AM-L was able to turn the original rubbish images into a diverse set of recognizable images of window screens (see Fig.~\ref{fig:repair_mode_collapse}a).
Quantitatively, the AM embedding improved BigGAN \class{window~screen} samples in all metrics: LPIPS (0.62 $\to$ 0.77), IS (2.76 $\to$ 2.91), and IA (0.56 $\to$ 0.7).

\begin{figure}[h]
	\centering
	\includegraphics[width=1.0\linewidth]{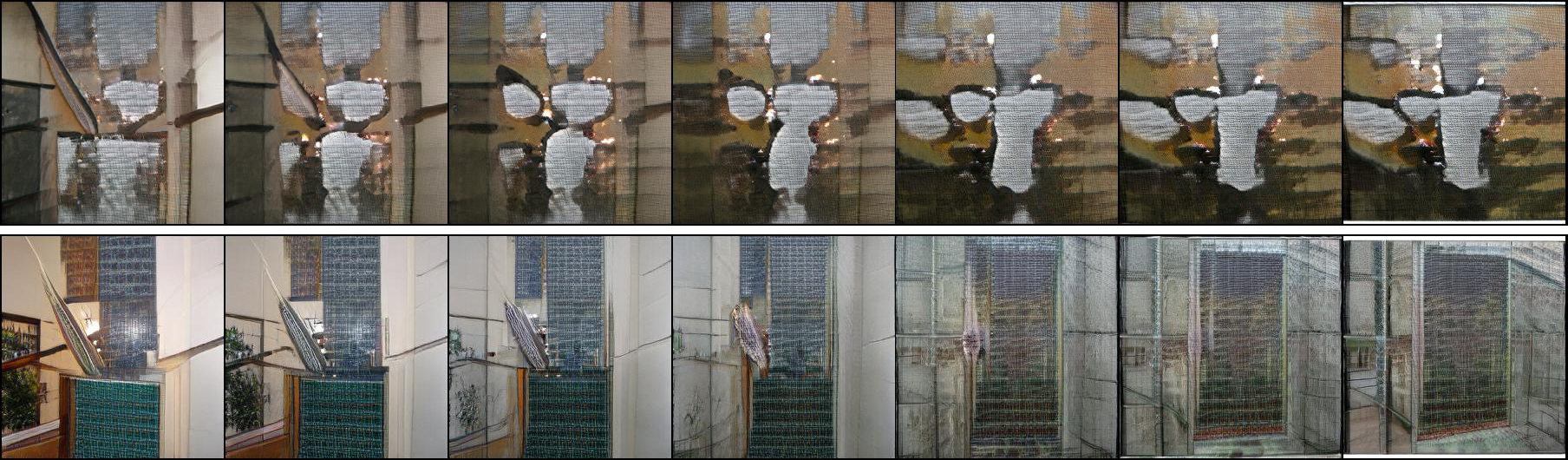}
	\caption{Interpolation between a $\vz$ pair in the \class{window~screen} class using the original BigGAN embedding (top) yields similar and unrealistic samples. 
		The same interpolation with the embedding found by AM (bottom) produced realistic intermediate samples between two window screen images.
	}
	\label{fig:interpolation}
\end{figure}

While the embeddings found by our AM methods changed the generated samples entirely, we observed that interpolating in the latent or embedding spaces still yields realistic intermediate samples (Fig.~\ref{fig:interpolation}).
% See Figs.~\ref{fig:inter_class_interpolation}--~\ref{fig:interpolate_z} for more examples of interpolation between $\vz$ pairs and between $\vc$ pairs \ie classes.

\noindent\textbf{Significantly faster computation~~}
According to a PyTorch BigGAN re-implementation by authors \cite{brock2019large}, BigGAN training can take at least 15 days on 8 V100 GPUs.
This is significantly more time-consuming and costly than our AM approach which takes at most 1 hour for generating 5 embeddings (from which users could choose to use one or more) on a single V100 GPU (see Table~\ref{tab:cost_table}).
The original DeepMind's training \cite{brock2019large} requires even more expensive and unique hardware of 256 Google Cloud TPU, which is not available to most of the community and so is not compared here.
% BigGAN training takes at least 4 weeks on 8 V100 GPUs

\begin{table}[h]
	\centering
	\tabcolsep=6pt
	\begin{tabular}{lrcr}
		&  &  &  \\
		%			\hline
		~~~~~~~~~Method \hfill & Time & Number of GPUs & AWS price  \\	
		& (hours) & (Tesla V100) & (USD)
		\\
		%			\hline
		\hline
		1. BigGAN training \cite{brock2019pytorch} & 24$\times$15 days=360 & 8 & 8812.8 \\ 	\hline
		2. AM optimization & 1  & 1  & 3.1 \\ 	\hline
		\vspace{0.1cm}
    \end{tabular} 
	\caption{BigGAN training is not only 360$\times$ more time-consuming but also almost 3,000$\times$ more costly.
	The AWS on-demand price-per-hour is \$ 24.48 for $8\times$V100 and \$ 3.06 for $1\times$V100 \cite{cost2020amazon}.}
% 	\qi{The comparison of the cost of training BigGAN and our AM. To find the new class embeddings for mode-collapse class, we can re-train the BigGAN, but the price is significantly high compared with our AM. On-Demand Price per hour is \$ 24.48 for $8\times$V100 and \$ 3.06 for $1\times$V100.}}
	\vspace{-0.8cm}
	\label{tab:cost_table}
\end{table}

Note that our method is essentially finding a new sampler for the same BigGAN model.
After a new embedding is found via optimization, the samples are generated fast via standard GAN sampling procedure \cite{goodfellow2014generative}.

% \subsection{Harnessing BigGAN pre-trained on ImageNet for synthesizing Places365 scene images}
\subsection{Synthesizing Places365 images using pre-trained ImageNet BigGAN}
\label{sec:places}

% \an{Editing here}

While original BigGAN is not able to synthesize realistic images for all 1000 ImageNet classes (see Fig.~\ref{fig:repair_mode_collapse}), it does so for a few hundred of classes.
% Our study has revealed that the BigGAN generator pre-trained on ImageNet is able to synthesize a wide variety of images.
Therefore, here, we test whether it is possible to re-use the same ImageNet BigGAN generator for synthesizing images for unseen categories in the target Places365 dataset \cite{zhou2017places}, which contains 365 classes of scene images.
For evaluation, we randomly chose 50 out of 365 classes in Places365 (hereafter, Places-50).

\subsec{Mean initialization} 
As we want to generate images for unseen classes, the Places365-optimal embeddings are intuitively far from the original ImageNet embeddings.
Therefore, we chose AM-L (instead of AM-S) for making larges jumps.
We ran the AM-L algorithm for 5 trials per class with the same hyperparameters as in Sec.~\ref{sec:AM} but with a ResNet-18 classifier \cite{he2016deep} pre-trained on Places365.

\begin{figure}[t]
	\centering
	\vspace{-0.2cm}
	{	
		\begin{flushleft}
			\hspace{0.4cm} (A) Places365 images
			\hspace{0.4cm} (B) Top-1 baseline (BigGAN)
			\hspace{0.4cm} (C) AM-L (ours)
		\end{flushleft}
	}
	\begin{subfigure}[b]{1.0\linewidth}
		\centering
		\includegraphics[width=1\linewidth]{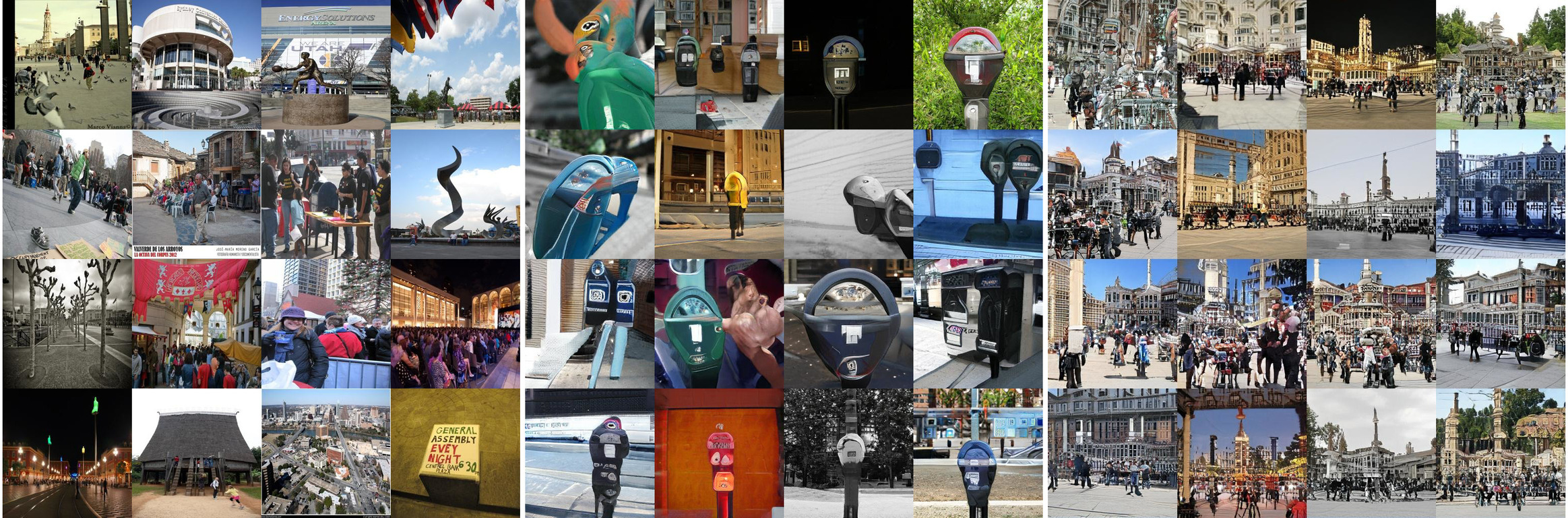}
		\vspace{-0.7cm}
		{
			\begin{flushleft}
				\hspace{1.6cm} \class{plaza}		
				\hspace{2.6cm}  \class{parking~meter}
				\hspace{2.5cm} \class{plaza}
			\end{flushleft}
		}
		
	\end{subfigure}
	
	\begin{subfigure}[b]{1.0\linewidth}
		\centering
		\includegraphics[width=1\linewidth]{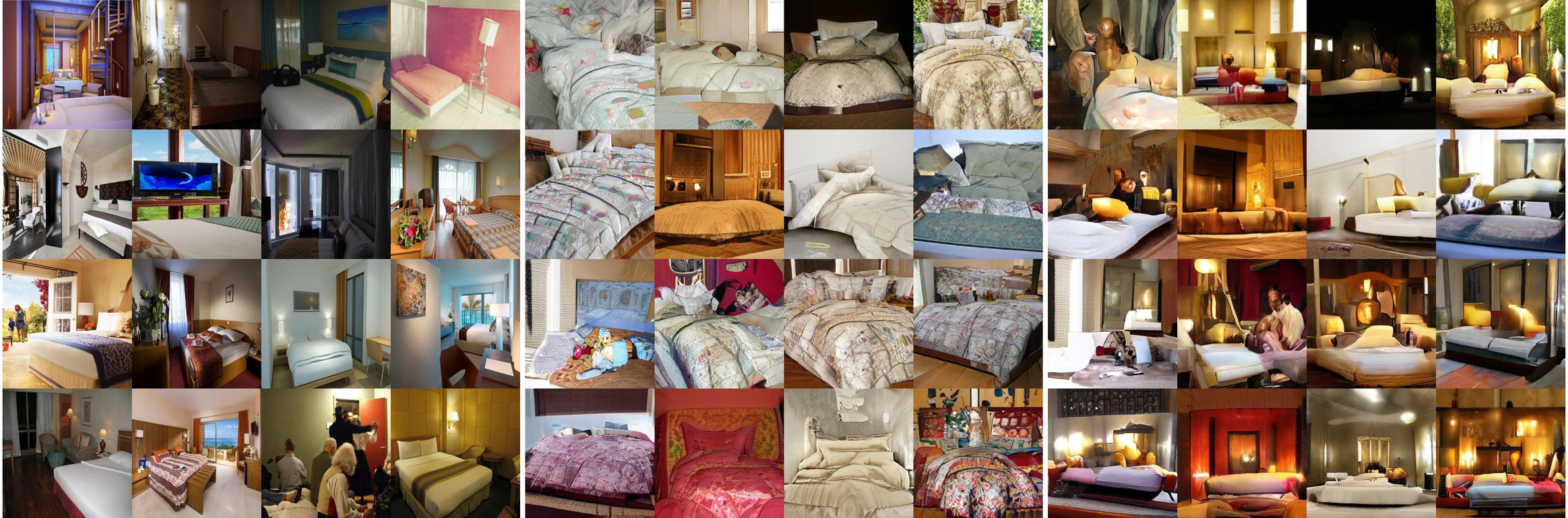}
		\vspace{-0.7cm}
		{
			\begin{flushleft}
				\hspace{1.2cm}  \class{hotel~room}		
				\hspace{2.8cm}  \class{quilt}
				\hspace{2.9cm}  \class{hotel~room}
			\end{flushleft}
		}
	\end{subfigure}

	\caption{
		AM-L generated plausible images for two Places365 classes, \class{plaza} (top) and \class{hotel~room} (bottom), which do \emph{not} exist in the ImageNet training set of the BigGAN generator.
		For example, AM-L synthesizes images of squares with buildings and people in the background for the \class{plaza} class (C) while the samples from the top-1 ImageNet class, here, \class{parking~meter}, shows parking meters on the street (B).
		Similarly, AM-L samples for the \class{hotel~room} class has the unique touches of lighting, lamps, and windows (C) that do not exist in the BigGAN samples for the \class{quilt} class (B).
		The latent vectors are held constant for corresponding images in (B) and (C).
		See Figs.~\ref{fig:mit_places_final_01}, \ref{fig:mit_places_final_02}, \ref{fig:mit_places_final_03}, and \ref{fig:mit_places_final_04} for more side-by-side image comparisons.
	}
	\vspace{-0.5cm}
	\label{fig:mit_teaser}
\end{figure}

\subsec{Top-5 initialization} Besides initializing from mean embeddings, we also tested initializing from the top-5 embeddings whose 10 random generated samples were given the highest average accuracy scores by the Places365 classifier.
For example, to synthesize the \class{hotel~room} images for Places365, the top-1 embedding in the ImageNet dataset is for class \class{quilt} (Fig.~\ref{fig:mit_teaser}).
We reproduced 5 AM-L trials but each was initialized with a unique embedding among the top-5.

\subsec{Baseline} We used the original BigGAN samples for the top-1 ImageNet classes found from the top-5 initialization procedure above as a baseline.

\begin{figure}[h]
	\centering
	\includegraphics[width=0.9\linewidth]{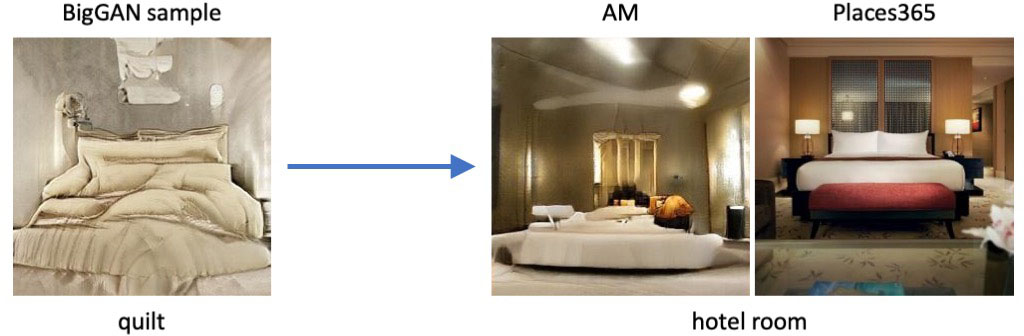}
	\caption{The closest ImageNet class that the BigGAN was pre-trained to generate is \class{quilt}, which contains mostly blankets and pillows.
	Surprisingly, with AM embeddings, the same BigGAN can generate remarkable images for unseen category of \class{hotel~room}. 
	The rightmost is an example Places365 image for reference.
	}
	\vspace{-0.4cm}
% 	\qi{The class embedding found by AM shows that it capture reasonable visual features of the target class \class{hotel room}}}
	\label{fig:am_mit_compare}
\end{figure}

\subsec{Qualitative Results} AM-L found many class embeddings that produced plausible images for Places365 scene classes using the same ImageNet BigGAN generator.
For example, to match the \class{hotel~room} class, which does not exist in ImageNet, AM-L synthesized bedroom scenes with lights and windows whereas the top-1 class (\class{quilt}) samples mostly shows beds with blankets (Fig.~\ref{fig:am_mit_compare}).
% (Fig.~\ref{fig:mit_teaser}).
See Fig.~\ref{fig:mit_teaser} for some qualitative differences between the generated images with original vs. AM embeddings for the same set of random latent vectors.

% More comparison for other classes are in 
% See Figs.~\ref{fig:mit_places_final_01}, \ref{fig:mit_places_final_02}, \ref{fig:mit_places_final_03}, \ref{fig:mit_places_final_04} for more image comparisons.

\subsec{Quantitative Results}
Compared to the baseline, AM-L samples have substantially higher realism in FID (41.25 vs. 53.15) and in ResNet-18 Accuracy scores (0.49 vs. 0.17).
In terms of diversity, AM-L and the baseline performed similarly and both were slightly worse than the real images in MS-SSIM (0.42 vs. 0.43) and LPIPS (0.65 vs. 0.70).
See Table~\ref{table:mit_eval} for detailed quantitative results.

\subsection{Improving sample diversity of 256$\times256$ BigGAN}
\label{sec:imagenet_256}

To evaluate the effectiveness of our method in improving sample diversity for many classes, here, we ran both AM-S and AM-L on 50 classes in ImageNet-50.
The goal is to compare the original BigGAN samples vs. a mixed set of samples generated from both the original BigGAN embeddings and AM embeddings found via our AM method.
That is, AM optimization is so inexpensive that users can generate many embeddings and use multiple of them to sample images.

\begin{figure*}[h]
	\centering
	\begin{subfigure}[b]{0.49\linewidth}
		\includegraphics[width=1.0\linewidth]{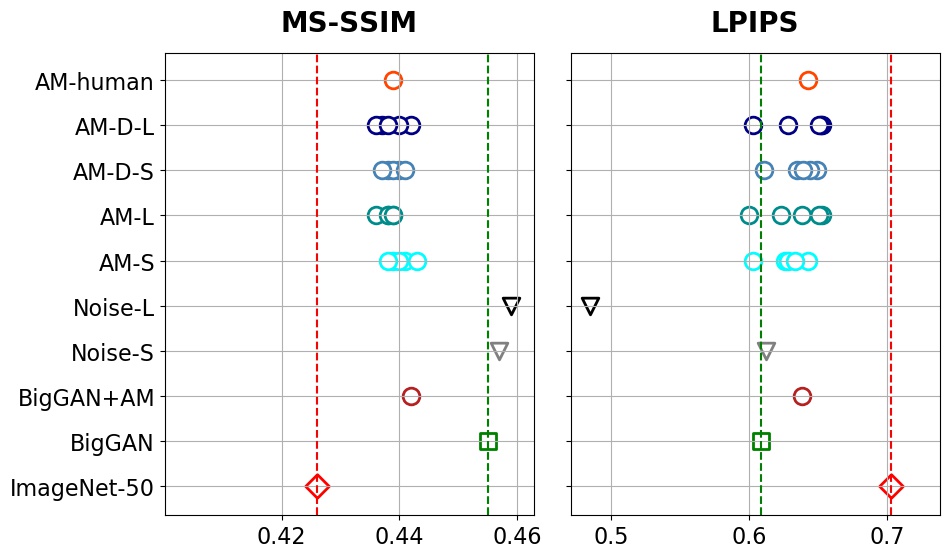}
% 		\caption{Diversity comparison in MS-SSIM and LPIPS metrics. }
		\caption{Diversity comparison.}
		\label{fig:biggan-256_ms-ssim_lpips_plots}
	\end{subfigure}
	\begin{subfigure}[b]{0.49\linewidth}
		\centering
		\includegraphics[width=1.0\linewidth]{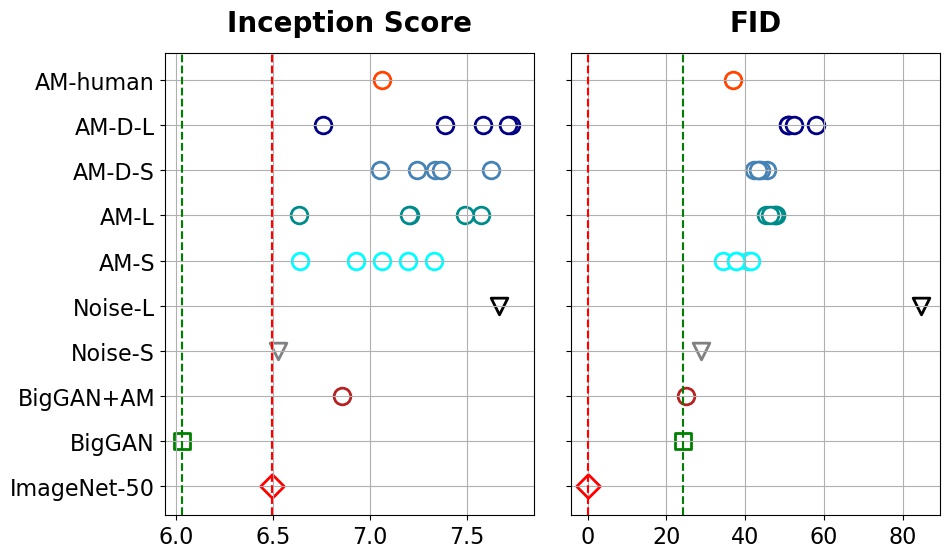}
% 		\caption{Realism comparison in IS and FID metrics. }
        \caption{Realism comparison.}
		\label{fig:biggan-256_is_fid_plots}
	\end{subfigure}
	\caption{
		Each point in the four plots is a mean score across 50 classes from one AM optimization trial or one BigGAN model.
		The ultimate goal here is to close the gap between the BigGAN samples (\textcolor{ao}{\small- - - -}) and the ImageNet-50 distribution (\textcolor{red}{\small- - - -}) in all four metrics.
		Naively adding noise degraded the embeddings in both diversity (MS-SSIM and LPIPS) and quality (IS and FID) scores \ie the black and gray $\nabla$ actually moved away from the red lines.
		Our optimization trials, on average, closed the \emph{diversity} gap by $\sim$50\% \ie the AM circles are half way in between the green and red dash lines (a).
% 		However, there was a trade-off between diversity vs. quality \ie on the IS and FID metrics, the AM circles went further away from the red line (b).
% 		\qi{By mixing new samples from AM to original samples from BigGAN (red \textcolor{red}{$\circ$}), the BigGAN+AM trail has better diversity (MS-SSIM and LPIPS) and similar quality (IS and FID) to BigGAN. It shows that multi-embeddings could improve the sample diversity of BigGAN without compromising the quality.}
        By mixing AM samples with the original BigGAN samples, the BigGAN+AM image-set (\textcolor{red}{$\circ$}) has substantially higher diversity (MS-SSIM and LPIPS) and similar quality (IS and FID) to BigGAN (\textcolor{green}{$\Box$}). That is, that multi-embeddings improved the sample diversity of BigGAN without compromising the quality.
		}
	\label{fig:biggan-256_eval_diversity_plots}
\end{figure*}

\subsec{BigGAN vs. AM}
Across 50 classes $\times$ 5 AM trials, we found that both AM-S and AM-L produced samples of higher diversity than the original BigGAN samples.
For both MS-SSIM and LPIPS, on average, our AM methods reduced the gap between the original BigGAN and the real data by $\sim$50\% (Fig.~\ref{fig:biggan-256_ms-ssim_lpips_plots}; AM-S and AM-L vs. BigGAN).

For all 50 classes, we always found at least 1 out of 10 trials (\ie both AM-S and AM-L combined) that yielded samples that match the real data in MS-SSIM or LPIPS scores.  
% (Fig.~\ref{fig:biggan-256_ms-ssim_lpips_plots}; AM-max vs. ImageNet-50).
The statistics also align with our qualitative observations that AM samples often contain a more diverse set of object poses, shapes and backgrounds than the BigGAN samples (see Figs.~\ref{fig:biggan-am_final_01}--\ref{fig:biggan-am_final_03}).
\label{sec:repair_mode_collapse}

% In terms of IA and FID scores, AM samples have lower realism than BigGAN samples (Table~\ref{table:main_result}).
% Given the known inflation issues with IS scores \cite{barratt2018note}, here, our IS scores (Fig.~\ref{fig:biggan-256_is_fid_plots}) suggest that AM did not improve the sample realism upon BigGAN.

\subsec{BigGAN vs. BigGAN+AM}
Most importantly, the set of images generated by both BigGAN and two AM embeddings obtained higher diversity in MS-SSIM and LPIPS while obtaining similar realism FID scores (Fig.~\ref{fig:biggan-256_eval_diversity_plots}; BigGAN vs. BigGAN$+$AM). We constructed each BigGAN$+$AM set per class using one BigGAN and one AM embedding (selected by humans out of 5 embeddings).

% In addition, the 50 embeddings found by AM when projected onto a 2-D t-SNE space still capture the class semantics similarly to the original BigGAN embeddings (see Fig.~\ref{fig:t-SNE_zoom_in} for side-by-side comparisons).

% \subsection{Semantically meaningful BigGAN class embeddings}
% \label{sec:low_diversity}

% We observed via t-SNE visualizations \cite{maaten2008visualizing} that the class embeddings learned by BigGAN captured well the semantics of the ImageNet classes.
% That is, we projected 1000 class embeddings $\vc^i \in \sR^{128}$ onto a 2-D t-SNE space.
% Interestingly, the embeddings for the low-diversity ImageNet-50 classes are far from being random \ie they were reasonably located in the neighborhood of related concepts (Fig.~\ref{fig:biggan_tsne_zoomin}; the \class{daisy} embedding is near other flowers and plants).
% The semantically meaningful t-SNE arrangements of the BigGAN class embeddings motivated us to search in the neighborhood of an original embedding to find a new vector that yields more diverse images (see the following sections).

\subsection{Adding noise to or finetuning the class embeddings did not improve diversity}
\label{sec:fintune}

\subsec{Adding noise}
A naive attempt to improve sample diversity is adding small random noise to the embedding vector of a low-diversity class.
Across 50 classes, we found that adding small noise $\sim \gN(0, 0.1)$ almost did not quantitatively change the image quality and diversity (Fig.~\ref{fig:biggan-256_eval_diversity_plots}; Noise-S) while adding larger noise $\sim \gN(0, 0.3)$ degraded the samples on both criteria (Fig.~\ref{fig:biggan-256_eval_diversity_plots}; Noise-L).

For example, \class{daisy} samples gradually turned into human-unrecognizable rubbish images as we increased the noise (Fig.~\ref{fig:biggan_noise}).

\subsec{Finetuning}
Another strategy to improve sample diversity is to finetune BigGANs.
However, how to finetune a BigGAN to improve its sample diversity is an open question.
The BigGAN pre-trained model would start to degrade if we kept training it using the original hyperparameters as reported in \cite{brock2019large}.

\begin{figure*}[t]
	\centering
	\begin{subfigure}[t]{0.49\linewidth}
		\centering
		\includegraphics[width=1.0\linewidth]{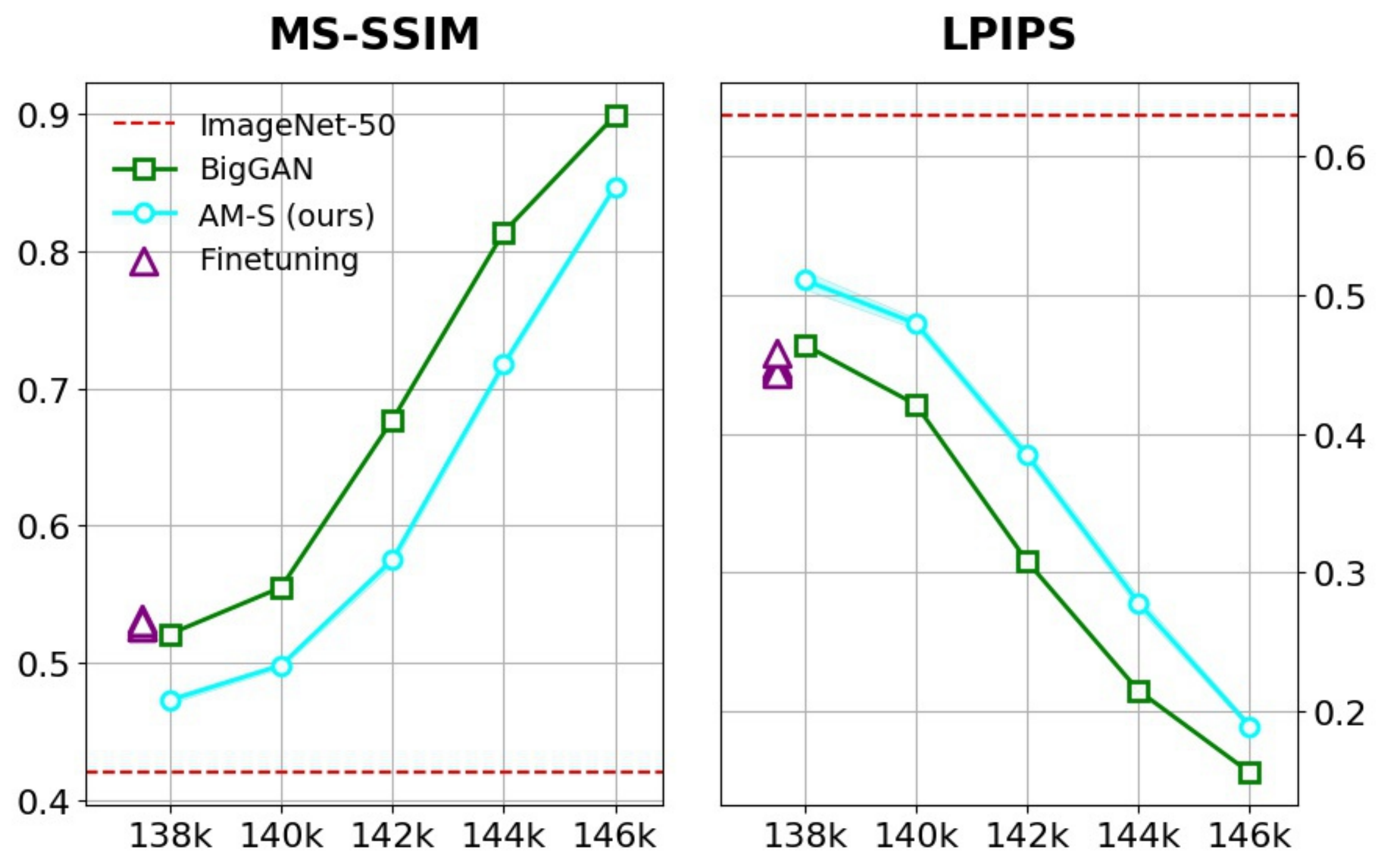}
% 		\caption{Diversity comparison in MS-SSIM and LPIPS metrics.}
        \caption{Diversity comparison}
		\label{fig:biggan-128_diversity}
	\end{subfigure}
	\begin{subfigure}[t]{0.49\linewidth}
		\centering
		\includegraphics[width=1.0\linewidth]{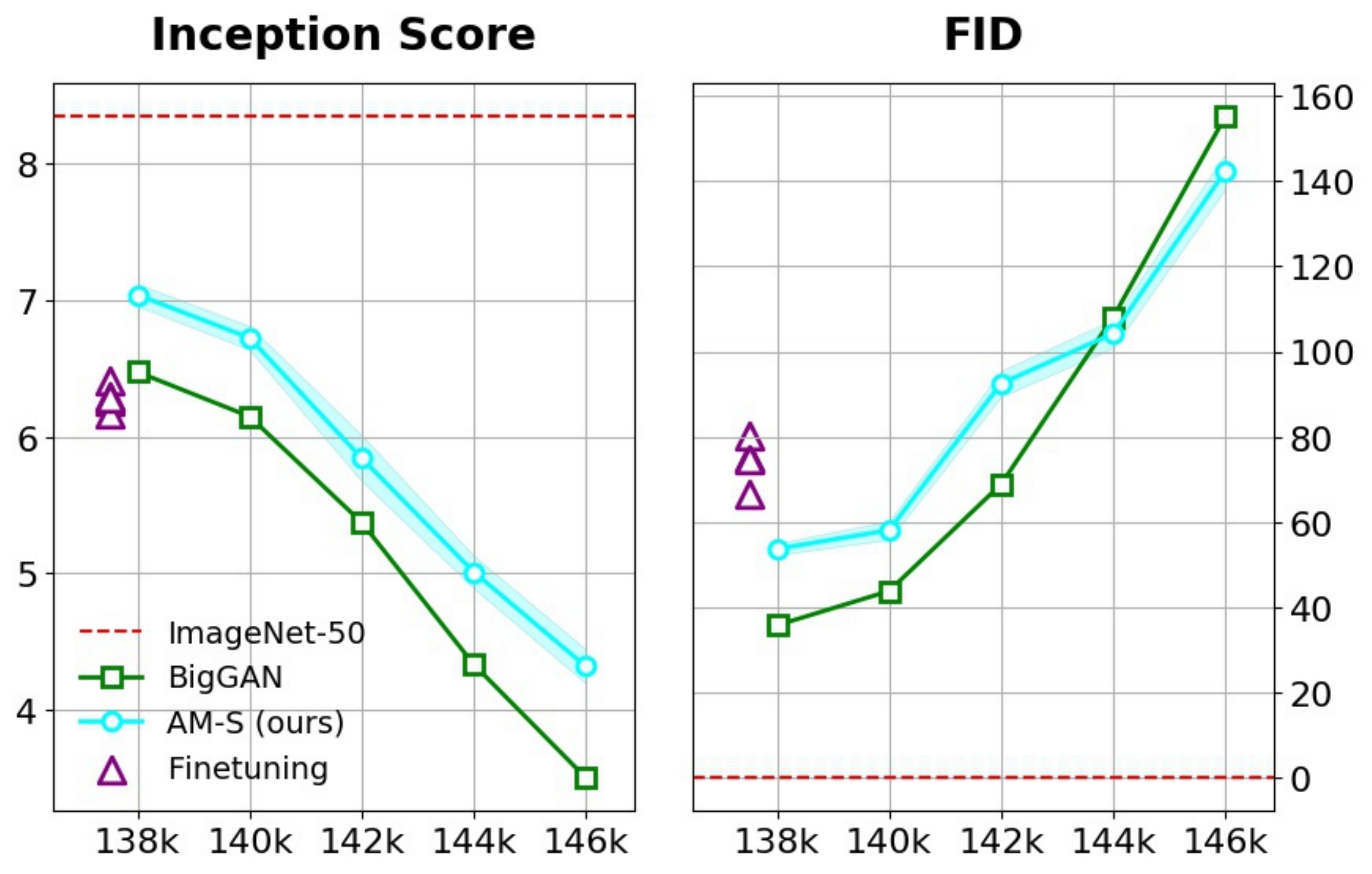}
% 		\caption{Realism comparison in IS and FID metrics.}
        \caption{Realism comparison}
		\label{fig:biggan-128_fid_plots}
	\end{subfigure}
	\caption{
		Each point in the four plots is a mean score across 50 classes and five AM-S trials or one $128\times128$ BigGAN model.
		Finetuning the 138k snapshot neither improved the sample diversity nor realism (purple 
		\textcolor{purple}{$\Delta$} vs. green \textcolor{ao}{$\Box$}).
		Optimizing the embeddings via AM-S consistently improved the diversity in both MS-SSIM and LPIPS (a).
		IS and FID metrics disagree on whether AM-S (cyan
		\textcolor{cyan}{$\circ$}) sample quality is better or worse than that of the BigGAN samples.
		See Fig.~\ref{fig:fix_biggan-128_701} for a side-by-side comparison of the samples from these five snapshots.
	}
	\label{fig:biggan-128_eval_quality_plots}
\end{figure*}

To minimize the GAN training instability and compare with other approaches in this paper, we only finetuned one embedding at a time, keeping the other embeddings and all parameters in the generator and discriminator frozen.
Because \cite{brock2019large} only released the discriminator for their $128\times128$ generator but not for the $256\times256$ model, we only finetuned the $128\times128$ model.
For each class, we added a small amount of noise $\sim \gN(0, 0.1)$ to the associated embedding vector and finetuned it using the original BigGAN training objective for 10 iterations until the training collapsed.
Across 50 classes $\times$ 5 trials, quantitatively, finetuning did not improve the sample diversity but lowered the realism (Fig.~\ref{fig:biggan-128_eval_quality_plots}; purple \textcolor{purple}{$\Delta$} vs. green \textcolor{ao}{$\Box$}).

\subsection{Explicitly encouraging diversity yielded worse sample realism}
\label{sec:amd}

Inspired by \cite{yang2019diversitysensitive}, here, we used the sample diversity further by incorporating a diversity term into the previous two AM-S and AM-L methods (Sec.~\ref{sec:objective}) to produce two new variants AM-D-S and AM-D-L.
We tested encouraging diversity in the (1) image space; (2) \layer{conv5} feature space; and (3) softmax outputs of AlexNet and found they can qualitatively bias the optimization towards different interesting spaces of diversity.

However, the addition of the diversity term quantitatively improved the diversity but at a large cost of lower sample quality (Fig.~\ref{fig:biggan-256_is_fid_plots} AM-S vs. AM-D-S and AM-L vs. AM-D-L).
Similarly, the IA scores of the AM-D methods were consistently lower than those of the original AM methods (Table~\ref{table:main_result}).
See Sec.~\ref{sec:amd_details} for more details.

We hypothesize that the intrinsic noise from mini-batch SGD \cite{wu2019noisy} also contributes to the increased sample diversity caused by AM embeddings.

\subsection{Humans rated AM samples more diverse and similarly realistic}
\label{sec:human}

Because quantitative image evaluation metrics are imperfect \cite{borji2019pros}, we ran a human study to compare the AM vs. original BigGAN samples.
For each class, across all 20 embeddings from 5 trials $\times$ 4 methods (AM-S, AM-L, AM-D-S, and AM-D-L), we manually chose one embedding that qualitatively is a balance between diversity and realism to sample images to represent our AM method in the study.
As a reference, this set of AM images were more diverse and less realistic than BigGAN samples according to the quantitative metrics  (Fig.~\ref{fig:biggan-256_eval_diversity_plots}; AM-human vs. BigGAN).

\subsec{Experiments}
We created two separate online surveys for diversity and realism, respectively.
For each class, the diversity survey showed a panel of $8\times8$ AM images side-by-side a panel of $8\times8$ BigGAN samples and asked participants to rate which panel is more diverse on the scale of 1--5.
That is, 1 or 5 denotes the left or right panel is clearly more diverse, while 3 indicates both sets are similarly diverse.
For each class, the AM and BigGAN panels were randomly positioned left or right.
The realism survey was a duplicate of the diversity except that each panel only showed $3\times3$ images so that participants could focus more on the details.

\subsec{Results}
For both tests, we had 52 participants who are mostly university students and do not work with Machine Learning or GANs.
On average, AM samples were rated to be more diverse and similarly realistic compared to BigGAN samples.
That is, AM images were given better than the neutral score of 3, \ie 2.24 $\pm$ 0.85 in diversity and 2.94 $\pm$ 1.15 in realism.

Also, AM samples were rated to be more diverse in 42/50 classes and more realistic in 22/50 classes.
See Figs.~\ref{fig:biggan-am_final_01}--\ref{fig:biggan-am_final_03} for your own comparisons.

% \subsection{AM embeddings still capture semantics and enable realistic interpolations}

% In addition, the 50 embeddings found by AM when projected onto a 2-D t-SNE space still capture the class semantics similarly to the original BigGAN embeddings (see Fig.~\ref{fig:t-SNE_zoom_in} for side-by-side comparisons).

\subsection{Generalization to a $128\times128$ BigGAN}
\label{sec:128x128}

To test whether our method generalizes to a different GAN at a lower resolution, we applied our AM-S method (see Sec.~\ref{sec:AM}) to a pre-trained $128\times128$ BigGAN released by \cite{brock2019pytorch}.
As in previous experiments, we ran 50 classes $\times$ 5 trials in total.
To evaluate each trial, we used the last-step embedding to sample 1000 images per class.

Consistent with the result on the $256\times256$ resolution, here, AM-S improved the diversity over the pre-trained model on both MS-SSIM and LPIPS (Fig.~\ref{fig:biggan-128_diversity}; 138k).
In terms of quality, FID and IS showed a mixed result of whether AM-S sample realism is lower or higher.
See Fig.~\ref{fig:biggan-am_128_final_01} for side-by-side comparisons.

\subsection{Generalization to different training snapshots of $128\times128$ BigGAN}
\label{sec:snapshots}

\begin{figure*}[t]
	\centering
	{	
		\begin{flushleft}
			\hspace{0.6cm} (A) Real
			\hspace{0.6cm} (B) 138k
			\hspace{0.6cm} (C) 140k
			\hspace{0.6cm} (D) 142k
			\hspace{0.6cm} (E) 144k
			\hspace{0.6cm} (F) 146k
		\end{flushleft}
	}
% 	\vspace*{0.3cm}
	{
		\begin{flushleft}
			\rotatebox{90}{\hspace{1cm}AM (ours)\hspace{0.6cm}BigGAN~~~}
		\end{flushleft}	
		\vspace{-3.3cm}
	}
	\vspace{-2.2cm}
	{
		\begin{flushright}
			\includegraphics[width=0.97\linewidth]{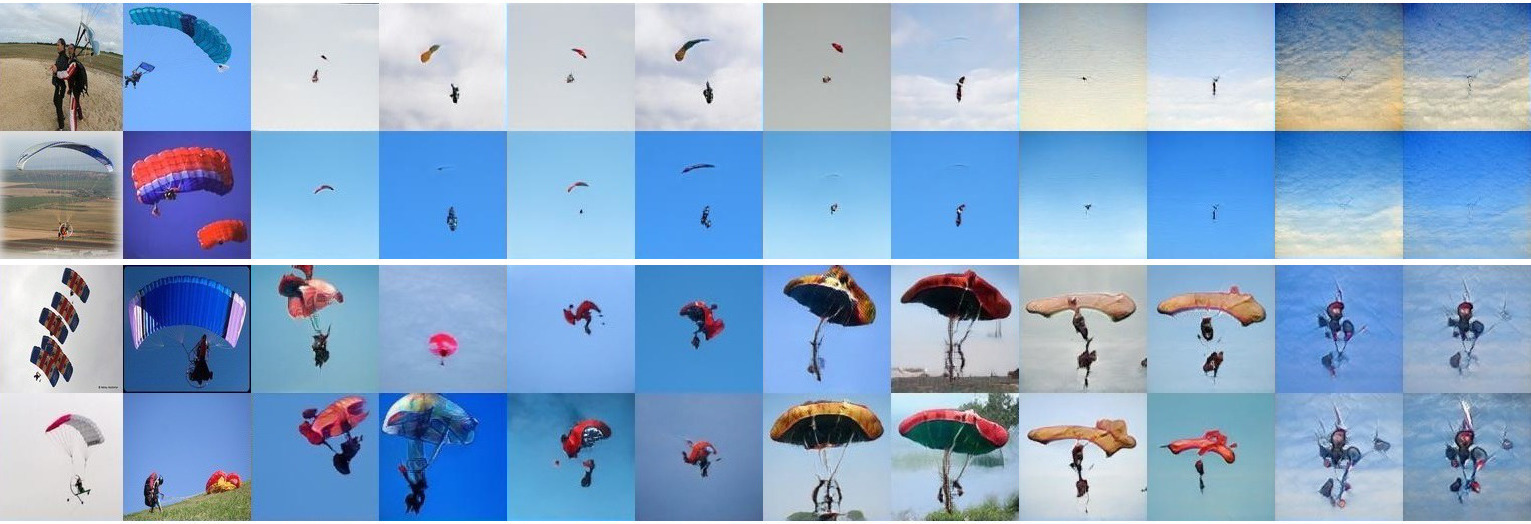}
		\end{flushright}
	}
	\caption{
		For the \class{parachute} class, the original $128\times128$ BigGAN samples (top panel) mostly contained tiny parachutes in the sky (B) and gradually degraded into blue sky images only (C--F).
		AM (bottom panel) instead exhibited a more diverse set of close-up and far-away parachutes (B) and managed to paint the parachutes for nearly-collapsed models (E--F).
		The samples in this figure correspond to the five snapshots (138k---146k) reported in the quantitative comparison in Fig.~\ref{fig:biggan-128_eval_quality_plots}.
		See Figs.~\ref{fig:fix_biggan_128_snapshot_701},  \ref{fig:fix_biggan_128_snapshot_715},  \ref{fig:fix_biggan_128_snapshot_530} for more qualitative comparisons.
	}
	\label{fig:fix_biggan-128_701}
\end{figure*}

We have shown that BigGAN sample diversity can be improved substantially by changing only the embeddings (Sec.~\ref{sec:AM}) which revealed that the generator was actually capable of synthesizing those diverse images.
Here, we test how much sample diversity and quality can be improved by AM as the BigGAN training gradually collapses, which might impair not only the embeddings but also the generator's parameters.

\subsec{Experiments}
We took the pre-trained $128\times128$ BigGAN model (saved at the 138k-th iteration) and continued training it for 9000 more iterations with the same hyperparameters as in \cite{brock2019pytorch}.
We applied the AM-S method with the same hyperparameters as in Sec.~\ref{sec:128x128} to four BigGAN snapshots captured at the 140k, 142, 144k, and 146k iteration, respectively.

\subsec{Results}
AM-S consistently improved the sample diversity of all snapshots.
For some classes, AM qualitatively improved both sample diversity and quality (Figs.~\ref{fig:fix_biggan-128_701} and~\ref{fig:fix_biggan_128_snapshot_701}--\ref{fig:fix_biggan_128_snapshot_530}).
However, the diversity and realism of both AM-S and the original BigGAN samples gradually dropped together (Fig.~\ref{fig:biggan-128_eval_quality_plots}; AM-S vs. BigGAN).
The result suggests that, as the GAN training gradually collapsed, the synthesis capability is so degraded that changing the class embeddings alone is not sufficient to significantly improve the samples.

\section{Related work}
\label{sec:related_work}

\textbf{Latent space traversal~} 
Searching in the latent space of a GAN generator network to synthesize images has been shown effective for many tasks including (1) in-painting \cite{yeh2017semantic}; (2) image editing \cite{zhu2016generative}; (3) creating natural adversarial examples \cite{zhao2018generating}; or (4) feature visualization \cite{nguyen2019understanding}. 
While all prior work in this line of research optimized the latent variable $\vz$, we instead optimize the class embeddings $\vc$ of a class-conditional generator over a set of random $\vz$ vectors.

Our method might be the most related to Plug \& Play Generative Networks (PPGN) \cite{nguyen2017plug} in that both methods sample from the distribution $p_G(\vx,\vy)$ jointly defined by a generator and a pre-trained classifier.
While \cite{nguyen2017plug} trained an unconditional generator that inverts the features of an ImageNet classifier, our method is generally applicable to any pre-trained class-conditional generator.
Importantly, our goal is novel---to improve the sample diversity of any pre-trained class-conditional generator (here, BigGANs) by changing its class embeddings.

\subsec{Improving sample quality}
Two methods, MH-GAN \cite{turner2019metropolis} and DRS \cite{azadi2018discriminator}, have recently been proposed to improve the samples of a pre-trained GAN by harnessing the discriminator to reject low-probability generated samples.
However, these methods are able to only improve sample \emph{quality} but not diversity.
In addition, they assume that the discriminator is (a) available, which may not always be the case \eg in the official BigGAN releases \cite{brock2019large}; and (b) optimally trained for their samplers to recover exactly the true distribution.
Similar to MH-GAN and PPGN, our method is similar to a Markov chain Monte Carlo (MCMC) sampler that has no rejection steps.
A major difference is that we only perform the iterative optimization \emph{once} to update the embedding matrix. 
After a desired embedding is found, our subsequent samplings of images are fast following standard GANs.
In contrast, MH-GAN, DRS, and PPGN samplers often require many rejection or update steps to produce a single image.

% \subsec{Improving sample diversity}
% Many GAN regularization tricks have been introduced to encourage the samples to be diverse (see \cite{wang2019generative} for a survey).
% However, all prior methods require re-training GANs from scratch, which can be computationally expensive \eg in the BigGAN's case.
% Fine-tuning GANs may be a more efficient approach \cite{wang2018transferring}. 
% However, finetuning (1) requires both the pre-trained generator and discriminator, which is not always available in practice, and (2) is subject to the known training instability issues (as in Sec.~\ref{sec:fintune}).
% Here, our method is not subject to the above issues and can be viewed as finetuning only the embedding layer but using a maximum likelihood objective instead of a GAN objective.

\subsec{Generalization}
Understanding the image synthesis capability of a trained GAN generator is an active research area.
Recent findings showed that GANs trained on a dataset of scene images contain neurons that can paint common objects such as ``trees'' or ``doors'' \cite{bau2018visualizing}.
\cite{jahanian2019steerability} found that BigGAN is able to perform some general image transforms such as zoom, rotate or brightness adjustment up to a certain limit.
However, these methods optimize only the latent variable \cite{jahanian2019steerability} or both the latent and the generator parameters \cite{bau2018visualizing}, but not the class embeddings as ours.

\section{Conclusion}

We showed that the low sample diversity of pre-trained GAN generators can be improved by simply changing the class embeddings, not the generator.
Note that one could ``recover'' the missing modes using our AM methods and improve the sample quality further by sampling from a truncated prior distribution \cite{brock2019large}. 
Our method is also a promising method for de-biasing GAN models.
Compared to finetuning or re-training BigGANs from scratch, our method is more tractable even considering that one has to run five 200-step optimization trials to find a desired class embedding.

\section*{Acknowledgment}

The authors thank Chirag Agarwal and Naman Bansal for valuable feedback.
AN is supported by the National Science Foundation under Grant No.~1850117, Adobe Research, and GPU donations from Nvidia.

%===========================================================
\newpage
\bibliographystyle{splncs}
\bibliography{references}

\clearpage
\appendix
\onecolumn

\renewcommand{\thesection}{S\arabic{section}}
\renewcommand{\thesubsection}{\thesection.\arabic{subsection}}

\newcommand{\beginsupplementary}{%
	\setcounter{table}{0}
	\renewcommand{\thetable}{S\arabic{table}}%
	\setcounter{figure}{0}
	\renewcommand{\thefigure}{S\arabic{figure}}%
	\setcounter{section}{0}
}

\beginsupplementary

%Supplementary material

\begin{center}
	{\Large\sc Supplementary Material}
\end{center}

%\section{SUPPLEMENTARY MATERIAL}

\begin{table}[h]
	\centering
	\scalebox{0.8}{
		\begin{tabular}{lccccc}
			&  &  &  &  &  \\
			%			\hline
			~~~~Method \hfill & IS (10 splits) & FID & Inception Accuracy & MS-SSIM & LPIPS \\
			& (higher=better) & (lower=better) & (higher=better) & (lower=better) & (higher=better) \\			
			%			\hline
			\hline
			1. ImageNet-50 (real) & 6.49 $\pm$ 0.40 & N/A & 0.90 & 0.43 $\pm$ 0.04 & 0.70 $\pm$ 0.08 \\ 	\hline
			2. BigGAN & 6.03 $\pm$ 0.76 & 24.34 & 0.87 & 0.46 $\pm$ 0.05 & 0.61 $\pm$ 0.09 \\ 	\hline
			3. BigGAN $+$ AM & 6.85 $\pm$ 0.58 & 24.93 & 0.80 & 0.44 $\pm$ 0.03 & 0.64 $\pm$ 0.08 \\ 	\hline
			%			\hline
			4. Noise-S & 6.53 $\pm$ 0.86 & 28.75 & 0.82 & 0.46 $\pm$ 0.05 & 0.61 $\pm$ 0.09 \\ 	\hline
			5. Noise-L & 7.67 $\pm$ 0.95 & 84.61 & 0.36 & 0.46 $\pm$ 0.05 & 0.49 $\pm$ 0.04 \\ 	\hline
			%			\hline
			\multicolumn{6}{l}{6. AM-S}   \\ 	%\hline
			
			~~~~a. Best LPIPS trial \hfill & 7.33 $\pm$ 0.73 & 40.82 & 0.72 & 0.44 $\pm$ 0.05 & 0.64 $\pm$ 0.08 \\
			~~~~b. Average \hfill & 7.03 $\pm$ 0.71 & 38.39 & 0.74 & 0.44 $\pm$ 0.05 & 0.63 $\pm$ 0.08 \\ 	\hline
			%			\hline
			\multicolumn{6}{l}{7. AM-L}   \\ 	%\hline
			
			~~~~a. Best LPIPS trial \hfill & 7.49 $\pm$ 0.81 & 47.25 & 0.64 & 0.44 $\pm$ 0.04 & 0.65 $\pm$ 0.08 \\ 	%\hline
			~~~~b. Average \hfill & 7.22 $\pm$ 0.79 & 46.86 & 0.68 & 0.44 $\pm$ 0.05 & 0.63 $\pm$ 0.08 \\ 	\hline
			%			\hline
			\multicolumn{6}{l}{8. AM-D-S}  \\ 	%\hline
			
			~~~~a. Best LPIPS trial \hfill & 7.62 $\pm$ 0.90 & 45.61 & 0.66 & 0.44 $\pm$ 0.04 & 0.65 $\pm$ 0.08 \\ 	%\hline
			~~~~b. Average \hfill & 7.32 $\pm$ 0.80 & 43.78 & 0.68 & 0.44 $\pm$ 0.05 & 0.64 $\pm$ 0.08 \\ 	\hline
			%			\hline
			\multicolumn{6}{l}{9. AM-D-L}   \\ 	%\hline
			
			~~~~a. Best LPIPS trial \hfill & 7.58 $\pm$ 0.84 & 50.94 & 0.64 & 0.44 $\pm$ 0.04 & 0.65 $\pm$ 0.08 \\ 	%\hline
			~~~~b. Average \hfill & 7.43 $\pm$ 0.85 & 52.68 & 0.61 & 0.44 $\pm$ 0.05 & 0.64 $\pm$ 0.08 \\ 	\hline
	\end{tabular}}
	\vspace{0.2cm}
	\caption{
		We compared Activation Maximization (AM) samples with the BigGAN samples and the real ImageNet-50 images on two diversity metrics (MS-SSIM and LPIPS) and three realism metrics, Inception Score (IS), Fr\'echet Inception Distance (FID), and Inception Accuracy (IA).
		%		We apply different metrics to evaluate the quality and diversity of images. 
		ImageNet-50 is a subset of ImageNet that contains 50 classes where BigGAN samples exhibit limited diversity (see Sec.~\ref{sec:dataset}).
		%		 for more description of the ImageNet-50 low-diversity classes.
		For each AM method, we ran 50 classes $\times$ 5 trials and reported here (a) the trial with the best LPIPS score and (b) the average across 5 runs.
		In MS-SSIM and LPIPS, all AM trials consistently produced more diverse samples than the BigGAN samples.
		However, FID and IA scores indicated that AM samples are worse in realism compared to the original BigGAN samples.
		See Fig.~\ref{fig:biggan-256_eval_diversity_plots} for some graphical plots of this table.  }
	\label{table:main_result}
\end{table}

\section{Explicitly encouraging diversity yielded worse sample realism}
\label{sec:amd_details}

We found that in $\sim$2\% of the AM-S and AM-L trials, the optimization converged at a class embedding that yields similar images for different random latent vectors.
Here, we try to improve the sample diversity further by incorporating a specific regularization term into the AM formulation (as described in Sec.~\ref{sec:objective}).

\begin{figure*}[h!]
	\centering
	\begin{subfigure}[b]{1.0\linewidth}
		\centering
		\includegraphics[width=1.0\linewidth]{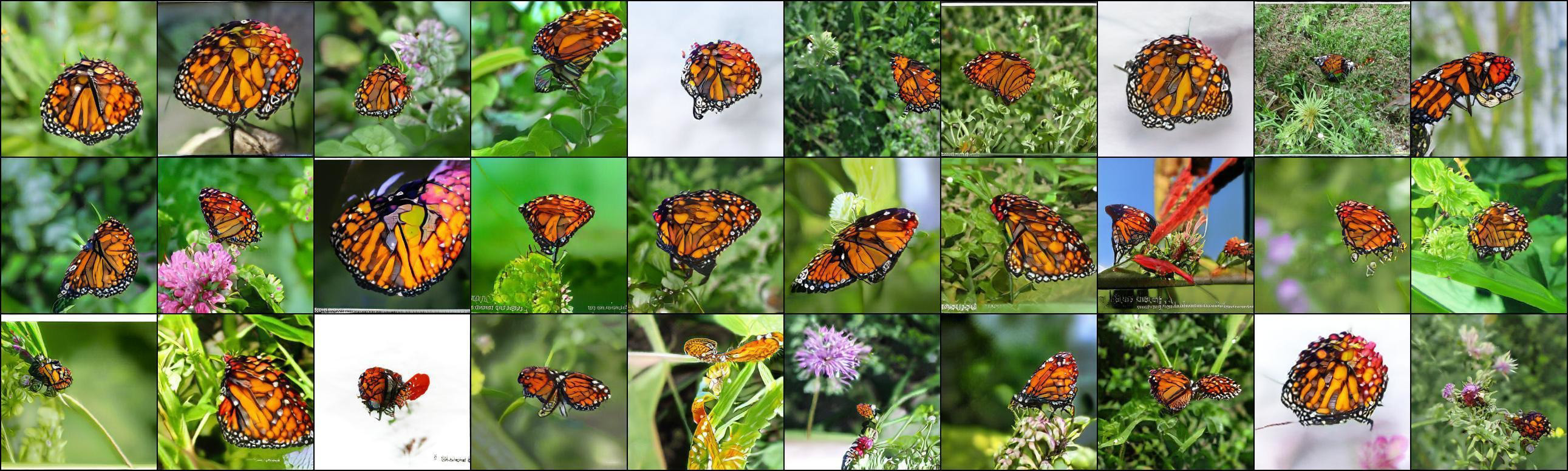}
		\caption{AM alone without the diversity term (\ie $\lambda = 0$ in Eq.~\ref{eq:amd}).}
	\end{subfigure}
	\begin{subfigure}[b]{1.0\linewidth}
		\centering
		\includegraphics[width=1.0\linewidth]{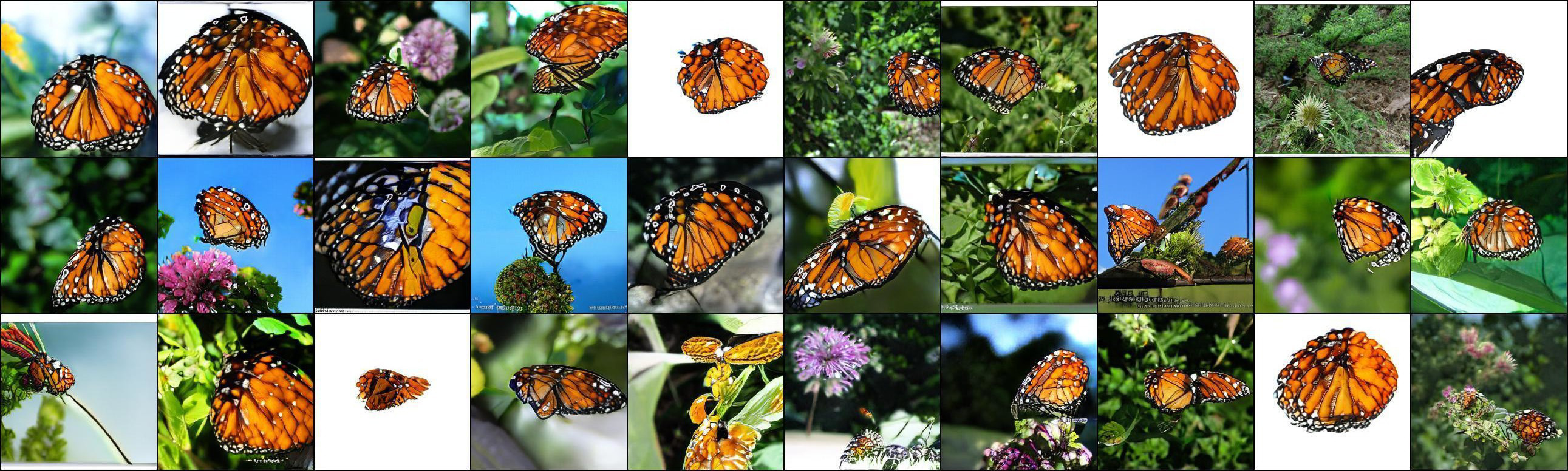}
		\caption{AM with the pixel-wise diversity term (\ie $\lambda = 0.01$ in Eq.~\ref{eq:amd}).\label{fig:pixel_diversity}}
	\end{subfigure}
	\begin{subfigure}[b]{1.0\linewidth}
		\centering
		\includegraphics[width=1.0\linewidth]{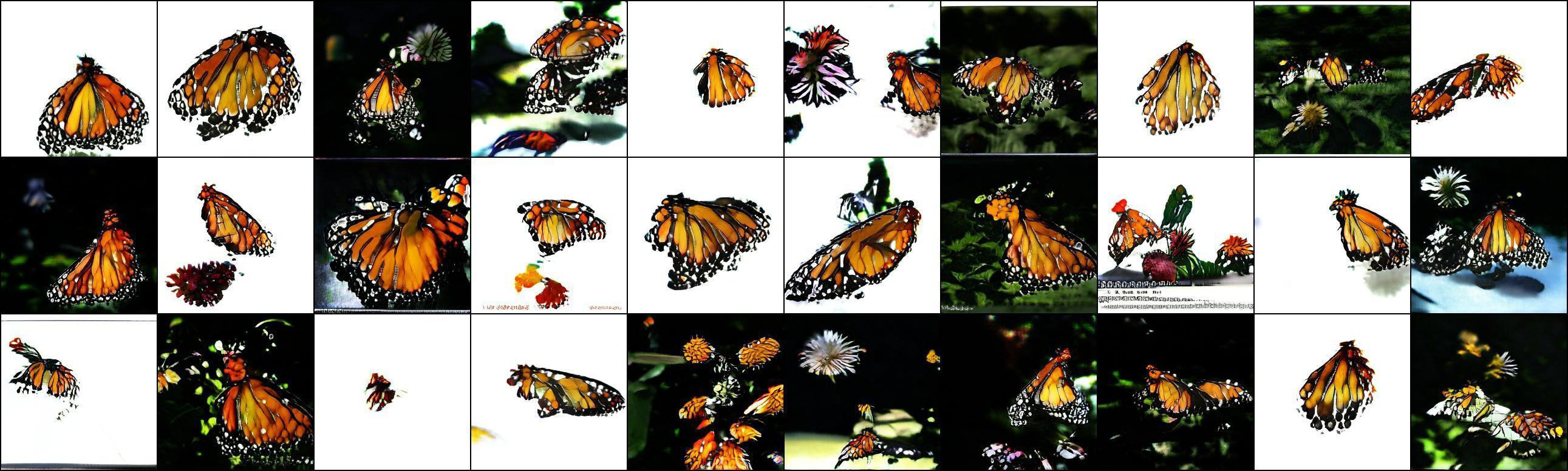}
		\caption{AM with the pixel-wise diversity term (\ie $\lambda = 0.1$ in Eq.~\ref{eq:amd}).\label{fig:high_alpha_pixel}}
	\end{subfigure}
	\begin{subfigure}[b]{1.0\linewidth}
		\centering
		\includegraphics[width=1.0\linewidth]{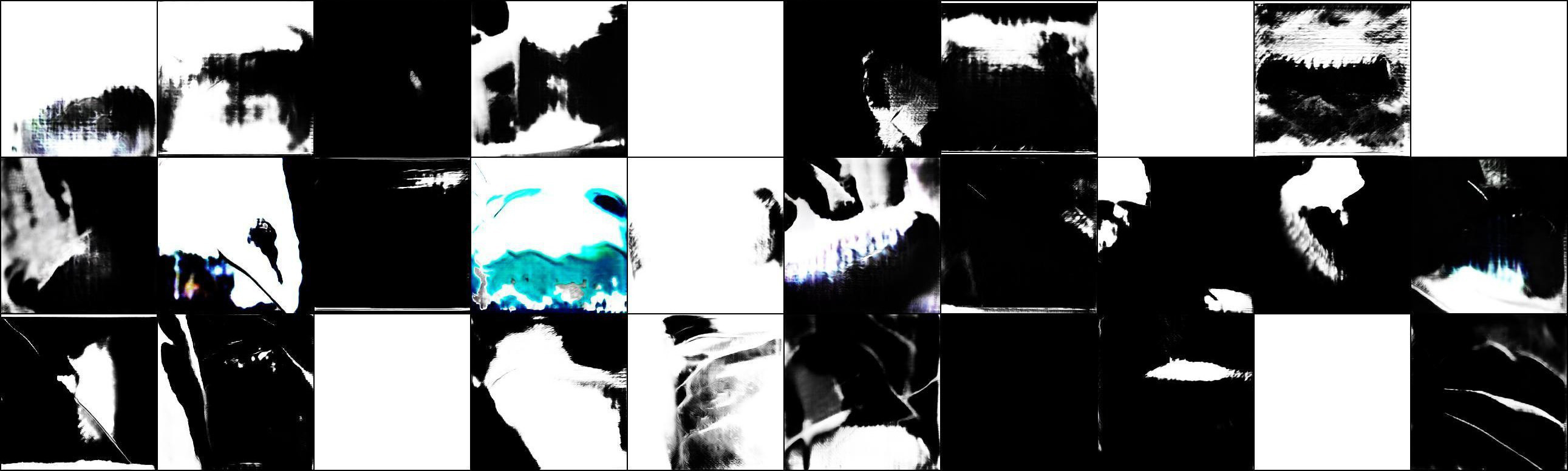}
		\caption{AM with the pixel-wise diversity term (\ie $\lambda = 1.0$ in Eq.~\ref{eq:amd}).}
	\end{subfigure}
	\caption{The \class{monarch~butterfly} class (323) samples generated by Activation Maximization (AM) methods when increasing the multiplier $\lambda$ of a pixel-wise diversity regularization term in Eq.~\ref{eq:amd}.}
	\label{fig:pixelwise_dl}
\end{figure*}

\begin{figure*}[h!]
	\centering
	\begin{subfigure}[b]{1.0\linewidth}
		\centering
		\includegraphics[width=1.0\linewidth]{images/323_biggan-am_without_dl_sample.jpg}
		\caption{AM alone without the diversity term (\ie $\lambda = 0$ in Eq.~\ref{eq:amd}).}
	\end{subfigure}
	\begin{subfigure}[b]{1.0\linewidth}
		\centering
		\includegraphics[width=1.0\linewidth]{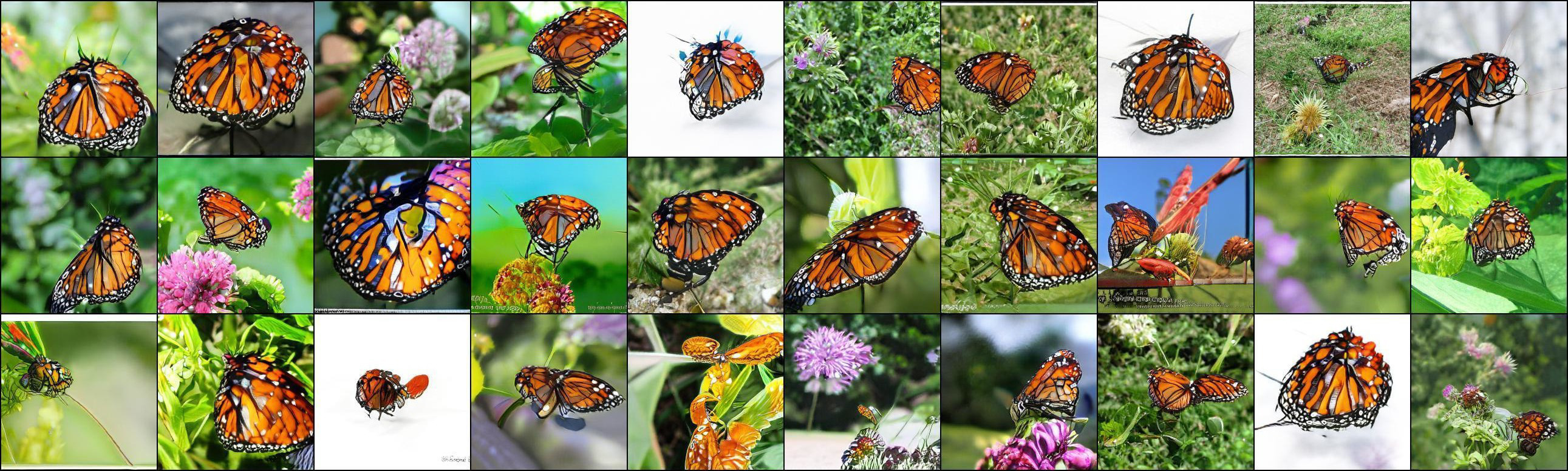}
		\caption{AM with a feature diversity term (\ie $\lambda = 0.01$ in Eq.~\ref{eq:amd}).\label{fig:conv5_diversity}}
	\end{subfigure}
	\begin{subfigure}[b]{1.0\linewidth}
		\centering
		\includegraphics[width=1.0\linewidth]{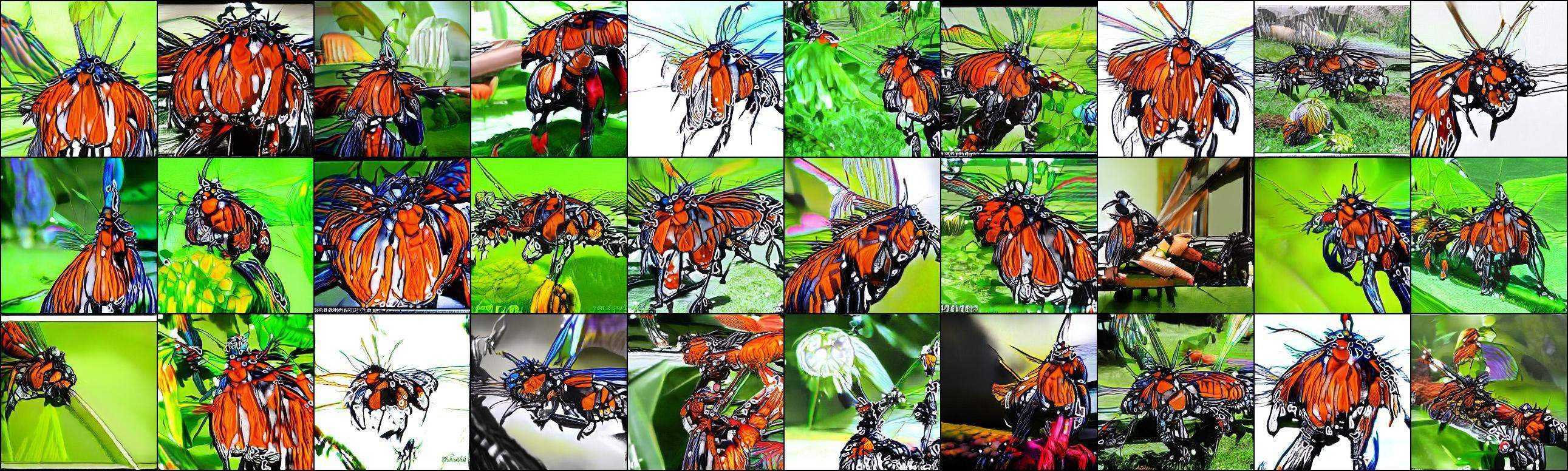}
		\caption{AM with a feature diversity term (\ie $\lambda = 0.1$ in Eq.~\ref{eq:amd}).}
	\end{subfigure}
	\begin{subfigure}[b]{1.0\linewidth}
		\centering
		\includegraphics[width=1.0\linewidth]{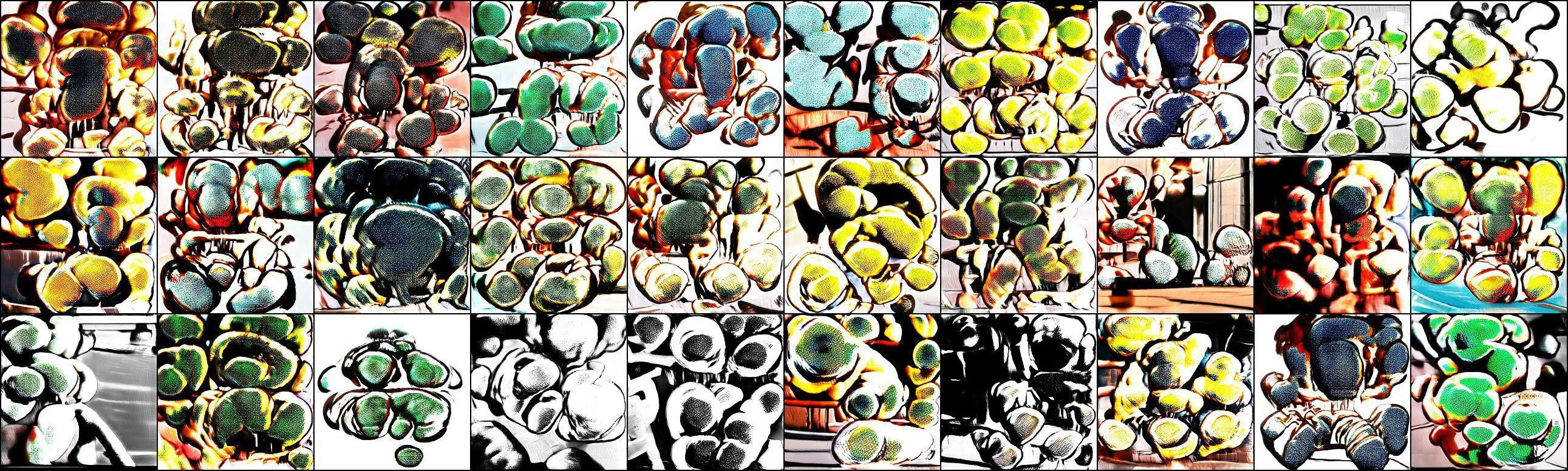}
		\caption{AM with a feature diversity term (\ie $\lambda = 1.0$ in Eq.~\ref{eq:amd}).\label{fig:conv5_high_alpha}}
	\end{subfigure}
	\caption{The \class{monarch~butterfly} class (323) samples generated by Activation Maximization (AM) methods when increasing the multiplier $\lambda$ of a \layer{conv5} feature diversity regularization term in Eq.~\ref{eq:amd}.}
	\label{fig:feature_dl}
\end{figure*}

\begin{figure*}[h!]
	\centering
	\begin{subfigure}[b]{1.0\linewidth}
		\centering
		\includegraphics[width=1.0\linewidth]{images/323_biggan-am_without_dl_sample.jpg}
		\caption{AM alone without the diversity term (\ie $\lambda = 0$ in Eq.~\ref{eq:amd}).}
	\end{subfigure}
	\begin{subfigure}[b]{1.0\linewidth}
		\centering
		\includegraphics[width=1.0\linewidth]{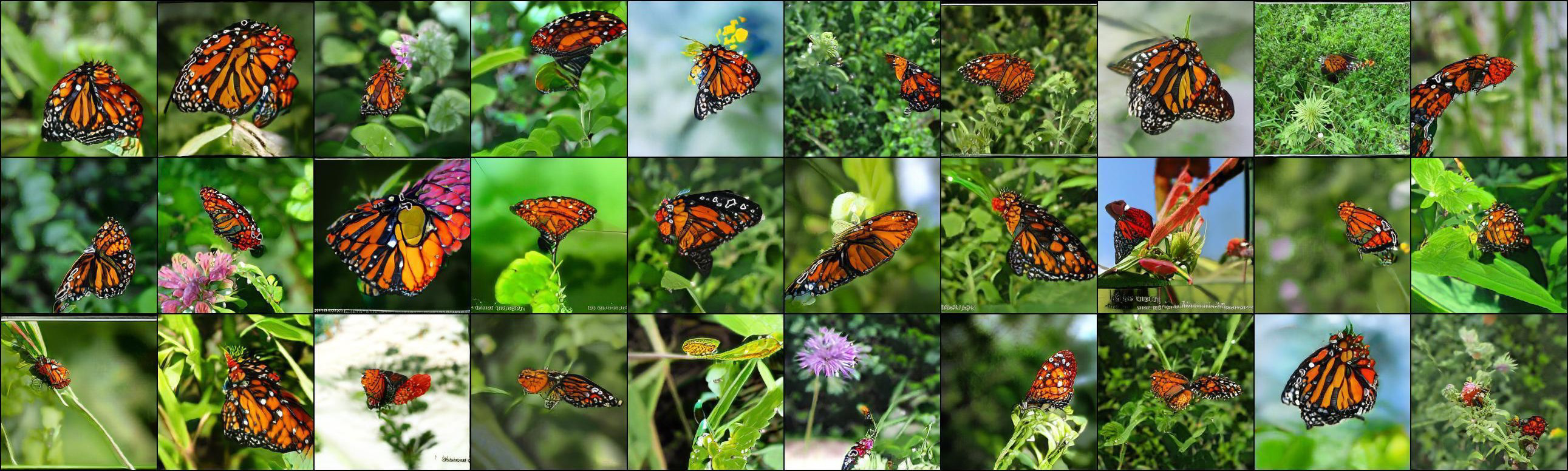}
		\caption{AM with a softmax diversity term (\ie $\lambda = 2$ in Eq.~\ref{eq:amd}).}
	\end{subfigure}
	\begin{subfigure}[b]{1.0\linewidth}
		\centering
		\includegraphics[width=1.0\linewidth]{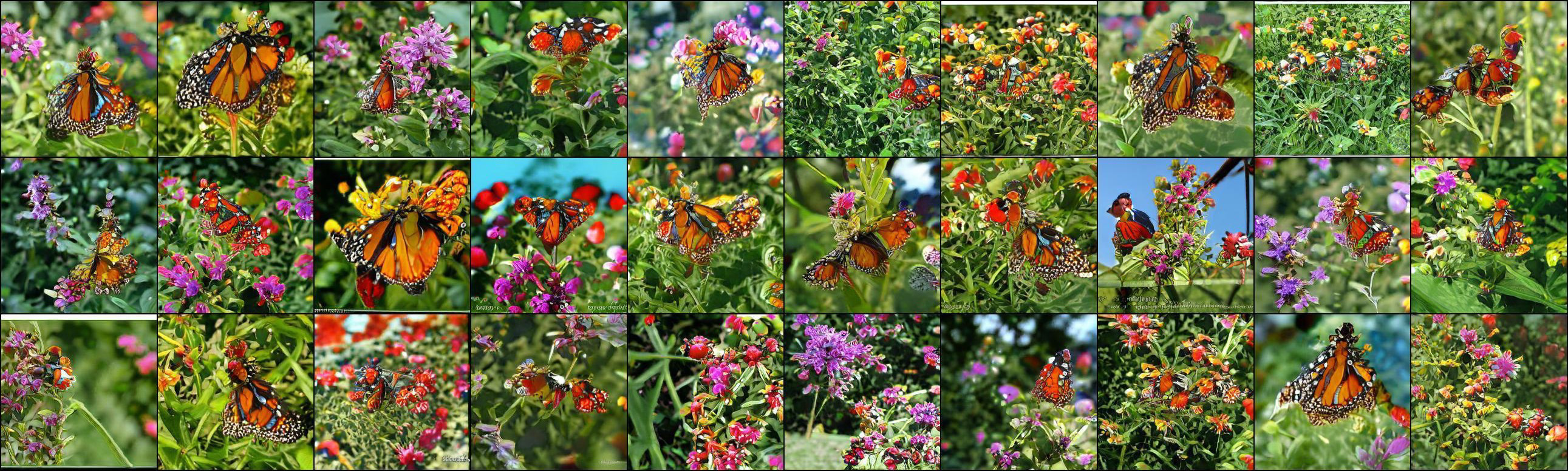}
		\caption{AM with a softmax diversity term (\ie $\lambda = 10$ in Eq.~\ref{eq:amd}).\label{fig:softmax_diversity}}
	\end{subfigure}
	\begin{subfigure}[b]{1.0\linewidth}
		\centering
		\includegraphics[width=1.0\linewidth]{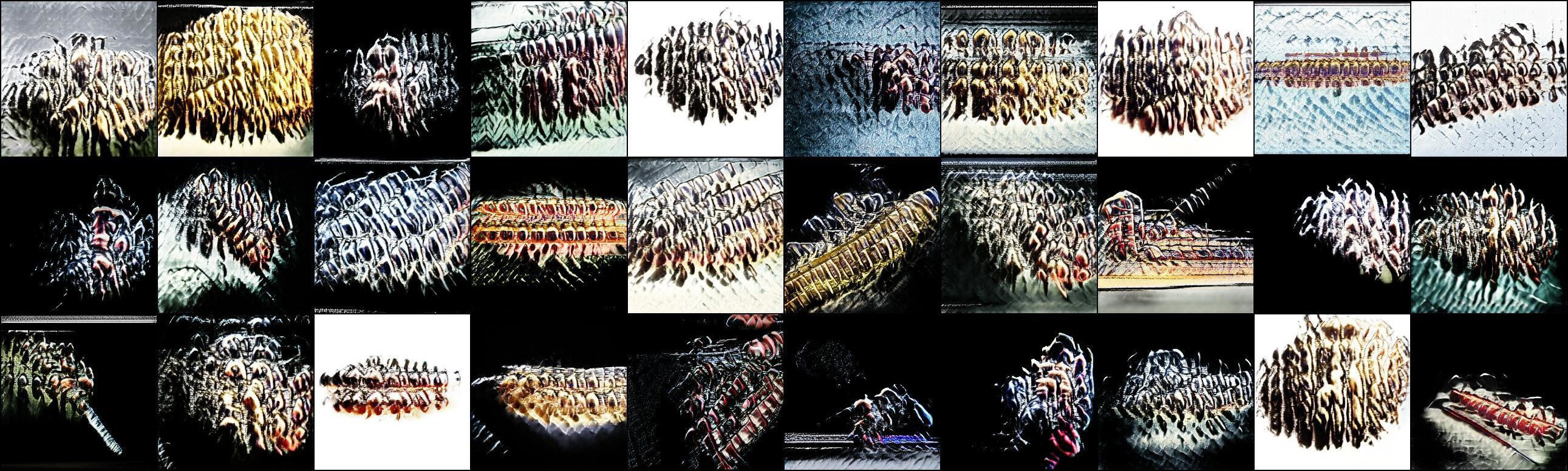}
		\caption{AM with a softmax diversity term (\ie $\lambda = 100$ in Eq.~\ref{eq:amd}).}
	\end{subfigure}
	\caption{
		The \class{monarch~butterfly} class (323) samples generated by Activation Maximization (AM) methods when increasing the multiplier $\lambda$ of a softmax probability diversity regularization term in Eq.~\ref{eq:amd}.
	}
	\label{fig:softmax_dl}
\end{figure*}

\textbf{Experiments~} In the preliminary experiments, we tested encouraging diversity in the (1) image space; (2) \layer{conv5} feature space; and (3) softmax outputs of AlexNet.
We observed that the pixel-wise regularizer can improve the diversity of background colors (Fig.~\ref{fig:pixelwise_dl}) and tends to increase the image contrast upon a high $\lambda$ multiplier (Fig.~\ref{fig:high_alpha_pixel}).
In contrast, the impact of the \layer{conv5} diversity regularizer is less noticeable (Fig.~\ref{fig:feature_dl}).
Encouraging diversity in the softmax output distribution can yield novel scenes \eg growing more flowers in \class{monarch~butterfly} images (Fig.~\ref{fig:softmax_diversity}).

While each level of diversity has its own benefits for specific applications, here, we chose to perform more tests with the softmax diversity to encourage samples to be more diverse \emph{semantically}.
That is, we re-ran the AM-S and AM-L experiments with an additional softmax diversity term (Eq.~\ref{eq:amd}) and a coefficient $\lambda = 2$ (see Fig.~\ref{fig:softmax_dl}).
We call these two AM methods with the diversity term AM-D-S and AM-D-L.

\textbf{Results~} We found that the addition of the regularizer did not improve the diversity substantially but lowered the sample quality  (Fig.~\ref{fig:biggan-256_is_fid_plots} AM-S vs. AM-D-S and AM-L vs. AM-D-L).
Similarly, the IA scores of the AM-D methods were consistently lower than those of the original AM methods (Table~\ref{table:main_result}).

\begin{table}[h]
	\centering
	\scalebox{0.8}{
		\begin{tabular}{lccccc}
			&  &  &  &  &  \\
			~~~~Method \hfill & IS (10 splits) & FID & Inception Accuracy & MS-SSIM & LPIPS \\
			& (higher=better) & (lower=better) & (higher=better) & (lower=better) & (higher=better) \\	 \hline
			
			1. ImageNet-30 (Real) & 4.18 $\pm$ 0.61 &  n/a & 0.92 & 0.42 $\pm$ 0.04 & 0.70 $\pm$ 0.08 \\ \hline
			2. BigGAN & 3.71 $\pm$ 0.74 & 31.36 & 0.91 & 0.45 $\pm$ 0.05 & 0.61 $\pm$ 0.09 \\ \hline
			
			\multicolumn{6}{l}{3. AM-L Random}   \\  
			~~~~a. AlexNet \hfill  & 5.06 $\pm$ 0.97 & 46.85 & 0.71 & 0.43 $\pm$ 0.04 & 0.66 $\pm$ 0.08  \\ 
			~~~~b. Inception-v3 \hfill  & 4.29 $\pm$ 0.56 & 31.62 & 0.87 & 0.44 $\pm$ 0.04 & 0.65 $\pm$ 0.08  \\ 
			~~~~c. ResNet-50 \hfill & 5.36 $\pm$ 0.75 & 47.23 & 0.70 & 0.44 $\pm$ 0.04 & 0.68 $\pm$ 0.09  \\ 
			~~~~d. Robust ResNet-50 \hfill & 4.59 $\pm$ 0.69 & 43.65 & 0.76 & 0.43 $\pm$ 0.05 & 0.63 $\pm$ 0.08  \\ \hline
			
			\multicolumn{6}{l}{4. AM-D-S}   \\  
			~~~~a. AlexNet \hfill  & 5.31 $\pm$ 0.60 & 48.74 & 0.69 & 0.43 $\pm$ 0.04 & 0.66 $\pm$ 0.08 \\ 
			~~~~b. Inception-v3 \hfill  & 4.23 $\pm$ 0.51 & 30.24 & 0.88 & 0.44 $\pm$ 0.04 & 0.65 $\pm$ 0.08  \\ 
			~~~~c. ResNet-50 \hfill & 5.78 $\pm$ 1.00 & 52.01 & 0.66 & 0.43 $\pm$ 0.04 & 0.68 $\pm$ 0.08  \\ 
			~~~~d. Robust ResNet-50 \hfill & 4.51 $\pm$ 0.79 & 41.74 & 0.78 & 0.44 $\pm$ 0.04 & 0.63 $\pm$ 0.09  \\ \hline
	\end{tabular}}
    \vspace{0.2cm}
	\caption{A comparison of four different classifiers (a--d) across two preliminary AM settings across 30 random classes from the ImageNet-50 low-diversity dataset (see Sec.~\ref{sec:dataset}). 
		The ImageNet-30 statistics here were computed from 30,000 images = 30 classes $\times$ 1000 images.
		%	All images in ImageNet 30 classes as the evaluation set (Row 2). 
		Similarly, for BigGAN (Row 2) and AM-L and AM-D-S methods (Row 3--4), we generated 1000 $256\times256$ samples per class.
		We computed the statistics for each initialization method from 5 trials, each with a different random seed.
		With AM-L (Sec.~\ref{sec:AM}), we maximized the log probabilities and used a large learning rate of $0.1$.
		With AM-D-S (Sec.~\ref{sec:amd}), we maximized both the log probabilities and a softmax diversity regularization term, and used a small learning rate of $0.01$.
		In sum, across both settings, AM consistently obtained the highest FID and Inception Accuracy (IA) scores with the Inception-v3 classifier (b).
		That is, it is possible to maximize the FID and IA scores when using Inception-v3 as the classifier in the AM formulation.
		However, qualitatively, we did not find the AM samples with Inception-v3 to be substantially different from the others. 
	}
	\label{tab:diff_classifiers}
\end{table}

% BigGAN-Places365 evaluation table
\begin{table}[h]
	\centering
	\scalebox{0.8}{
		\begin{tabular}{lccccc}
			&  &  &  &  &  \\
			%			\hline
			~~~~Method \hfill & IS (10 splits) & FID & ResNet-18 Accuracy & MS-SSIM & LPIPS \\
			& (higher=better) & (lower=better) & (higher=better) & (lower=better) & (higher=better) \\			
			%			\hline
			\hline
			1. Places-50 (real) & 12.17 $\pm$ 1.01 & N/A & 0.57 & 0.42 $\pm$ 0.04 & 0.70 $\pm$ 0.06 \\ 	\hline
			2. BigGAN & 8.19 $\pm$ 0.9 & 53.15 & 0.17 & 0.42 $\pm$ 0.05 & 0.66 $\pm$ 0.07 \\ 	\hline
			\multicolumn{6}{l}{3. AM-L with Mean Initialization}    \\ 	%\hline
			\hfill Trial 1 & 8.32 $\pm$ 0.89 & 42.38 & 0.51 & 0.43 $\pm$ 0.05 & 0.64 $\pm$ 0.07 \\
			\hfill Trial 2 & 8.39 $\pm$ 0.83 & 44.11 & 0.48 & 0.43 $\pm$ 0.05 & 0.64 $\pm$ 0.07 \\ 
			\hfill Trial 3 & 8.45 $\pm$ 0.84 & 42.98 & 0.46 &  0.43 $\pm$ 0.05 & 0.65 $\pm$ 0.07 \\
			\hfill Trial 4 & 7.03 $\pm$ 0.71 & 38.39 & 0.49 & 0.43 $\pm$ 0.05 & 0.64 $\pm$ 0.07 \\ 
			\hfill Trial 5 & 7.03 $\pm$ 0.71 & 38.39 & 0.49 & 0.43 $\pm$ 0.04 & 0.65 $\pm$ 0.07 \\ 	
			\hfill Average & 7.03 $\pm$ 0.51 & 41.25 & 0.49 & 0.43 $\pm$ 0.05 & 0.65 $\pm$ 0.07 \\ 	\hline
			%			\hline
			\multicolumn{6}{l}{4. AM-L with Top-5 Initialization}   \\ 	%\hline
			
			\hfill Trial 1 & 8.60 $\pm$ 0.88 & 46.92 & 0.47 & 0.43 $\pm$ 0.05 & 0.65 $\pm$ 0.07 \\
			\hfill Trial 2 & 8.45 $\pm$ 0.81 & 41.09 & 0.52 & 0.43 $\pm$ 0.05 & 0.65 $\pm$ 0.07 \\ 
			\hfill Trial 3 & 8.13 $\pm$ 0.71 & 40.35 & 0.48 & 0.43 $\pm$ 0.05 & 0.65 $\pm$ 0.07 \\
			\hfill Trial 4 & 8.20 $\pm$ 0.79 & 43.56 & 0.47 & 0.43 $\pm$ 0.05 & 0.65 $\pm$ 0.07 \\ 
			\hfill Trial 5 & 8.37 $\pm$ 0.75 & 39.49 & 0.50 & 0.43 $\pm$ 0.05 & 0.65 $\pm$ 0.07 \\ 	
			\hfill Average & 8.35 $\pm$ 0.79 & 42.28 & 0.49 & 0.43 $\pm$ 0.05 & 0.65 $\pm$ 0.07 \\ 	\hline
			%			\hline
	\end{tabular}}
	\vspace{0.2cm}
	\caption{A comparison of Places-50, BigGAN and AM images. 
		We randomly chose 50 classes in Places365 (\ie Places-50) to be the evaluation dataset for the experiments in Sec.~\ref{sec:places}. 
		The Places-50 statistics here were computed from 50,000 images = 50 classes $\times$ 1000 images that were randomly selected from the training set of Places365. 
		For BigGAN (Sec.~\ref{sec:places}), we chose the class embedding whose 10 random samples yielded the highest accuracy score for each target Places-50 class and generated 1000 samples per class. 
		With AM-L mean initialization and AM-L top-5 initialization (Sec.~\ref{sec:places}), we maximized the log probabilities and used a large learning rate of $0.1$.
		We found that samples from AM (Row 3-4) are of similar diversity but better quality than BigGAN samples.
		%		\todo{Replace Run --> Trial in Tables/Caption in Appendix}
	}
	\label{table:mit_eval}
\end{table}

% Choose the alpha for different diversity loss methods

% The BigGAN samples genreted by original embedding with diff noise 
\begin{figure*}[h!]
	\centering
	\begin{subfigure}[b]{1.0\linewidth}
		\centering
		\includegraphics[width=1.0\linewidth]{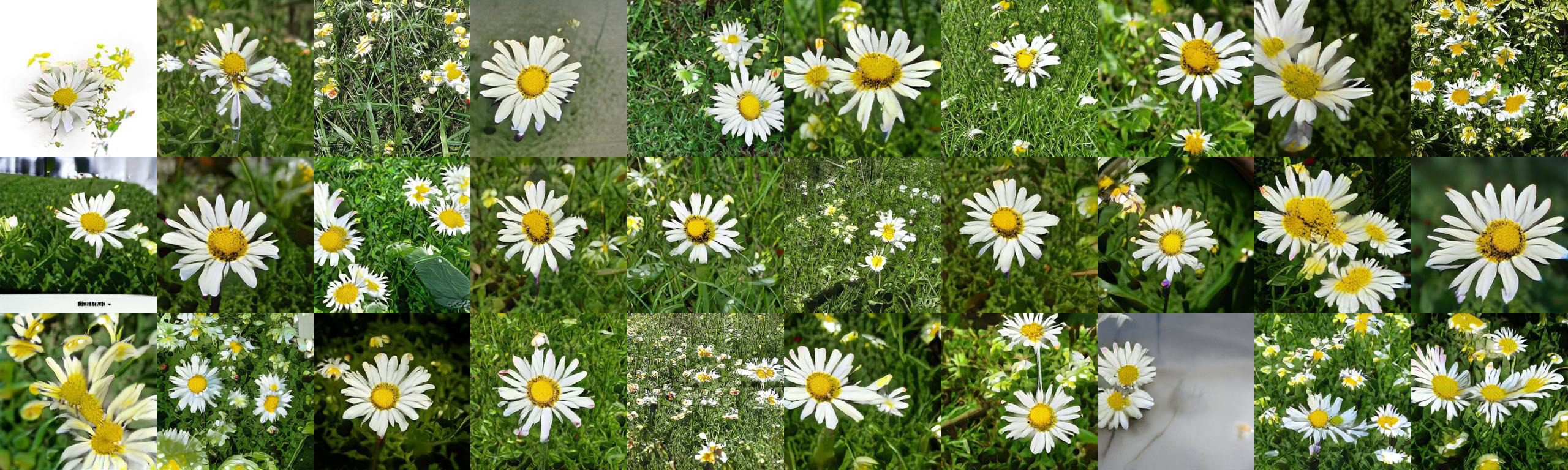}
		\caption{BigGAN samples generated with the original \class{daisy} class embedding (no noise).}
	\end{subfigure}
	\begin{subfigure}[b]{1.0\linewidth}
		\centering
		\includegraphics[width=1.0\linewidth]{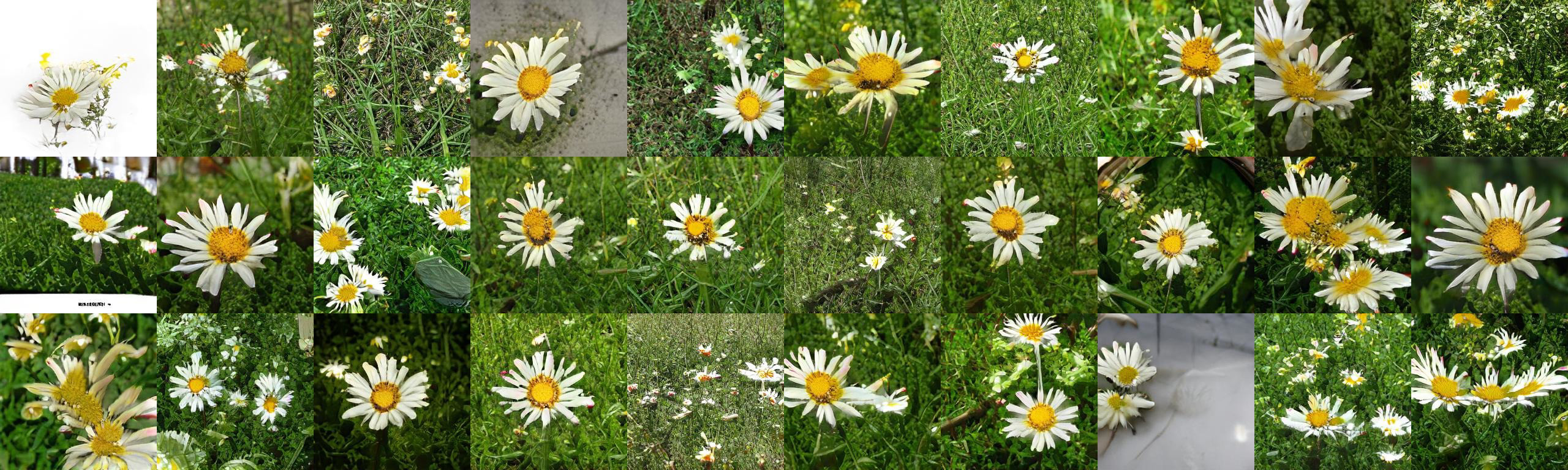}
		\caption{BigGAN samples generated with the \class{daisy} class embedding $\vc' = \vc + \epsilon$ where noise $\epsilon \sim \gN(0,0.1)$.}
	\end{subfigure}
	\begin{subfigure}[b]{1.0\linewidth}
		\centering
		\includegraphics[width=1.0\linewidth]{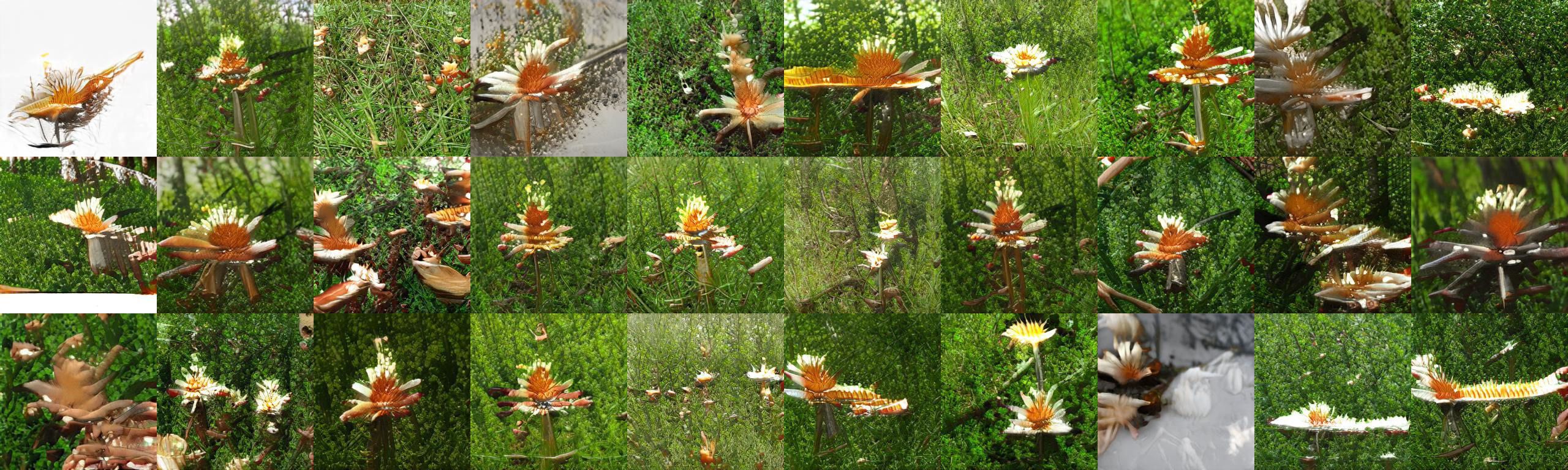}
		\caption{BigGAN samples generated with the \class{daisy} class embedding $\vc' = \vc + \epsilon$ where noise $\epsilon \sim \gN(0,0.3)$.}
	\end{subfigure}
	\begin{subfigure}[b]{1.0\linewidth}
		\centering
		\includegraphics[width=1.0\linewidth]{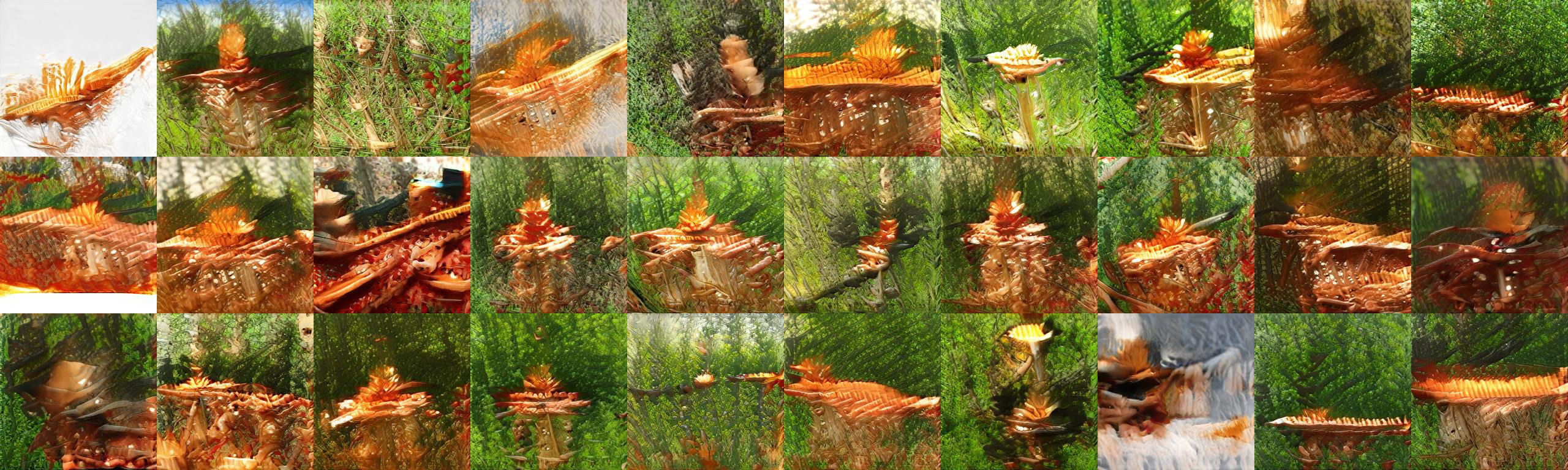}
		\caption{BigGAN samples generated with the \class{daisy} class embedding $\vc' = \vc + \epsilon$ where noise $\epsilon \sim \gN(0,0.5)$.}
	\end{subfigure}
	\caption{
		BigGAN samples when increasing the amount of noise added to the original \class{daisy} class embedding vector.
		That is, four panels (a--d) are generated using the same set of 30 latent vectors $\{\vz^i\}_{30}$ but with a different class embedding $\vc'$.
	}
	%		The samples of class 985 (\class{daisy}) generated by BigGAN which class embedding add $\gN(0,I)$ noise.}
	\label{fig:biggan_noise}
\end{figure*}

% ImageNet vs Bad BigGAN classes
\begin{figure*}[h!]
	\centering
	{	
		\begin{flushleft}
			\hspace{1.5cm} (A) ImageNet images
			\hspace{2.5cm} (B) BigGAN samples \cite{brock2019large}
		\end{flushleft}
	}
	\begin{subfigure}[b]{1.0\linewidth}
		\centering
		\includegraphics[width=1.0\linewidth]{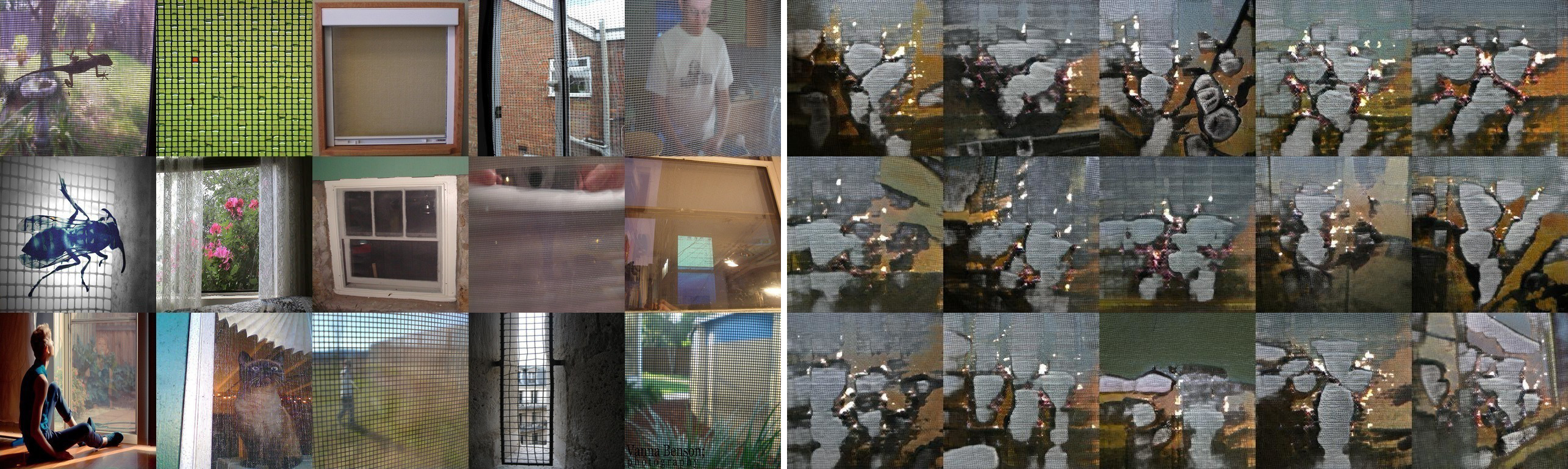}
		\caption{Samples from the \class{window~screen} class (904).}
	\end{subfigure}
	\begin{subfigure}[b]{1.0\linewidth}
		\centering
		\includegraphics[width=1.0\linewidth]{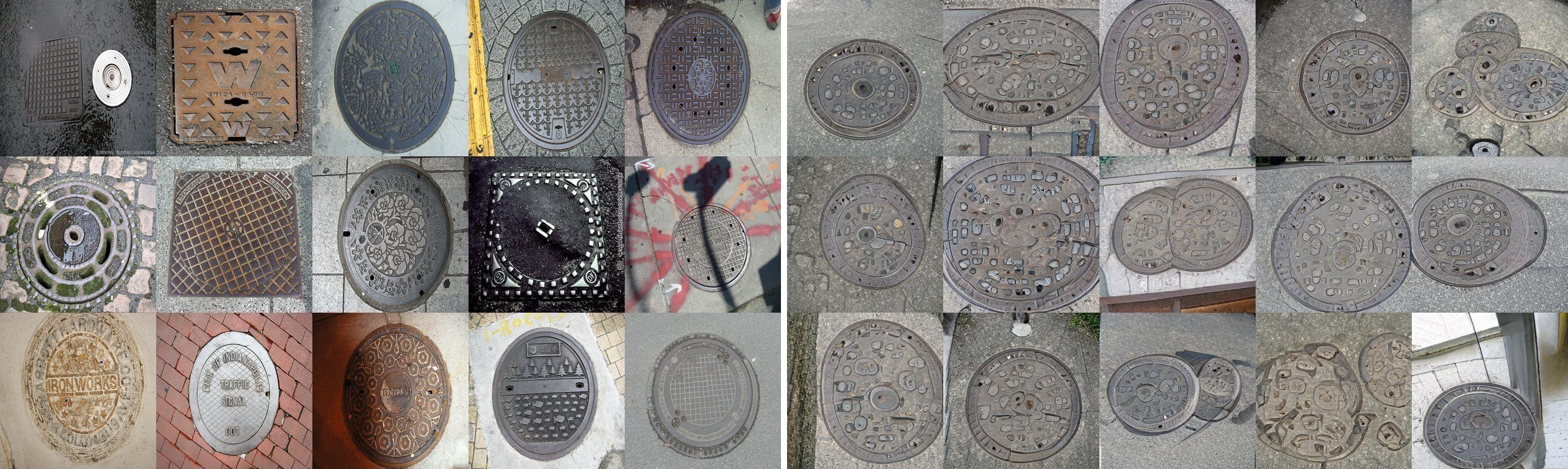}
		\caption{Samples from the \class{manhole~cover} class (640).}
	\end{subfigure}
	\begin{subfigure}[b]{1.0\linewidth}
		\centering
		\includegraphics[width=1.0\linewidth]{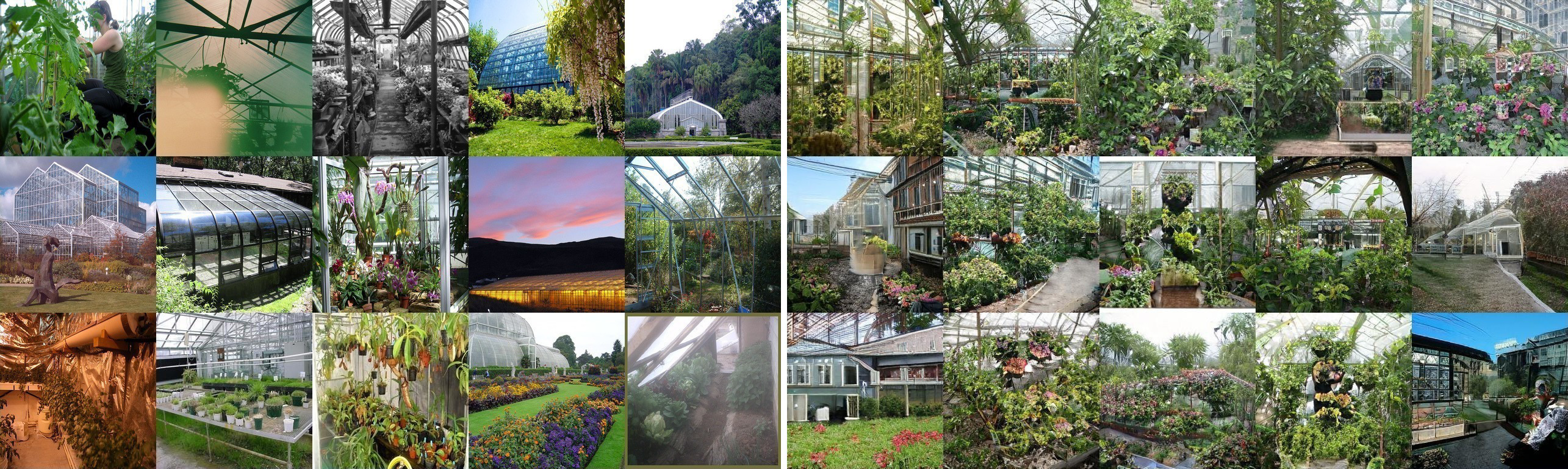}
		\caption{Samples from the \class{green house} class (580). }
	\end{subfigure}
	\begin{subfigure}[b]{1.0\linewidth}
		\centering
		\includegraphics[width=1.0\linewidth]{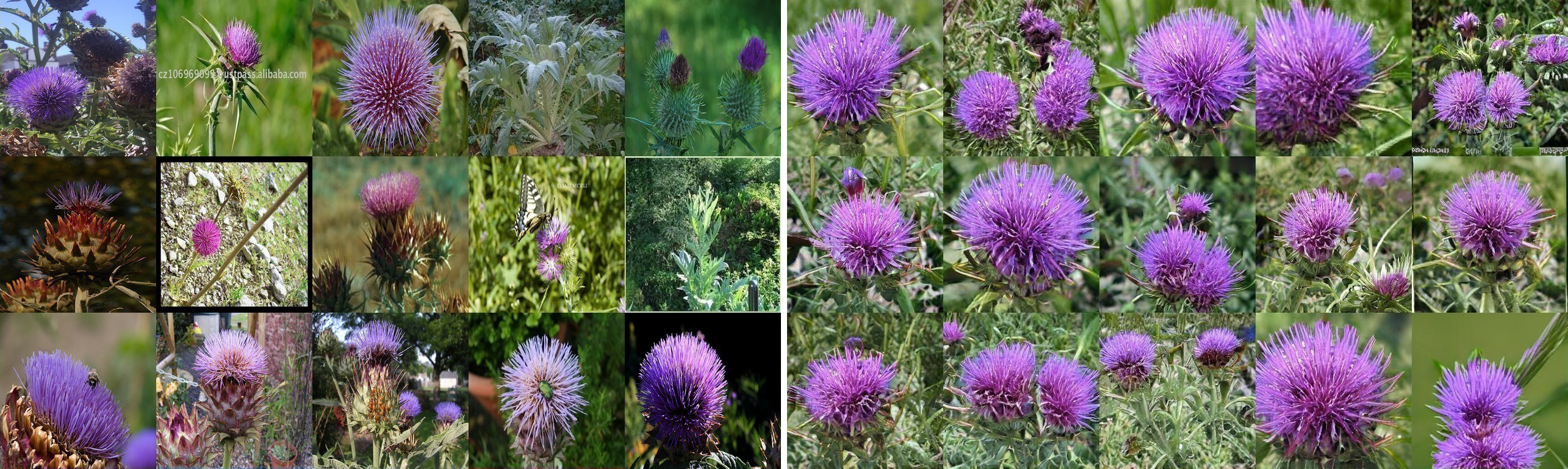}
		\caption{Samples from the \class{cardoon} class (946). }
	\end{subfigure}
	\caption{Example mode-collapse classes from the ImageNet-50 subset where BigGAN samples (right) exhibit substantially lower diversity compared to the real data (left).}
	\label{fig:imagenet_vs_bad_biggan}
\end{figure*}

% fix the snapshots of BigGAN-128
\begin{figure*}[h!]
	\centering
	\begin{subfigure}[b]{1.0\linewidth}
		\centering
		\includegraphics[width=1\linewidth]{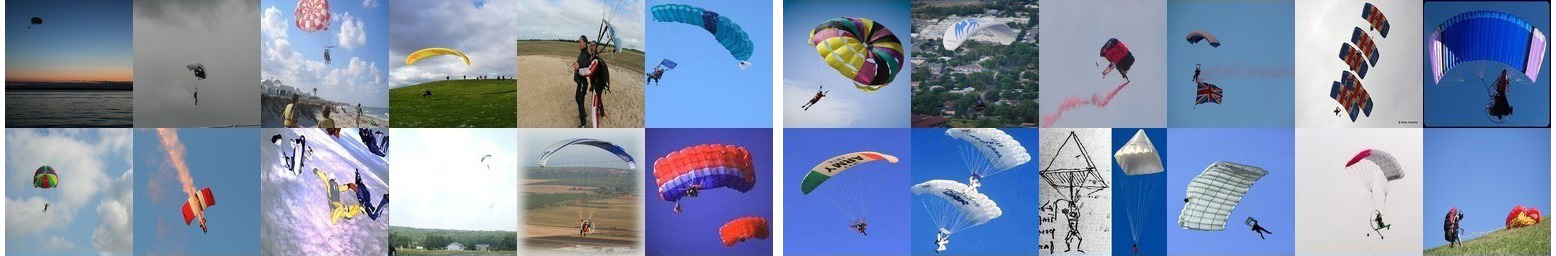}
		\caption{ImageNet samples from the \class{parachute} class.}
	\end{subfigure}
	\begin{subfigure}[b]{1.0\linewidth}
		\centering
		\includegraphics[width=1\linewidth]{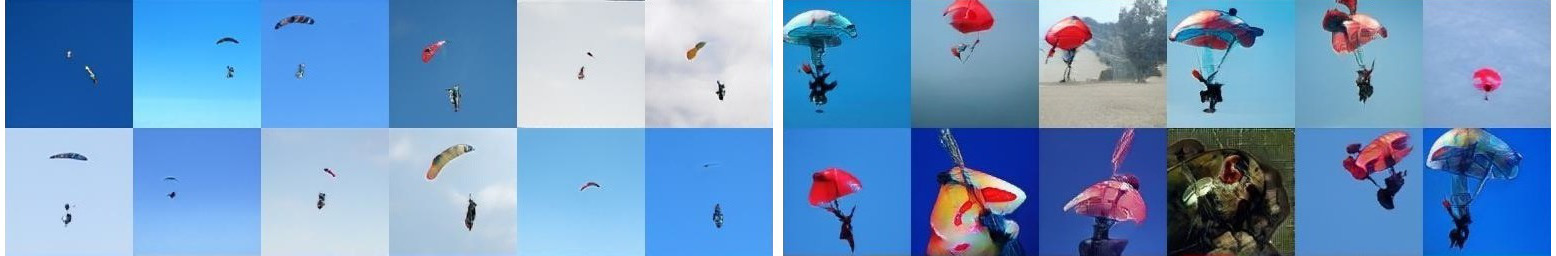}
		\caption{BigGAN samples (left) and AM samples (right), both generated using the BigGAN 138k snapshot. }
	\end{subfigure}
	\begin{subfigure}[b]{1.0\linewidth}
		\centering
		\includegraphics[width=1\linewidth]{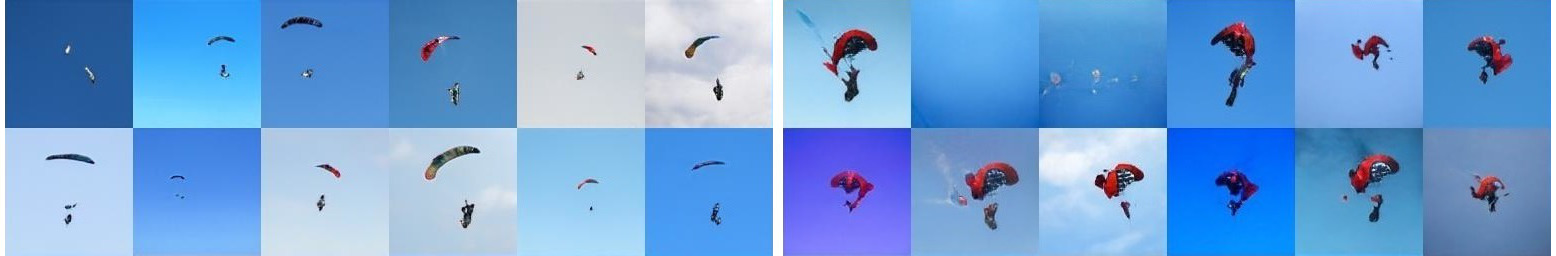}
		%		\caption{Samples from BigGAN (left) and samples from AM (right) correspond to the snapshot (140k). }
		\caption{BigGAN samples (left) and AM samples (right), both generated using the BigGAN 140k snapshot. }
	\end{subfigure}
	\begin{subfigure}[b]{1.0\linewidth}
		\centering
		\includegraphics[width=1\linewidth]{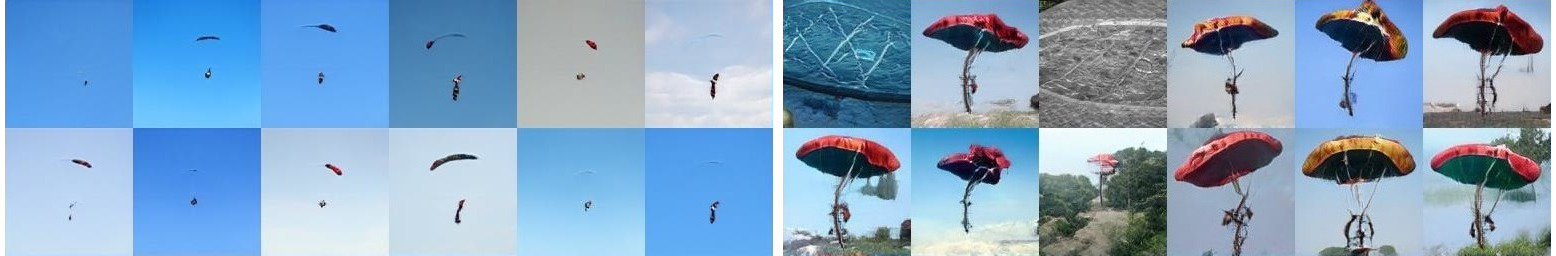}
		%		\caption{Samples from BigGAN (left) and samples from AM (right) correspond to the snapshot (142k).  }
		\caption{BigGAN samples (left) and AM samples (right), both generated using the BigGAN 142k snapshot. }
	\end{subfigure}
	\begin{subfigure}[b]{1.0\linewidth}
		\centering
		\includegraphics[width=1\linewidth]{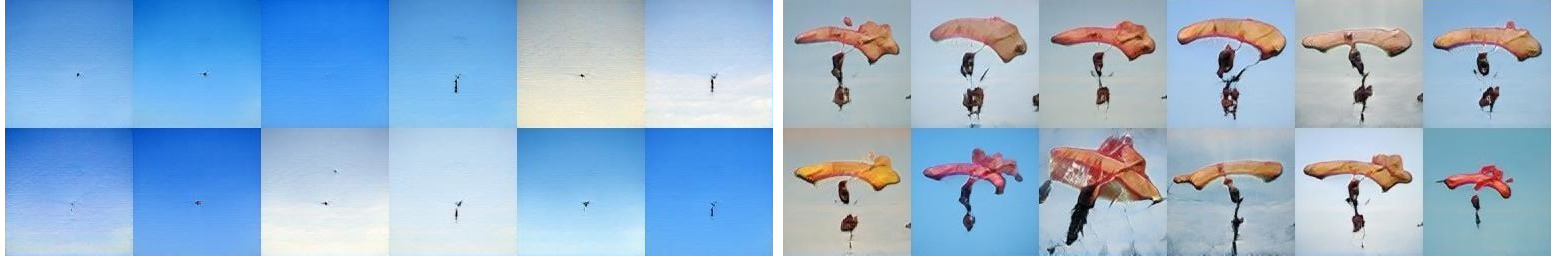}
		%		\caption{Samples from BigGAN (left) and samples from AM (right) correspond to the snapshot (144k).  }
		\caption{BigGAN samples (left) and AM samples (right), both generated using the BigGAN 144k snapshot. }
	\end{subfigure}
	\begin{subfigure}[b]{1.0\linewidth}
		\centering
		\includegraphics[width=1\linewidth]{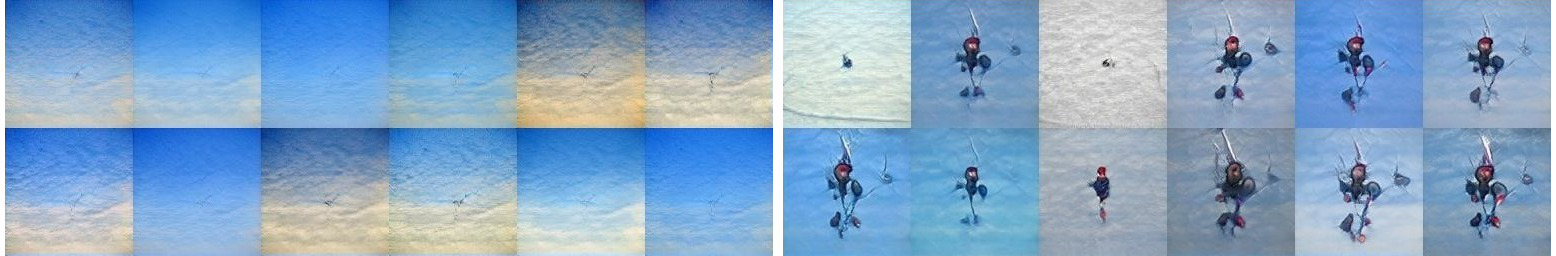}
		%		\caption{Samples from BigGAN (left) and samples from AM (right) correspond to the snapshot (146k). }
		\caption{BigGAN samples (left) and AM samples (right), both generated using the BigGAN 146k snapshot. }
	\end{subfigure}
	\caption{
		Applying our AM method to 5 different $128\times128$ BigGAN training snapshots (b--f) yielded samples (right) that qualitatively are more diverse and recognizable to be from the \class{parachute} class compared to the original BigGAN samples (left).
		While the original BigGAN samples are almost showing only the blue sky (d--f), AM samples show large and colorful parachutes.
		%		AM examples from mode-collapse \class{parachute} class (701) where BigGAN samples (left) exhibit substantially lower diversity compared to the AM (right).
	}
	\label{fig:fix_biggan_128_snapshot_701}
\end{figure*}

\begin{figure*}[h!]
	\centering
	\begin{subfigure}[b]{1.0\linewidth}
		\centering
		\includegraphics[width=1\linewidth]{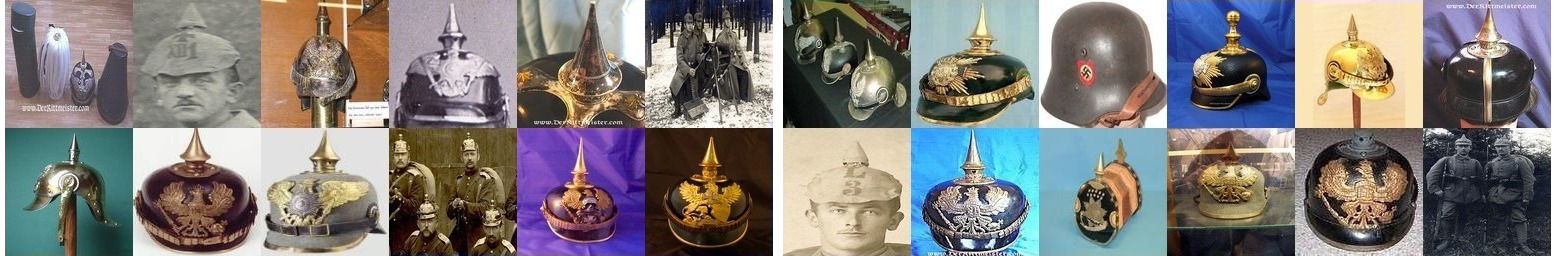}
		%		\caption{Samples from ImageNet.}
		\caption{ImageNet samples from the \class{pickelhaube} class.}
	\end{subfigure}
	\begin{subfigure}[b]{1.0\linewidth}
		\centering
		\includegraphics[width=1\linewidth]{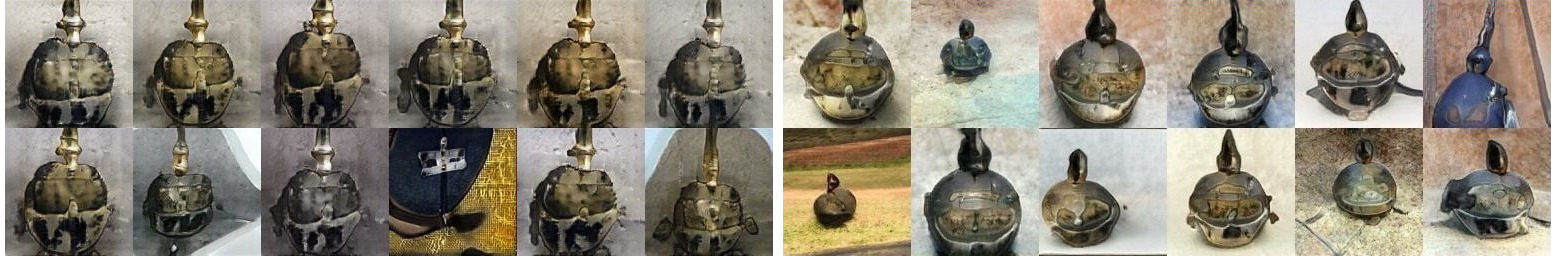}
		%		\caption{Samples from BigGAN (left) and samples from AM (right) correspond to the snapshot (138k).}
		\caption{BigGAN samples (left) and AM samples (right), both generated using the BigGAN 138k snapshot.}
	\end{subfigure}
	\begin{subfigure}[b]{1.0\linewidth}
		\centering
		\includegraphics[width=1\linewidth]{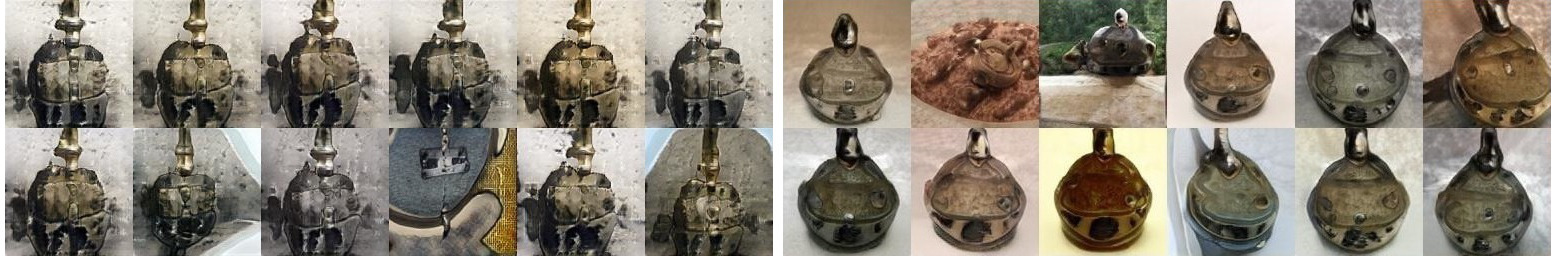}
		%		\caption{Samples from BigGAN (left) and samples from AM (right) correspond to the snapshot (140k).}
		\caption{BigGAN samples (left) and AM samples (right), both generated using the BigGAN 140k snapshot. }
	\end{subfigure}
	\begin{subfigure}[b]{1.0\linewidth}
		\centering
		\includegraphics[width=1\linewidth]{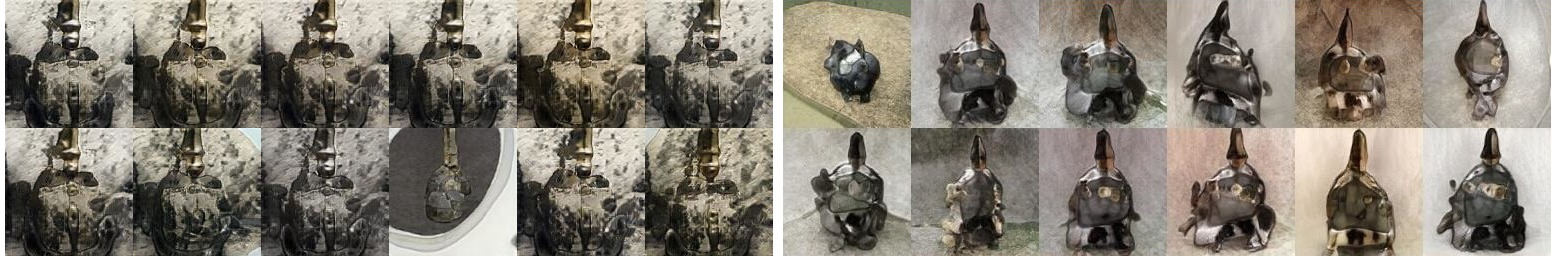}
		%		\caption{Samples from BigGAN (left) and samples from AM (right) correspond to the snapshot (142k).}
		\caption{BigGAN samples (left) and AM samples (right), both generated using the BigGAN 142k snapshot. }
	\end{subfigure}
	\begin{subfigure}[b]{1.0\linewidth}
		\centering
		\includegraphics[width=1\linewidth]{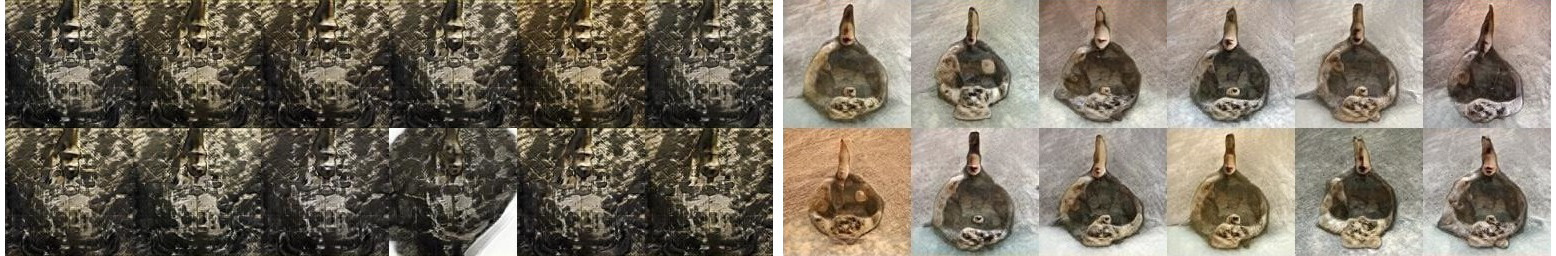}
		%		\caption{Samples from BigGAN (left) and samples from AM (right) correspond to the snapshot (144k).}
		\caption{BigGAN samples (left) and AM samples (right), both generated using the BigGAN 144k snapshot. }
	\end{subfigure}
	\begin{subfigure}[b]{1.0\linewidth}
		\centering
		\includegraphics[width=1\linewidth]{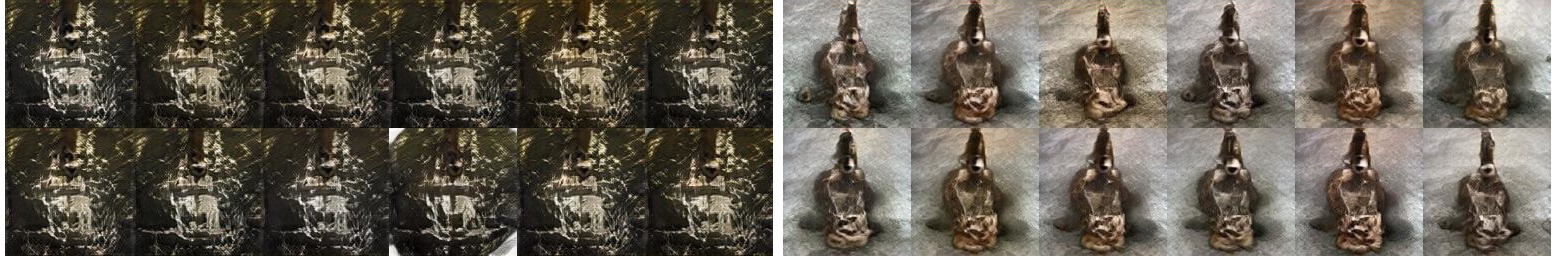}
		%		\caption{Samples from BigGAN (left) and samples from AM (right) correspond to the snapshot (146k).}
		\caption{BigGAN samples (left) and AM samples (right), both generated using the BigGAN 146k snapshot. }
	\end{subfigure}
	\caption{
		%		AM examples from mode-collapse \class{pickelhaube} class (715) where BigGAN samples (left) exhibit substantially lower diversity compared to the AM (right).
		The same figure as Fig.~\ref{fig:fix_biggan_128_snapshot_701} but for the \class{pickelhaube} class (715).
	}
	\label{fig:fix_biggan_128_snapshot_715}
\end{figure*}

\begin{figure*}[h!]
	\centering
	\begin{subfigure}[b]{1.0\linewidth}
		\centering
		\includegraphics[width=1\linewidth]{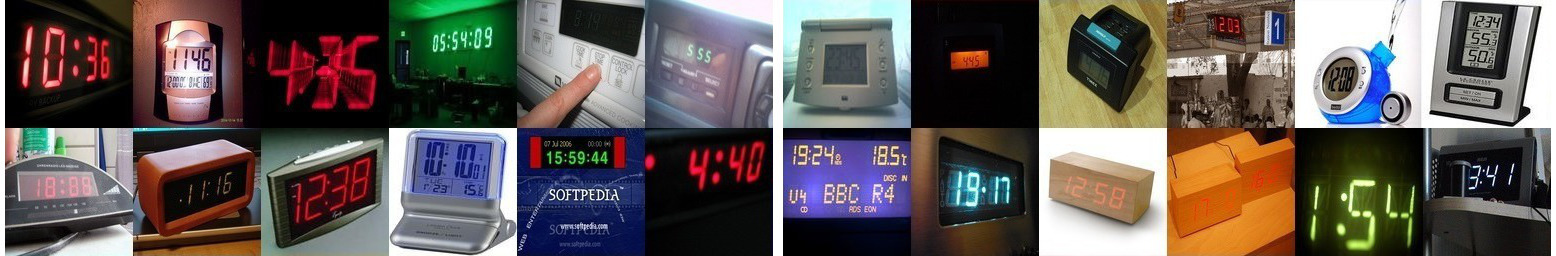}
		%		\caption{Samples from ImageNet. }
		\caption{ImageNet samples from the \class{digital~clock} class.}
	\end{subfigure}
	\begin{subfigure}[b]{1.0\linewidth}
		\centering
		\includegraphics[width=1\linewidth]{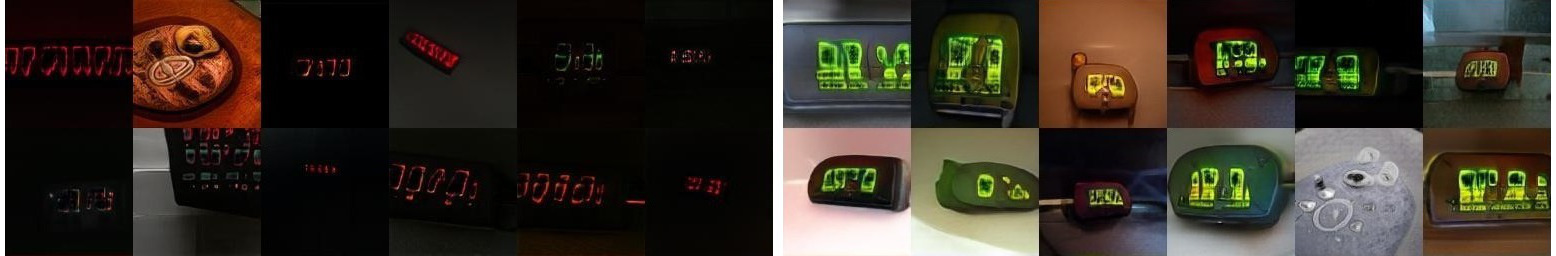}
		%		\caption{Samples from BigGAN (left) and samples from AM (right) correspond to the snapshot (138k).}
		\caption{BigGAN samples (left) and AM samples (right), both generated using the BigGAN 138k snapshot. }
	\end{subfigure}
	\begin{subfigure}[b]{1.0\linewidth}
		\centering
		\includegraphics[width=1\linewidth]{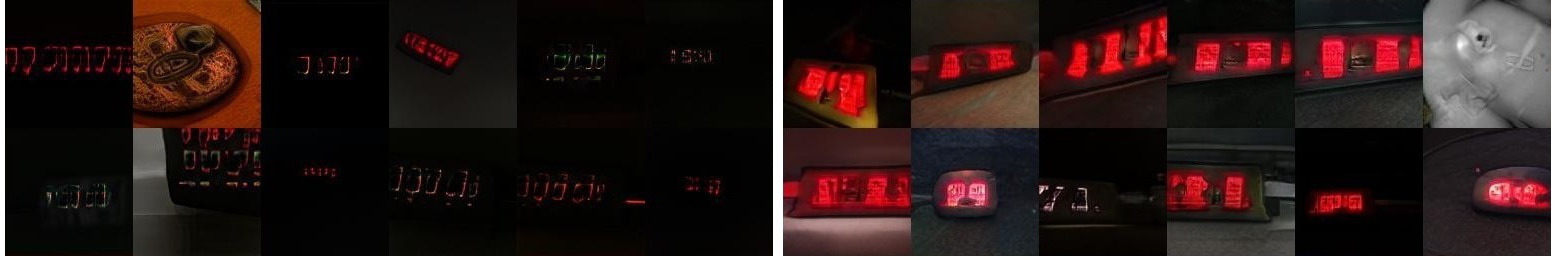}
		%		\caption{Samples from BigGAN (left) and samples from AM (right) correspond to the snapshot (140k).}
		\caption{BigGAN samples (left) and AM samples (right), both generated using the BigGAN 140k snapshot. }
	\end{subfigure}
	\begin{subfigure}[b]{1.0\linewidth}
		\centering
		\includegraphics[width=1\linewidth]{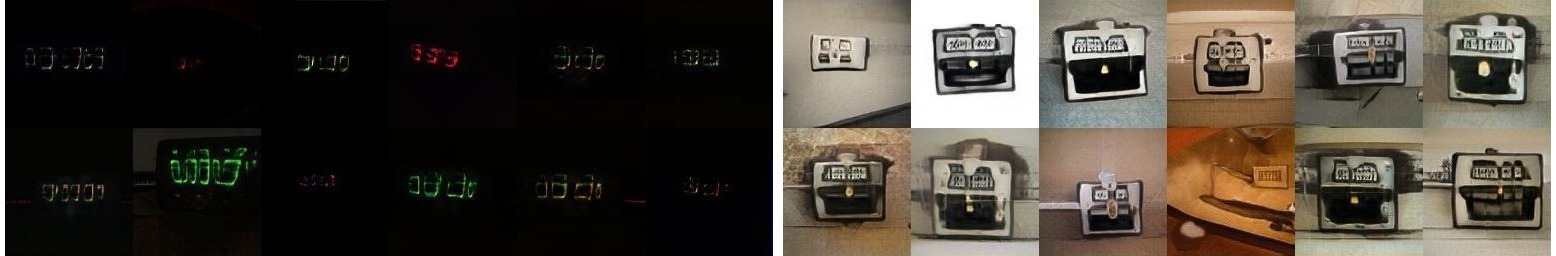}
		%		\caption{Samples from BigGAN (left) and samples from AM (right) correspond to the snapshot (142k).}
		\caption{BigGAN samples (left) and AM samples (right), both generated using the BigGAN 142k snapshot. }
	\end{subfigure}
	\begin{subfigure}[b]{1.0\linewidth}
		\centering
		\includegraphics[width=1\linewidth]{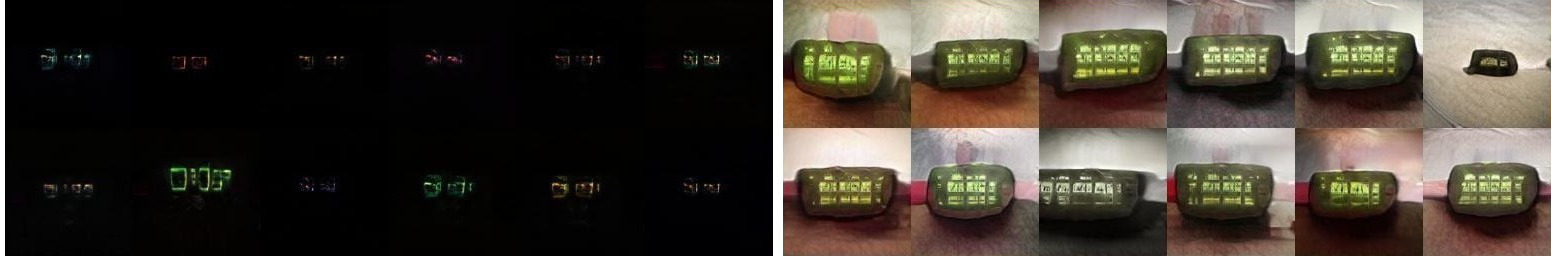}
		%		\caption{Samples from BigGAN (left) and samples from AM (right) correspond to the snapshot (144k).}
		\caption{BigGAN samples (left) and AM samples (right), both generated using the BigGAN 144k snapshot. }
	\end{subfigure}
	\begin{subfigure}[b]{1.0\linewidth}
		\centering
		\includegraphics[width=1\linewidth]{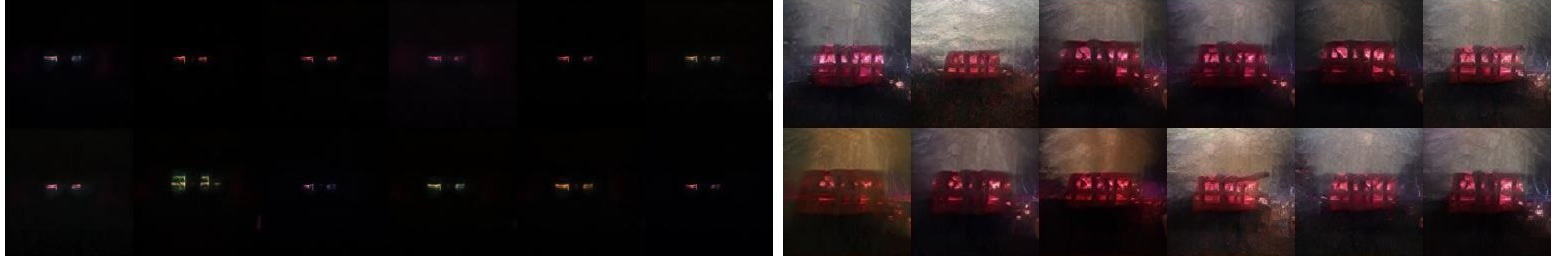}
		%		\caption{Samples from BigGAN (left) and samples from AM (right) correspond to the snapshot (146k). }
		\caption{BigGAN samples (left) and AM samples (right), both generated using the BigGAN 146k snapshot. }
	\end{subfigure}
	%	\caption{AM examples from mode-collapse \class{digital~clock} class (530) where BigGAN samples (left) exhibit substantially lower diversity compared to the AM (right).}
	\caption{The same figure as Fig.~\ref{fig:fix_biggan_128_snapshot_701} but for the \class{digital~clock} class (530).}
	\label{fig:fix_biggan_128_snapshot_530}
\end{figure*}

% BigGAN-AM 256 final results
\begin{figure*}
	\centering
	{	
		\begin{flushleft}
			\hspace{1.6cm} (A) ImageNet
			\hspace{1.1cm} (B) BigGAN \cite{brock2019large}
			\hspace{1.1cm} (C) AM (ours)
		\end{flushleft}
	}
	\begin{subfigure}[b]{1.0\linewidth}
		\centering
		\includegraphics[width=0.9\linewidth]{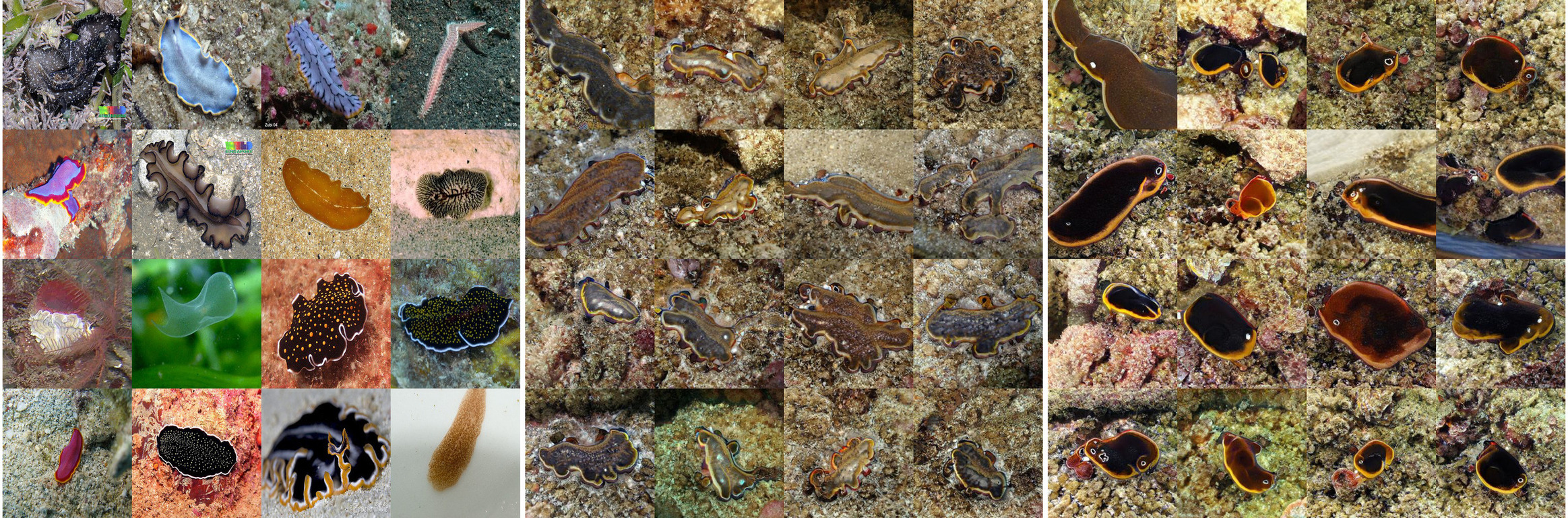}
		\caption{Samples from the \class{flatworm} class (110).}
	\end{subfigure}
	\begin{subfigure}[b]{1.0\linewidth}
		\centering
		\includegraphics[width=0.9\linewidth]{images/256_imagenet_biggan_biggan-am_111_final_show.jpg}
		\caption{Samples from the \class{nematode} class (111).}
	\end{subfigure}
	\begin{subfigure}[b]{1.0\linewidth}
		\centering
		\includegraphics[width=0.9\linewidth]{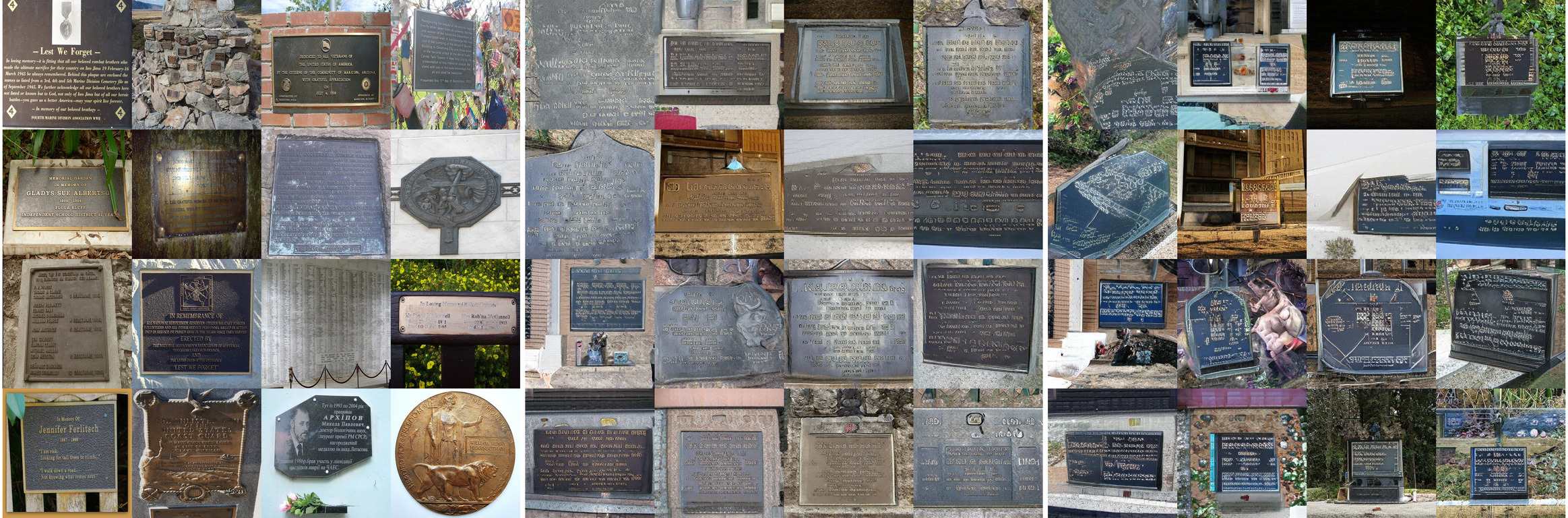}
		\caption{Samples from the \class{brass} class (458).}		
	\end{subfigure}
	\begin{subfigure}[b]{1.0\linewidth}
		\centering
		\includegraphics[width=0.9\linewidth]{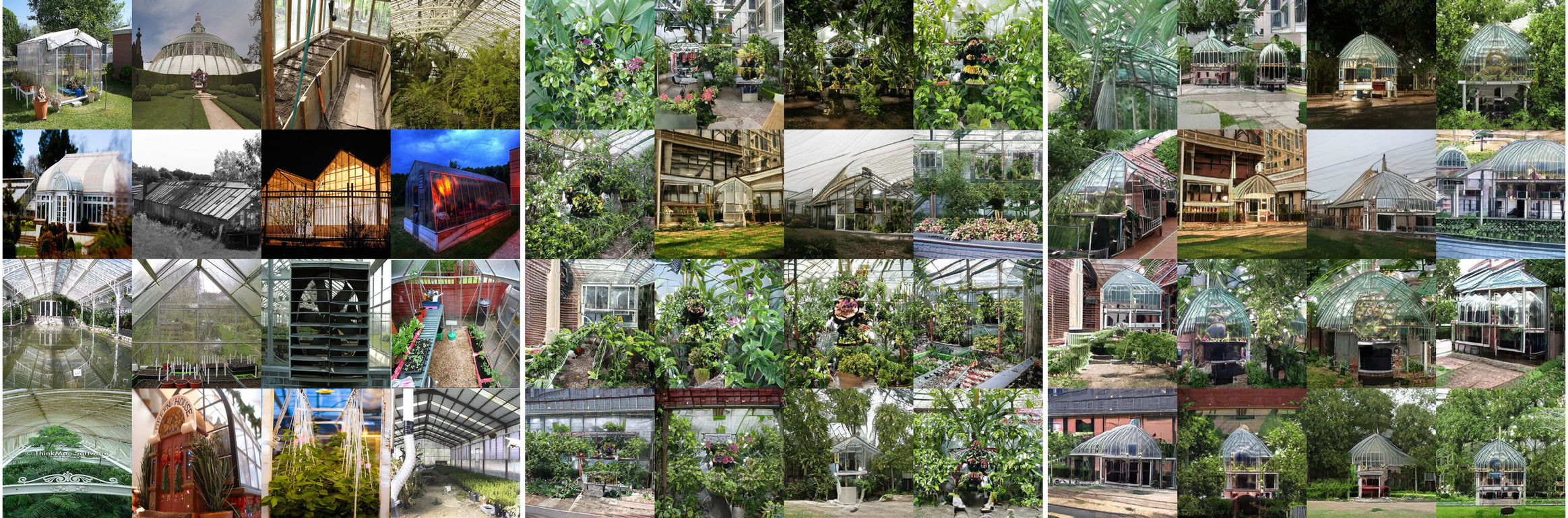}
		\caption{Samples from the \class{greenhouse} class (580).}				
	\end{subfigure}
	\caption{A comparison between the $256\times256$ samples from the ImageNet training set (A), the original BigGAN model (B), and our AM method (C) for four ImageNet-50 low-diversity classes.
		AM samples (C) are of similar quality but higher diversity than the original BigGAN samples (B). See \url{https://drive.google.com/drive/folders/14qiLdaslnxfsCMnlBa4n1iEO1EUUYUjQ?usp=sharing} for the high-resolution version of this figure.
	}
	\label{fig:biggan-am_final_01}
\end{figure*}

\begin{figure*}
	\centering
	{	
		\begin{flushleft}
			\hspace{1.6cm} (A) ImageNet
			\hspace{1.1cm} (B) BigGAN \cite{brock2019large}
			\hspace{1.1cm} (C) AM (ours)
		\end{flushleft}
	}
	\begin{subfigure}[b]{1.0\linewidth}
		\centering
		\includegraphics[width=0.9\linewidth]{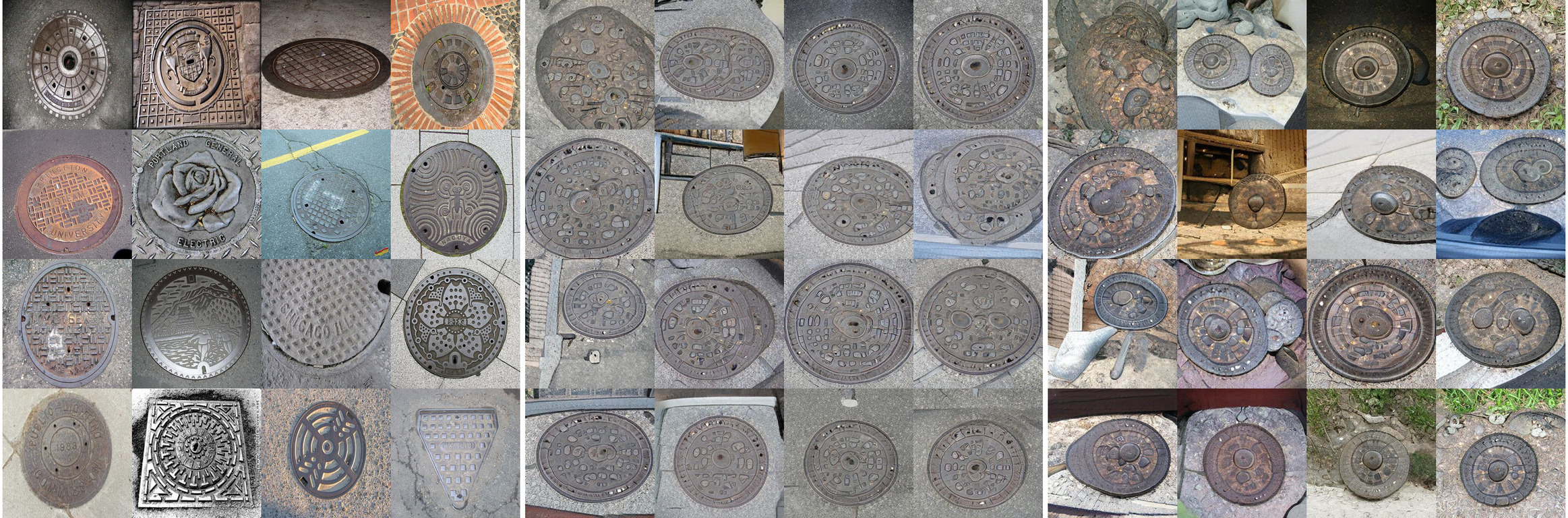}
		%		\caption{Class 640 (\class{manhole~cover}) }
		\caption{Samples from the \class{manhole~cover} class (640).}				
	\end{subfigure}
	\begin{subfigure}[b]{1.0\linewidth}
		\centering
		\includegraphics[width=0.9\linewidth]{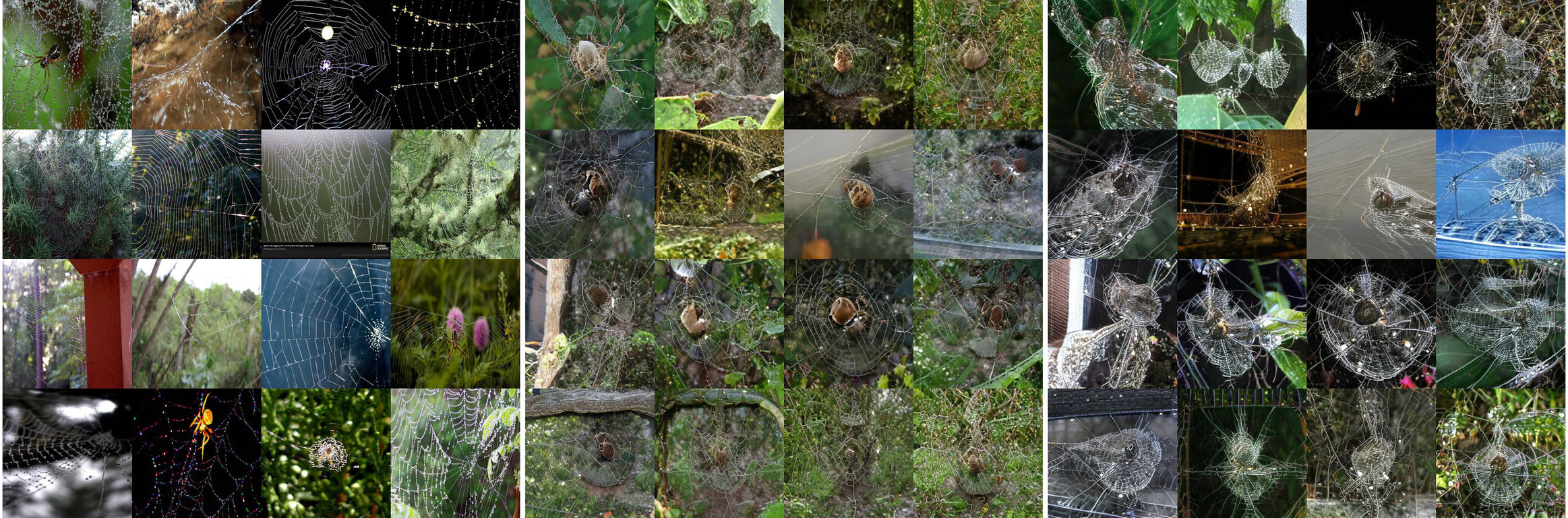}
		%		\caption{Class 815 (\class{spider~web})}
		\caption{Samples from the \class{spider~web} class (815).}
	\end{subfigure}
	\begin{subfigure}[b]{1.0\linewidth}
		\centering
		\includegraphics[width=0.9\linewidth]{images/256_imagenet_biggan_biggan-am_904_final_show.jpg}
		%		\caption{Class 904 (\class{window~screen}) }
		\caption{Samples from the \class{window~screen} class (904).\label{fig:window_screen}}		
	\end{subfigure}
	\begin{subfigure}[b]{1.0\linewidth}
		\centering
		\includegraphics[width=0.9\linewidth]{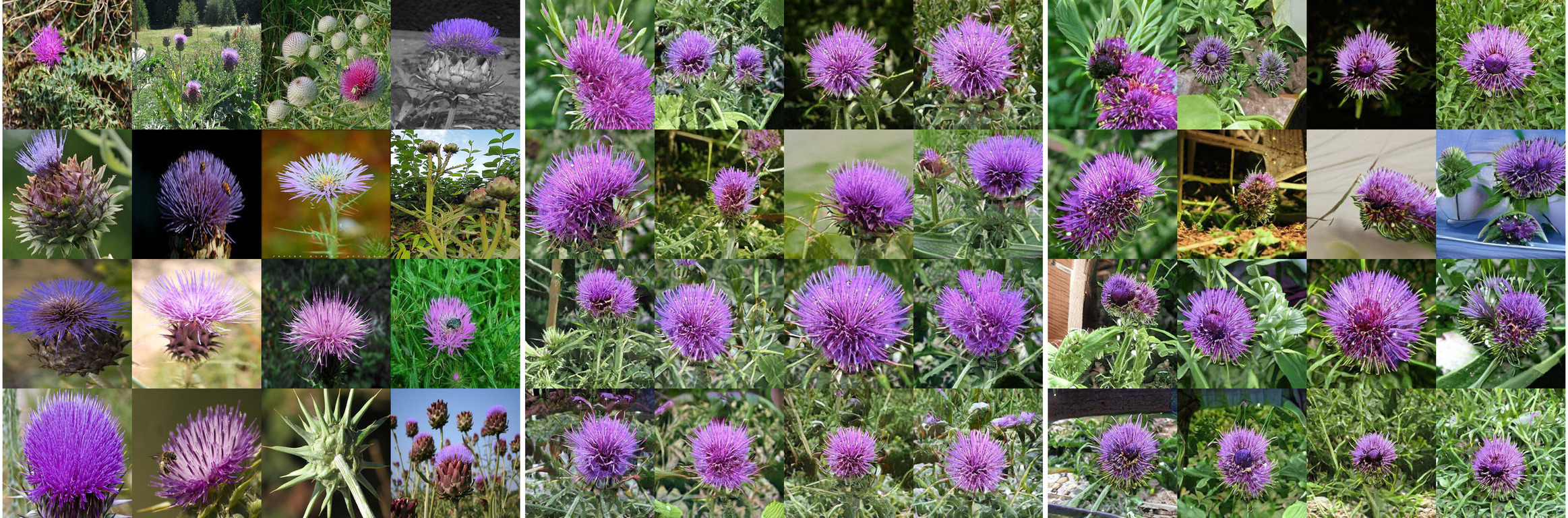}
		%		\caption{Class 946 (\class{cardoon}) }
		\caption{Samples from the \class{cardoon} class (946).}
	\end{subfigure}
	\caption{A comparison between the $256\times256$ samples from the ImageNet training set (A), the original BigGAN model (B), and our AM method (C) for four ImageNet-50 low-diversity classes.
		AM samples (C) are of similar quality but higher diversity than the original BigGAN samples (B). See \url{https://drive.google.com/drive/folders/14qiLdaslnxfsCMnlBa4n1iEO1EUUYUjQ?usp=sharing} for the high-resolution version of this figure.
	}
	\label{fig:biggan-am_final_02}
\end{figure*}

\begin{figure*}
	\centering
	{	
		\begin{flushleft}
			\hspace{1.6cm} (A) ImageNet
			\hspace{1.1cm} (B) BigGAN \cite{brock2019large}
			\hspace{1.1cm} (C) AM (ours)
		\end{flushleft}
	}
	\begin{subfigure}[b]{1.0\linewidth}
		\centering
		\includegraphics[width=0.9\linewidth]{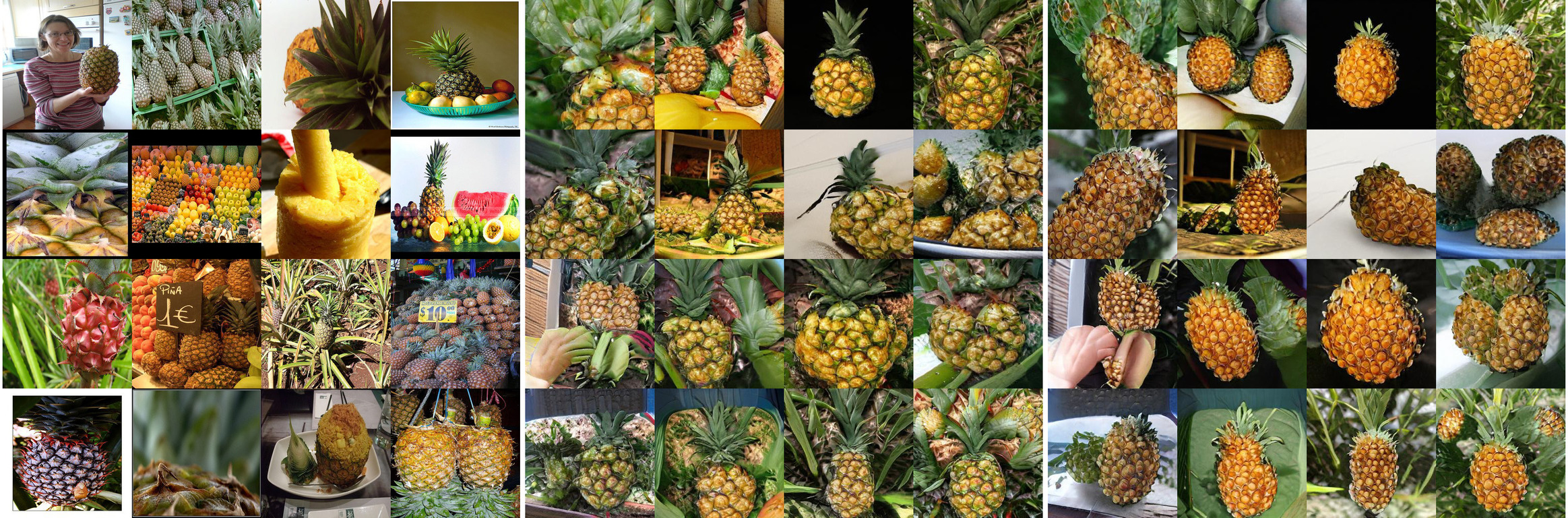}
		\caption{Samples from the \class{pineapple} class (953).}
	\end{subfigure}
	\begin{subfigure}[b]{1.0\linewidth}
		\centering
		\includegraphics[width=0.9\linewidth]{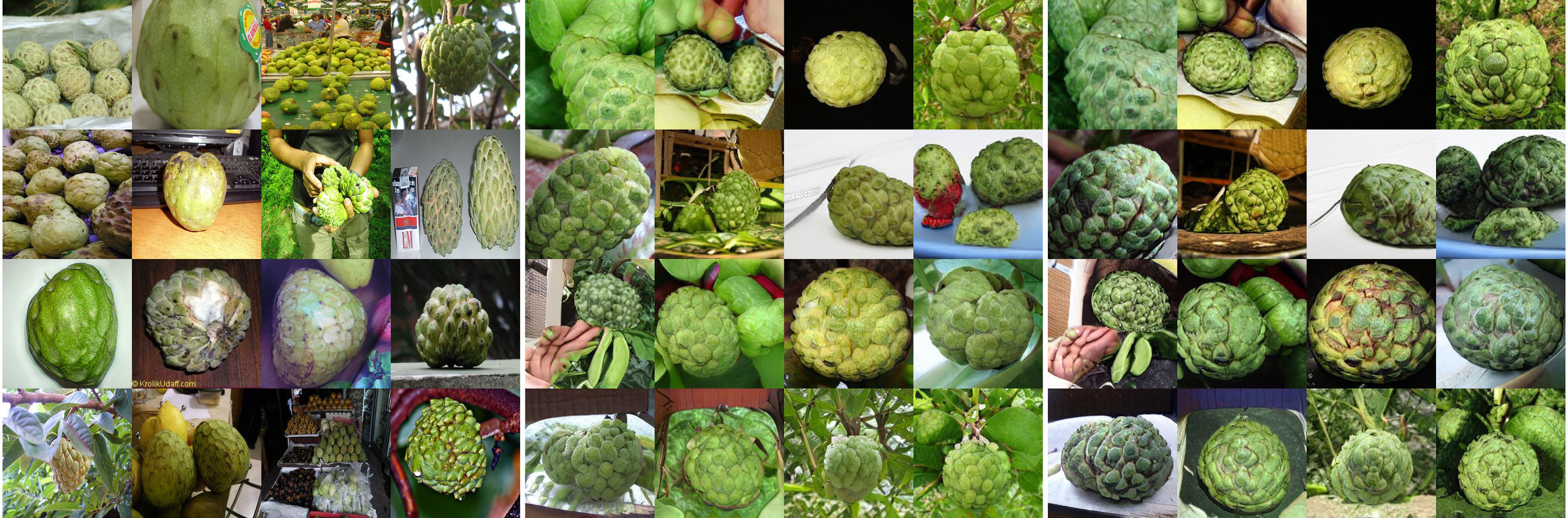}
		\caption{Samples from the \class{custard~apple} class (956).}
	\end{subfigure}
	\begin{subfigure}[b]{1.0\linewidth}
		\centering
		\includegraphics[width=0.9\linewidth]{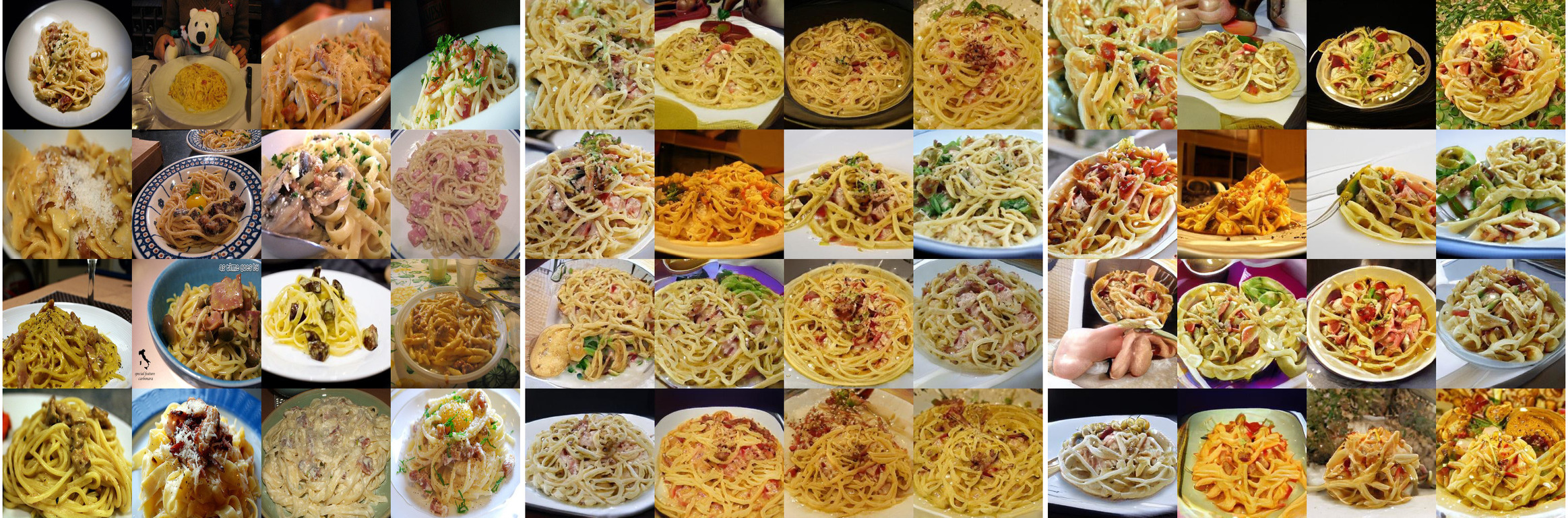}
		\caption{Samples from the \class{carbonara} class (959).}
	\end{subfigure}
	\begin{subfigure}[b]{1.0\linewidth}
		\centering
		\includegraphics[width=0.9\linewidth]{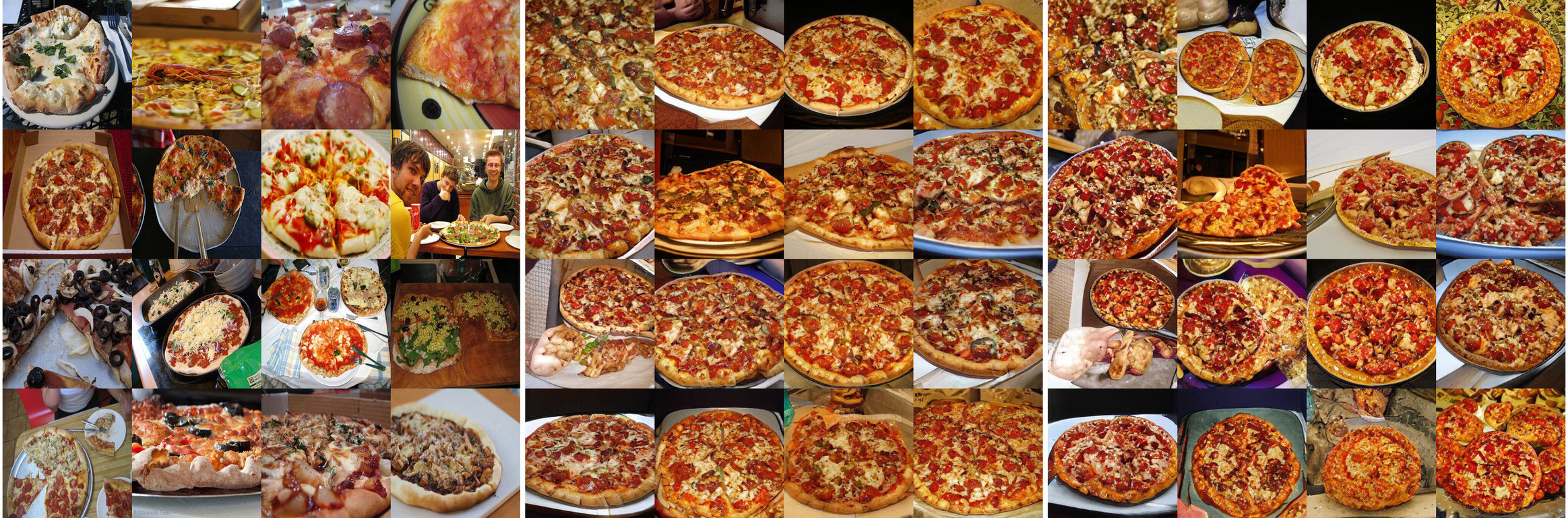}
		\caption{Samples from the \class{pizza} class (963).}
	\end{subfigure}
	\caption{A comparison between the $256\times256$ samples from the ImageNet training set (A), the original BigGAN model (B), and our AM method (C) for four ImageNet-50 low-diversity classes.
		AM samples (C) are of similar quality but higher diversity than the original BigGAN samples (B). See \url{https://drive.google.com/drive/folders/14qiLdaslnxfsCMnlBa4n1iEO1EUUYUjQ?usp=sharing} for the high-resolution version of this figure.
	}
	\label{fig:biggan-am_final_03}
\end{figure*}

% BigGAN vs AM bigger comparisons
\begin{figure*}
	\centering
\
	\begin{subfigure}[b]{1.0\linewidth}
		\centering
		\includegraphics[width=1.0\linewidth]{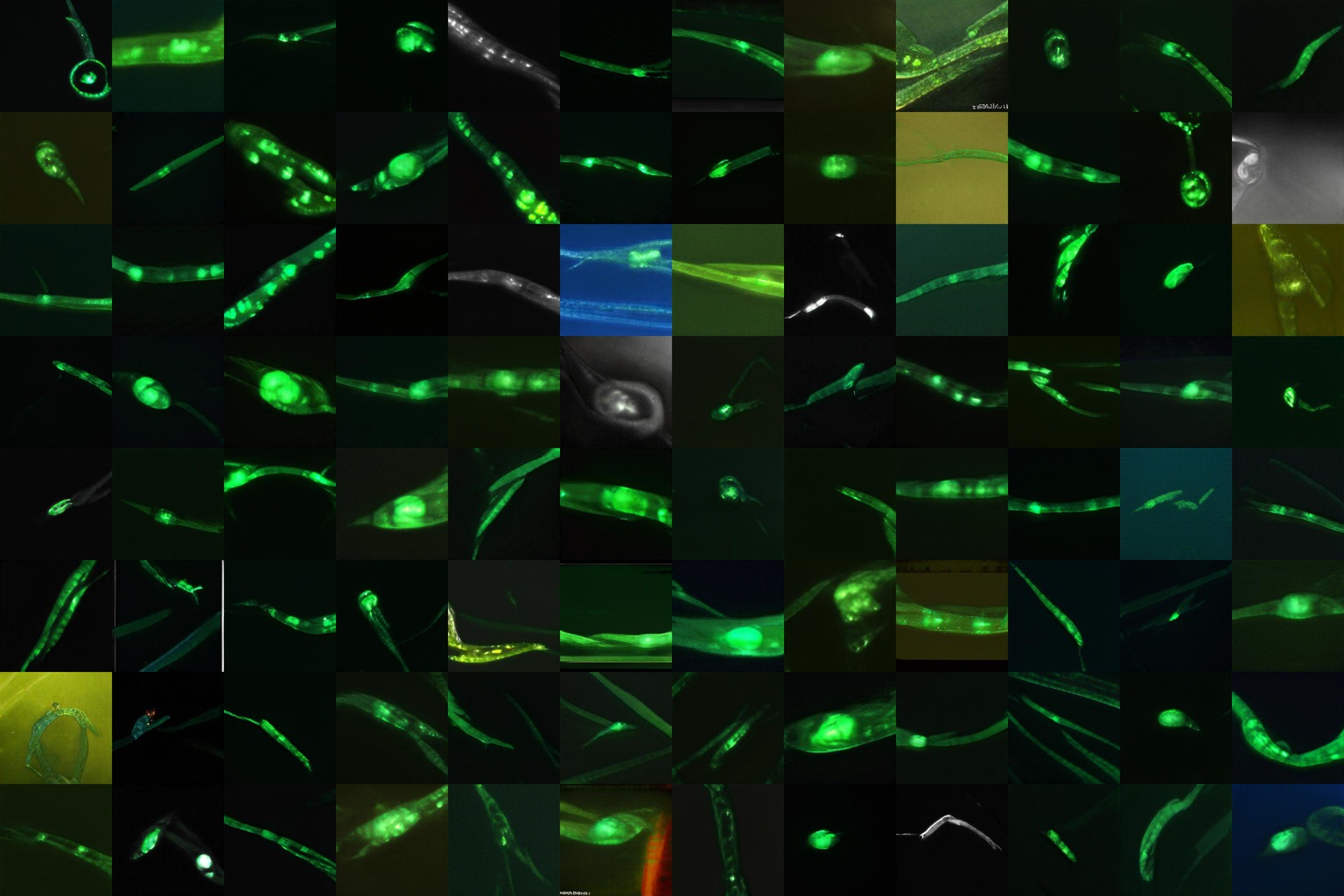}
		\caption{Samples from BigGAN.}
	\end{subfigure}
	\begin{subfigure}[b]{1.0\linewidth}
		\centering
		\includegraphics[width=1.0\linewidth]{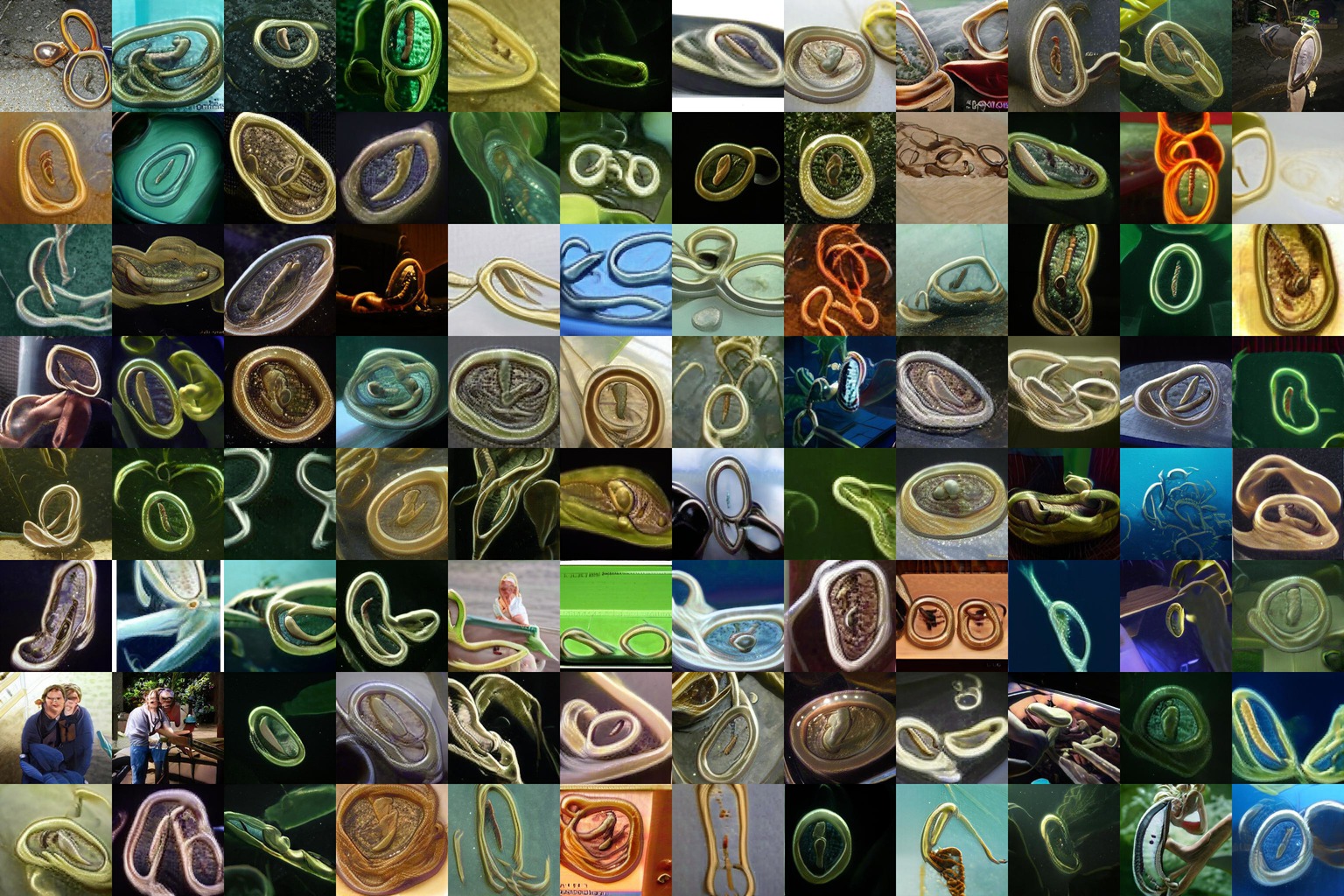}
		\caption{Samples from AM.}
	\end{subfigure}
	\caption{A comparison between the $256\times256$ samples from the original BigGAN model (a), and our AM method (b) for the \class{nematode} class (111).
		AM samples (b) are of similar quality but higher diversity than the original BigGAN samples (a). 
	}
	\label{fig:biggan_vs_am_bigger_111}
\end{figure*}

\begin{figure*}
	\centering
	\begin{subfigure}[b]{1.0\linewidth}
		\centering
		\includegraphics[width=1.0\linewidth]{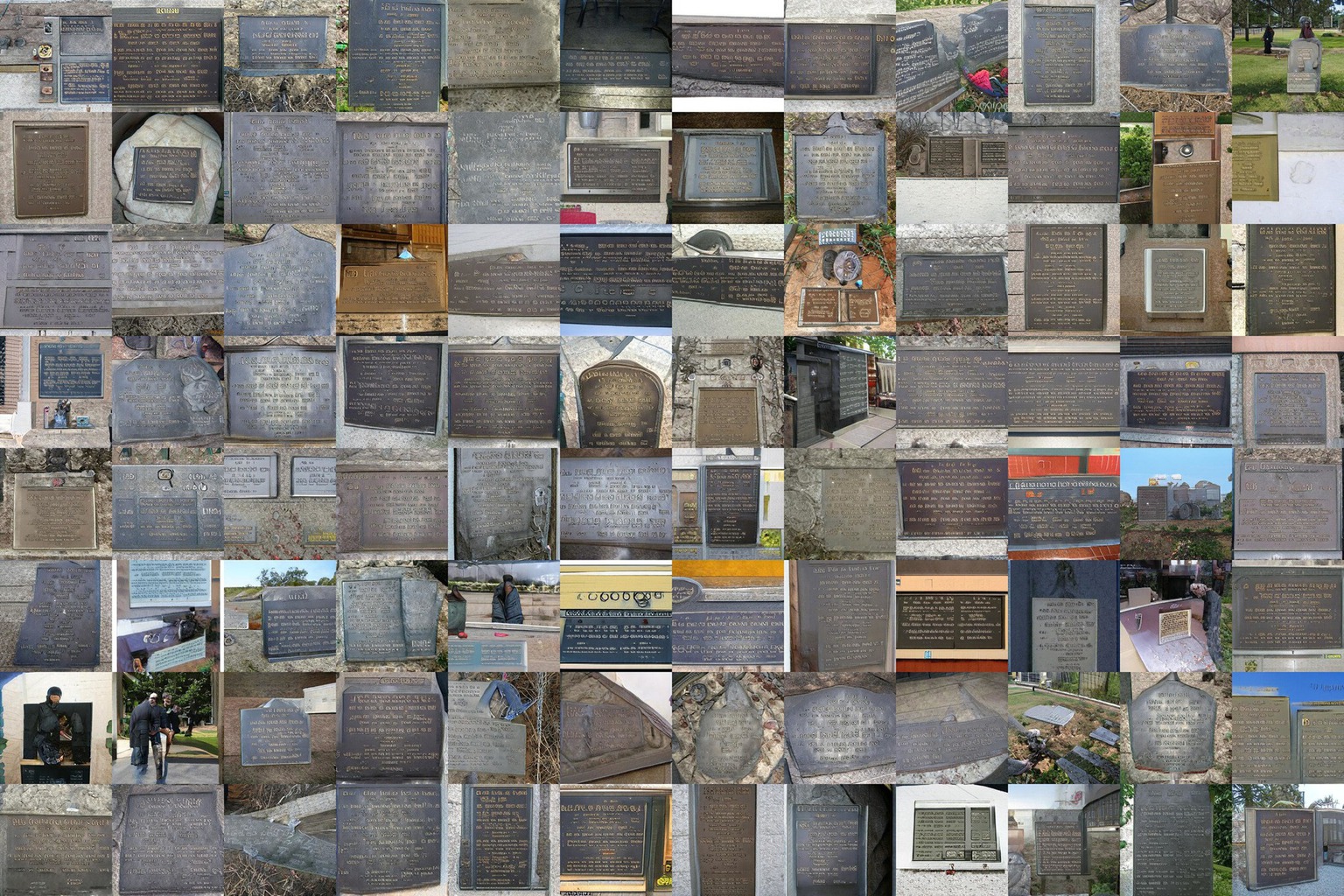}
		\caption{Samples from BigGAN.}
	\end{subfigure}
	\begin{subfigure}[b]{1.0\linewidth}
		\centering
		\includegraphics[width=1.0\linewidth]{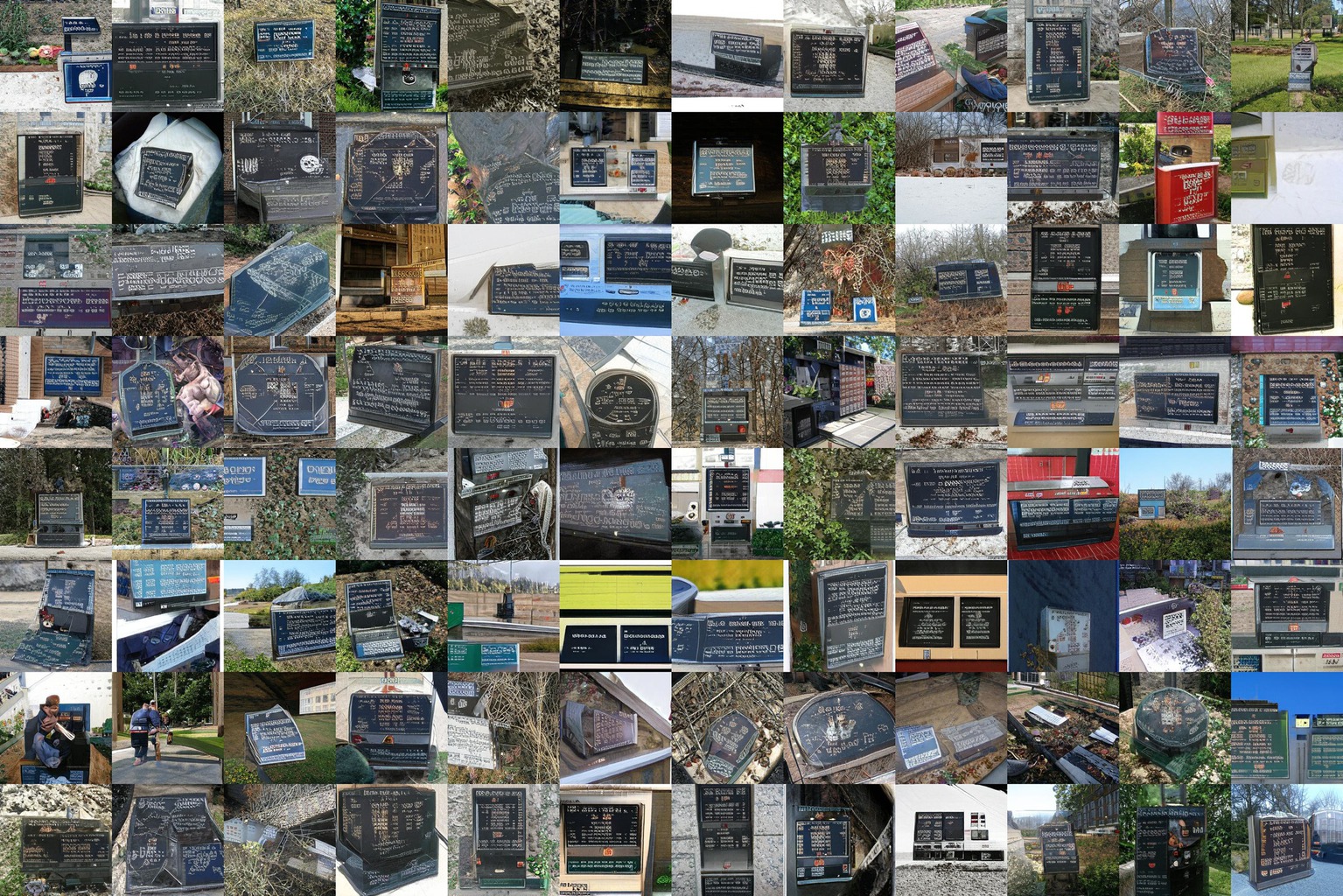}
		\caption{Samples from AM.}
	\end{subfigure}
	\caption{A comparison between the $256\times256$ samples from the original BigGAN model (a), and our AM method (b) for the \class{brass} class (458).
		AM samples (b) are of similar quality but higher diversity than the original BigGAN samples (a). 
	}
	\label{fig:biggan_vs_am_bigger_458}
\end{figure*}

\begin{figure*}
	\centering
	\begin{subfigure}[b]{1.0\linewidth}
		\centering
		\includegraphics[width=1.0\linewidth]{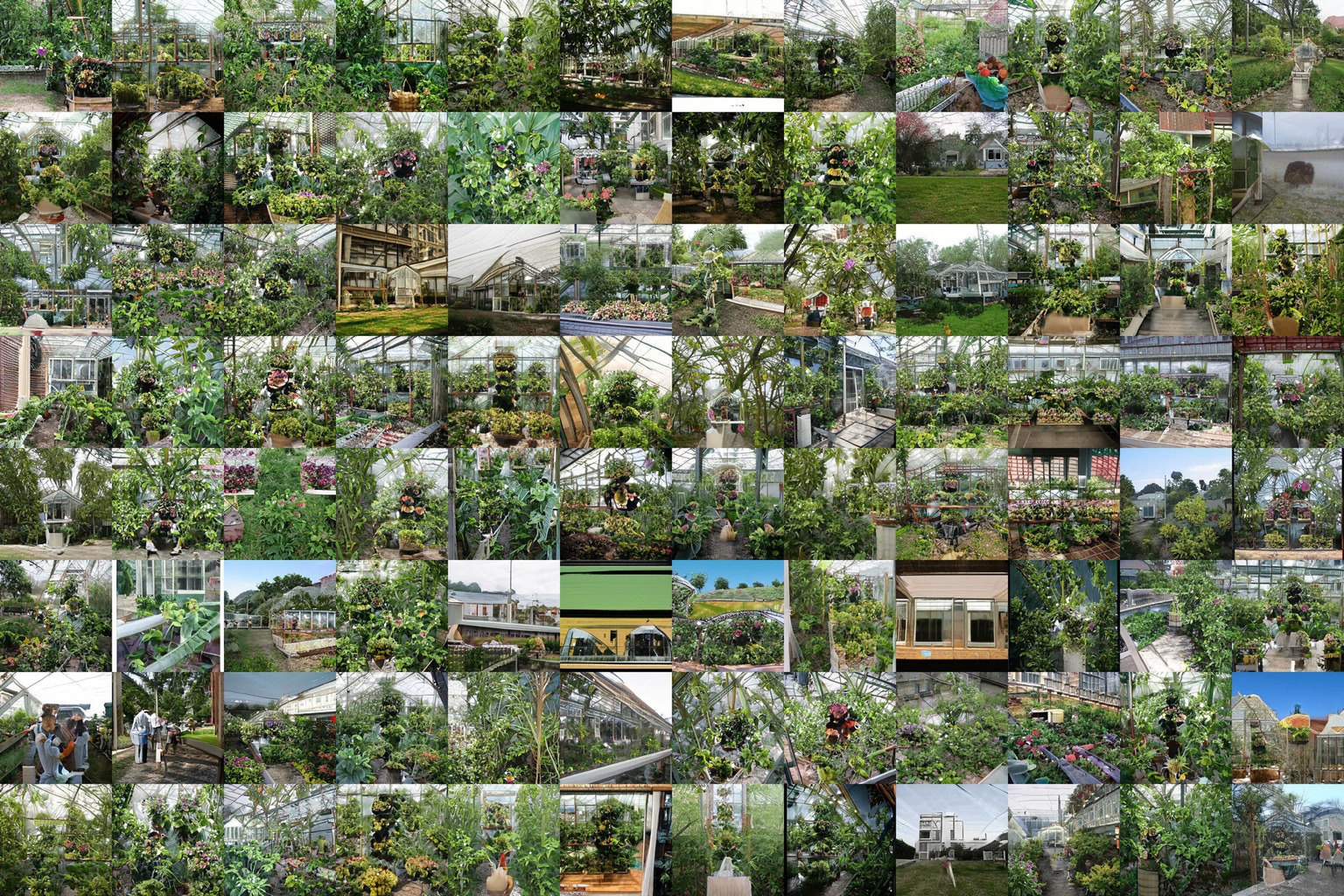}
		\caption{Samples from BigGAN.}
	\end{subfigure}
	\begin{subfigure}[b]{1.0\linewidth}
		\centering
		\includegraphics[width=1.0\linewidth]{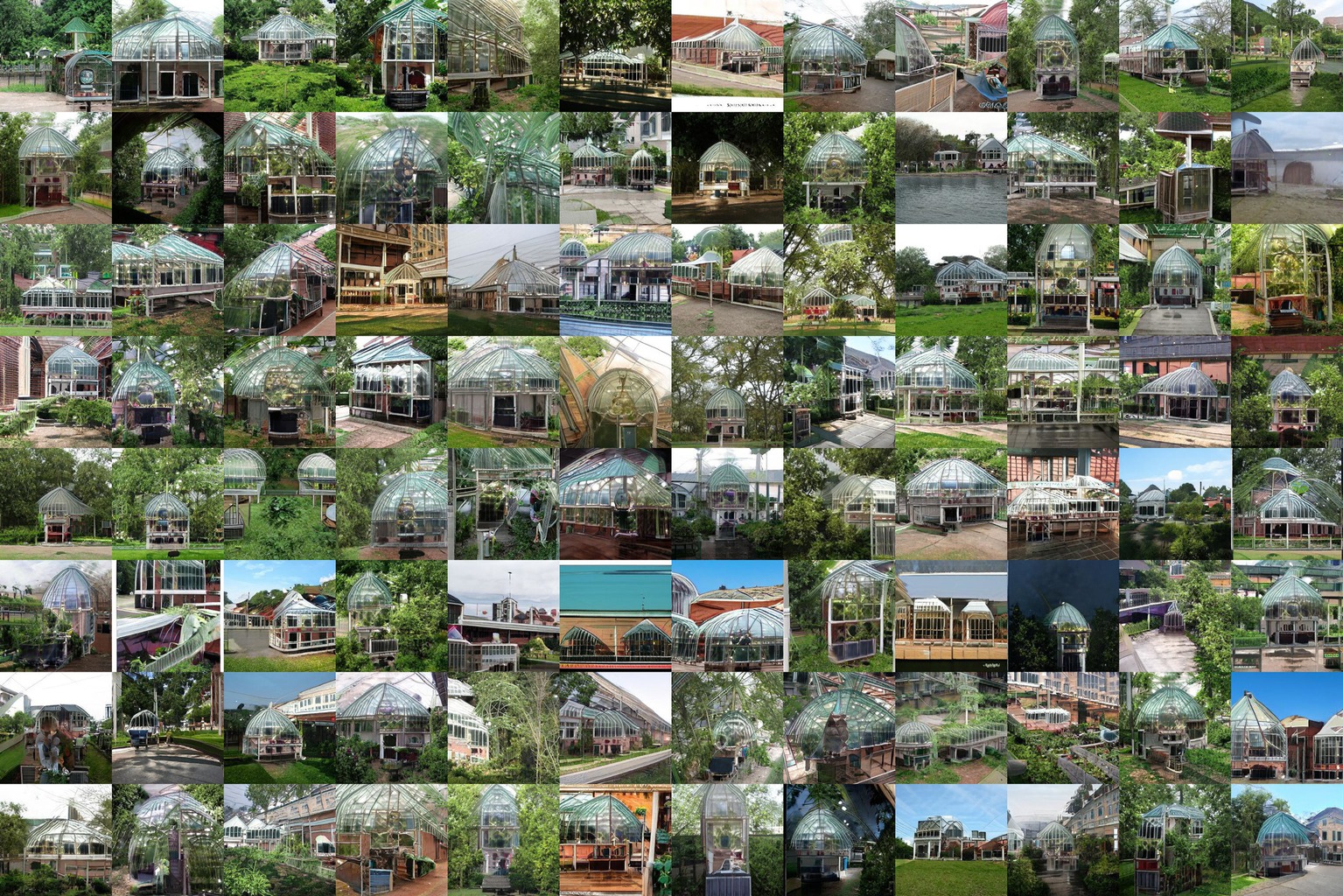}
		\caption{Samples from AM.}
	\end{subfigure}
	\caption{A comparison between the $256\times256$ samples from the original BigGAN model (a), and our AM method (b) for the \class{greenhouse} class (580).
		AM samples (b) are of similar quality but higher diversity than the original BigGAN samples (a). 
	}
	\label{fig:biggan_vs_am_bigger_580}
\end{figure*}

\begin{figure*}
	\centering
	\begin{subfigure}[b]{1.0\linewidth}
		\centering
		\includegraphics[width=1.0\linewidth]{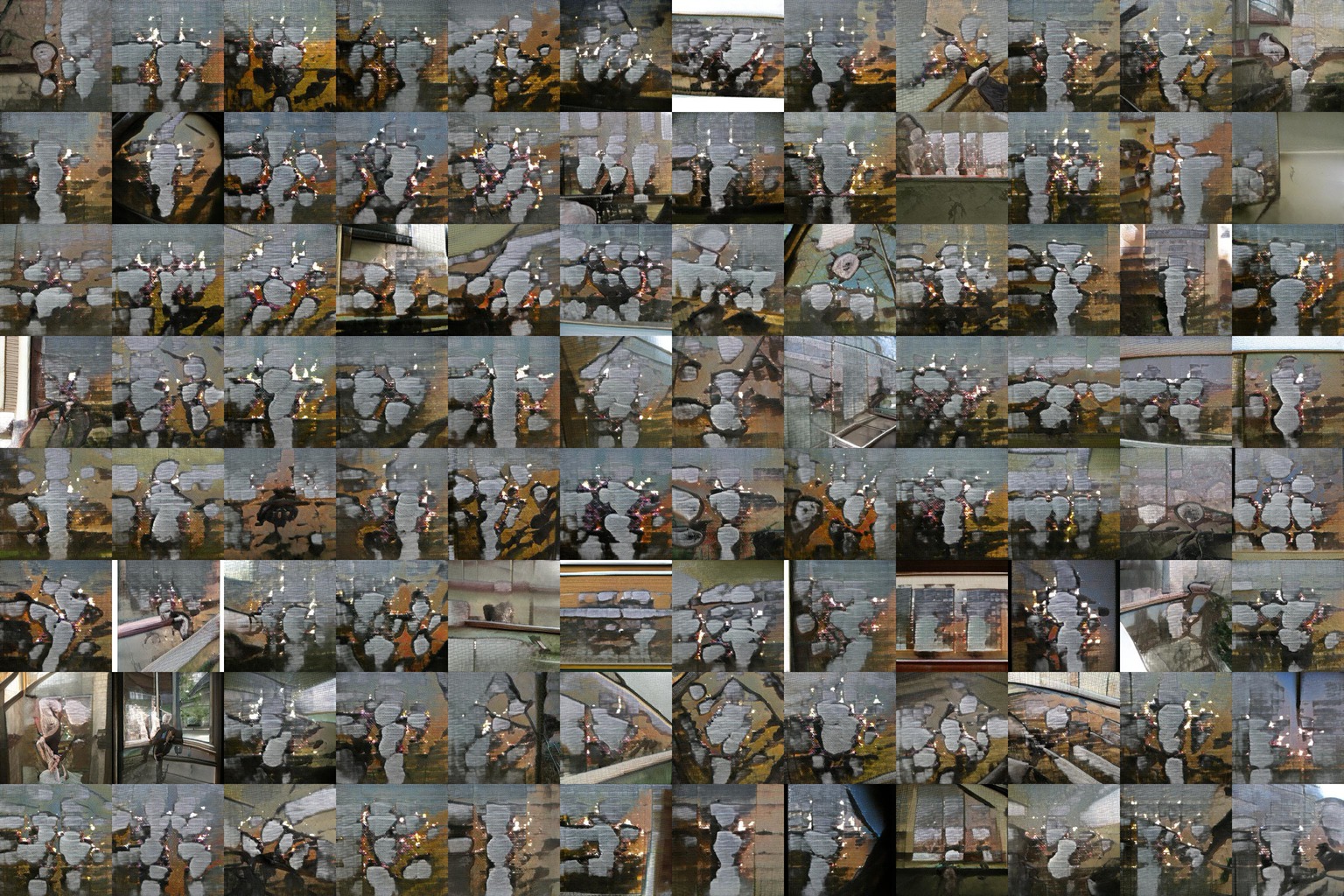}
		\caption{Samples from BigGAN.}
	\end{subfigure}
	\begin{subfigure}[b]{1.0\linewidth}
		\centering
		\includegraphics[width=1.0\linewidth]{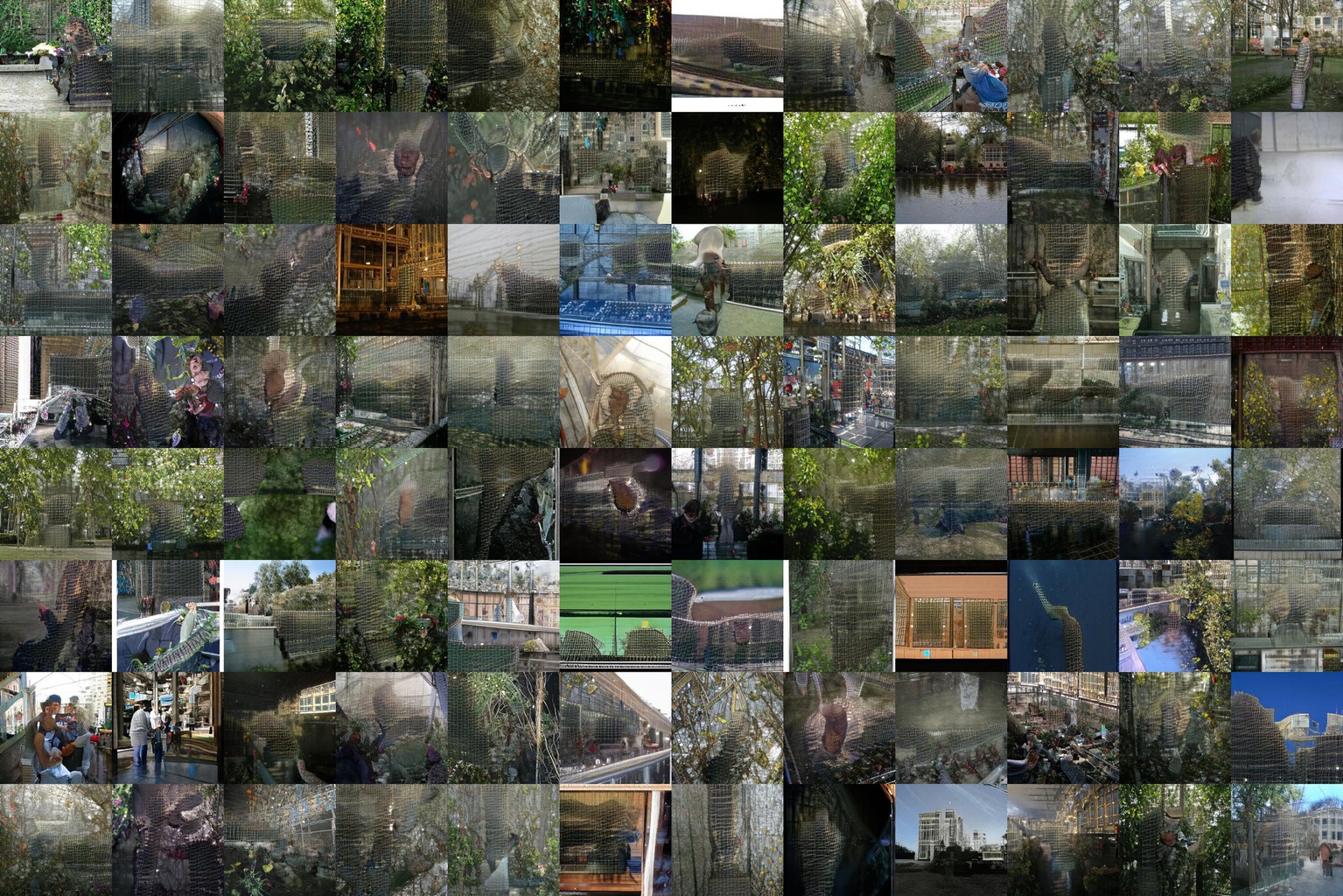}
		\caption{Samples from AM.}
	\end{subfigure}
	\caption{A comparison between the $256\times256$ samples from the original BigGAN model (a), and our AM method (b) for the \class{window~screen} class (904).
		AM samples (b) are both of higher quality and higher diversity than the original BigGAN samples (a). 
	}
	\label{fig:biggan_vs_am_bigger_904}
\end{figure*}

\begin{figure*}
	\centering
	\begin{subfigure}[b]{1.0\linewidth}
		\centering
		\includegraphics[width=1.0\linewidth]{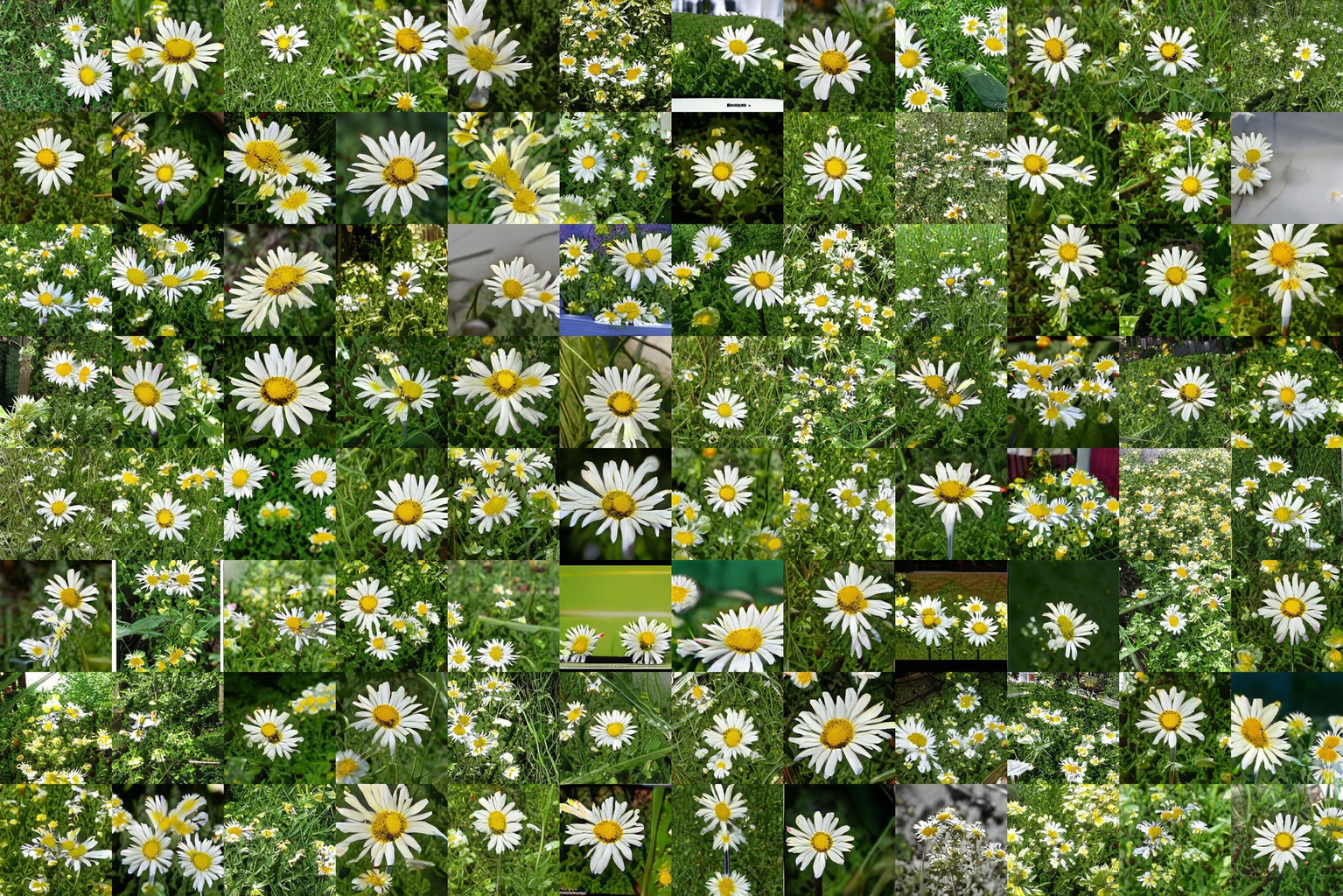}
		\caption{Samples from BigGAN.}
	\end{subfigure}
	\begin{subfigure}[b]{1.0\linewidth}
		\centering
		\includegraphics[width=1.0\linewidth]{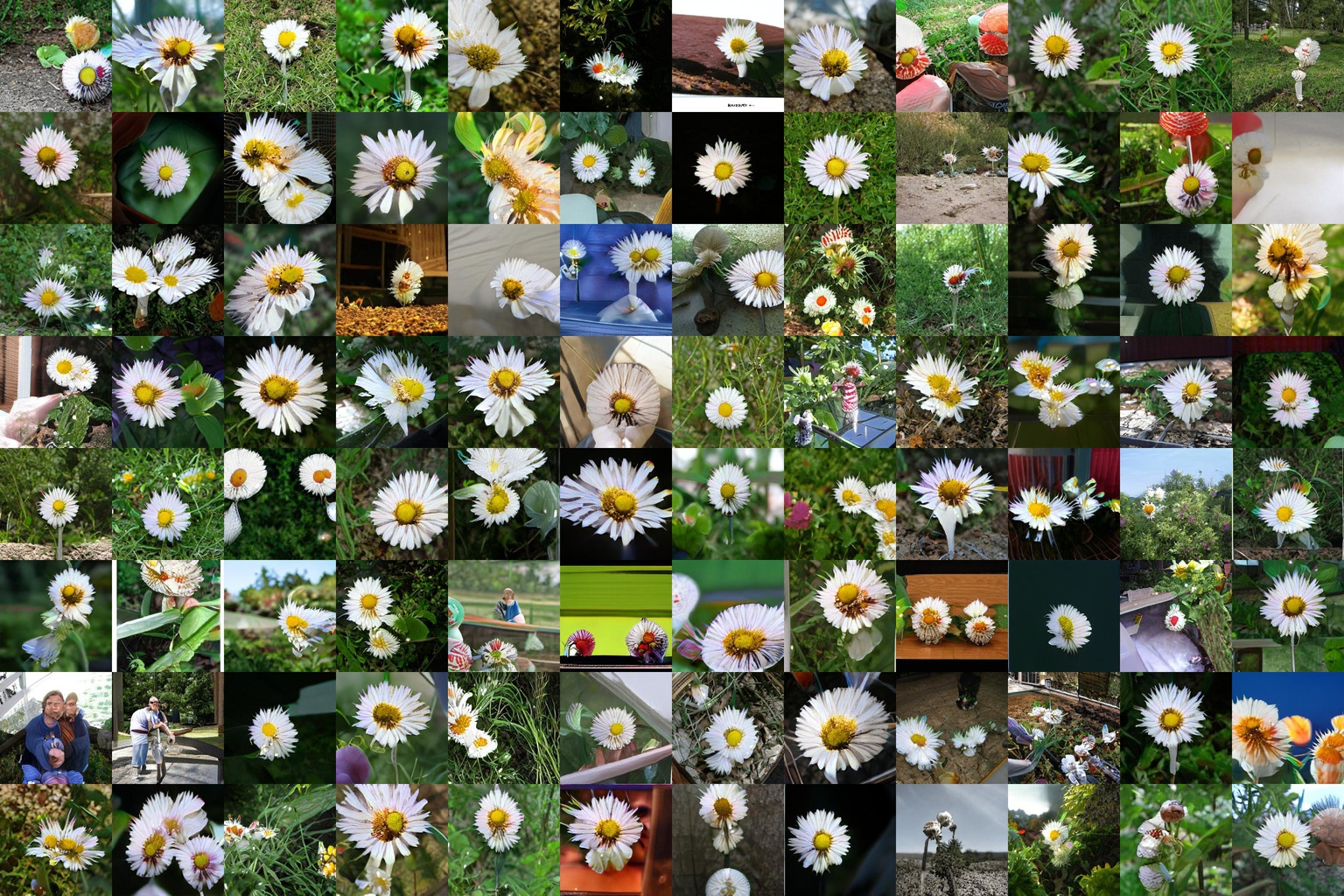}
		\caption{Samples from AM.}
	\end{subfigure}
	\caption{A comparison between the $256\times256$ samples from the original BigGAN model (a), and our AM method (b) for the \class{daisy} class (985).
		AM samples (b) are of similar quality but higher diversity than the original BigGAN samples (a). 
	}
	\label{fig:biggan_vs_am_bigger_985}
\end{figure*}
%BigGAN 128 final results

\begin{figure*}[h!]
	\centering
	{	
		\begin{flushleft}
			\hspace{1.4cm} (A) ImageNet
			\hspace{1.4cm} (B) BigGAN \cite{brock2019large}
			\hspace{1.3cm} (C) AM (ours)
		\end{flushleft}
	}
	
	\begin{subfigure}[b]{1.0\linewidth}
		\centering
		\includegraphics[width=0.9\linewidth]{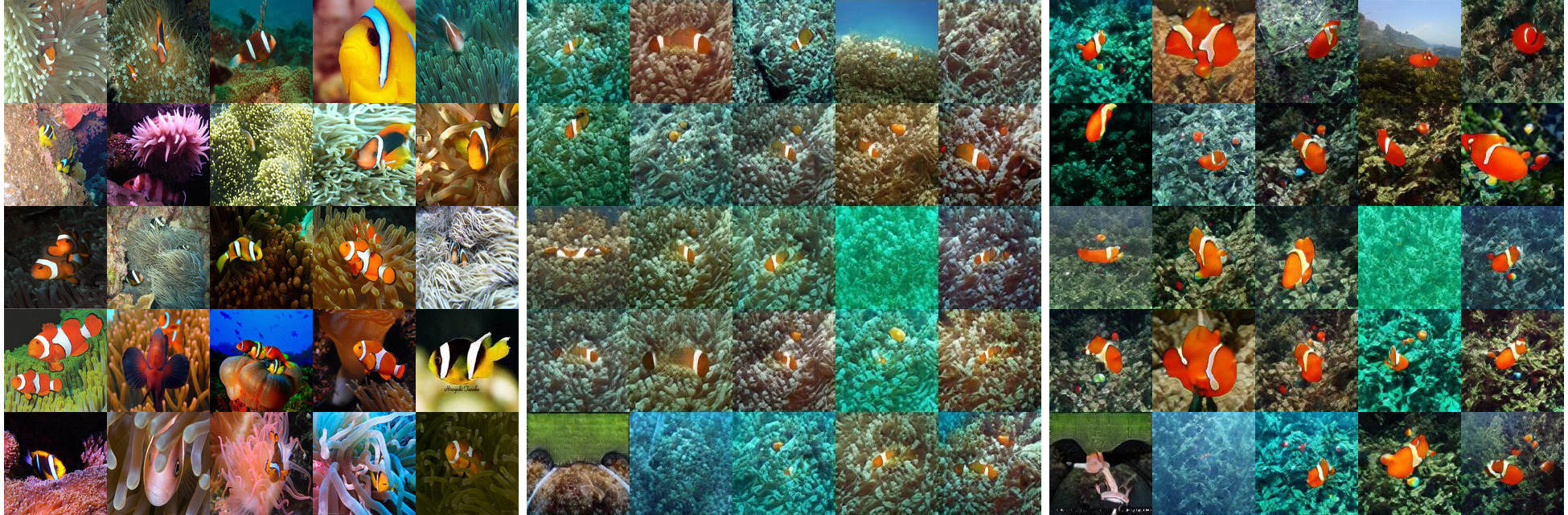}
		\caption{Samples from the \class{anemone~fish} class (393).}
	\end{subfigure}
	\begin{subfigure}[b]{1.0\linewidth}
		\centering
		\includegraphics[width=0.9\linewidth]{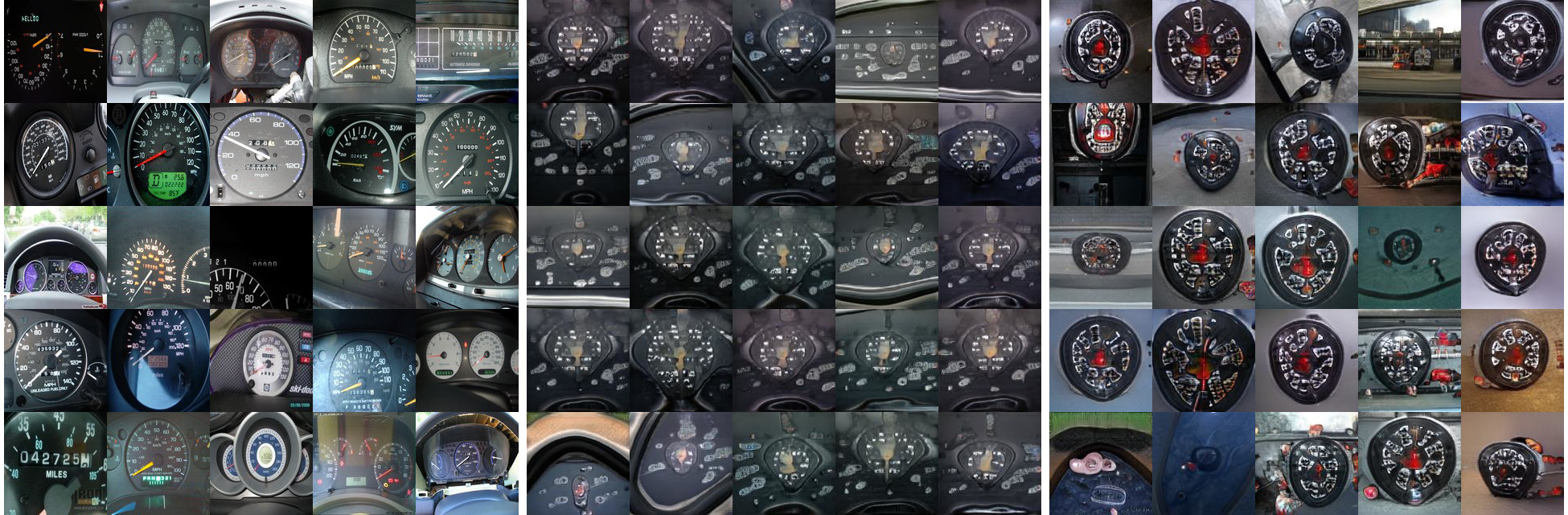}
		\caption{Samples from the \class{odometer} class (685).}
	\end{subfigure}
	\begin{subfigure}[b]{1.0\linewidth}
		\centering
		\includegraphics[width=0.9\linewidth]{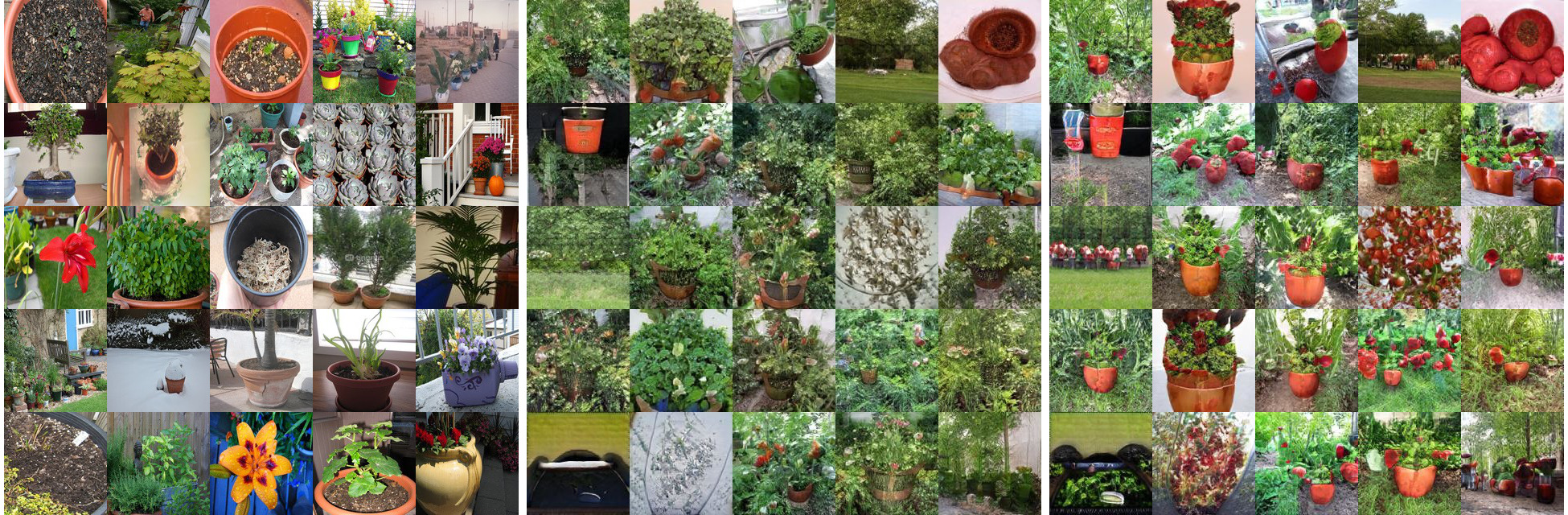}
		\caption{Samples from the \class{flowerpot} class (738).}
	\end{subfigure}
	\begin{subfigure}[b]{1.0\linewidth}
		\centering
		\includegraphics[width=0.9\linewidth]{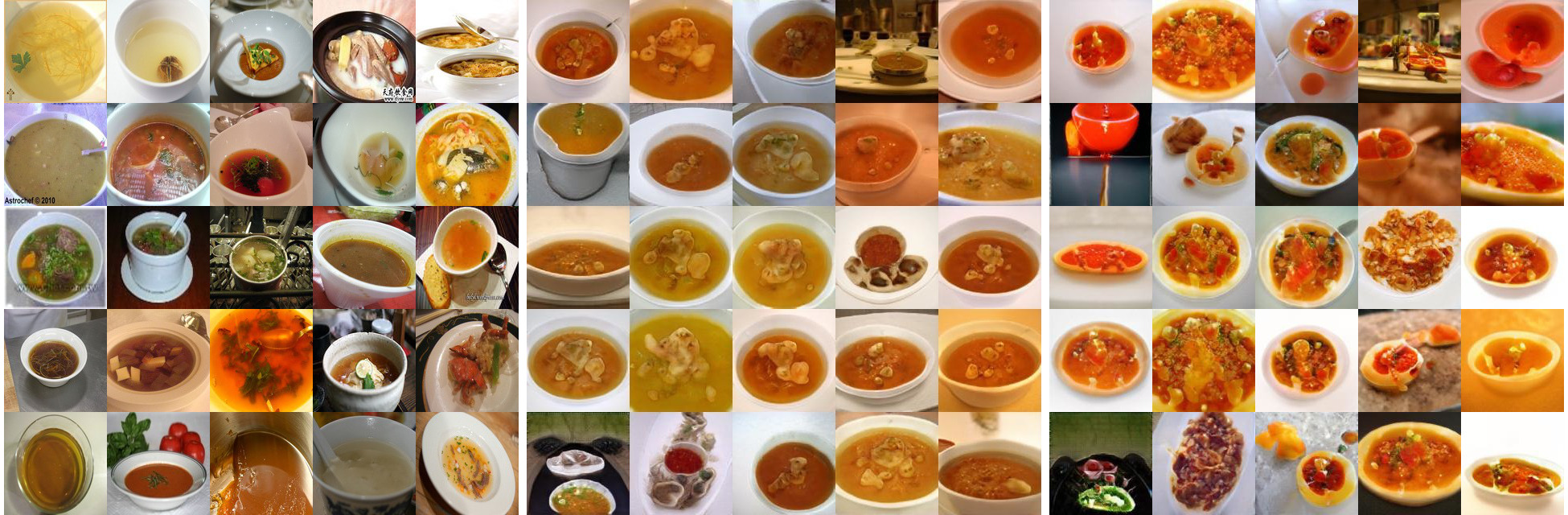}
		\caption{Samples from the \class{consomme} class (925).}
	\end{subfigure}
	\caption{A comparison between the $128\times128$ samples from the ImageNet training set (A), the original BigGAN model (B), and our AM method (C) for four ImageNet-50 low-diversity classes.
		AM samples (C) are of similar quality but higher diversity than the original BigGAN samples (B).}
	\label{fig:biggan-am_128_final_01}
\end{figure*}

\begin{figure*}[h!]
	\centering
	\begin{subfigure}[b]{1.0\linewidth}
		\includegraphics[width=1.0\linewidth]{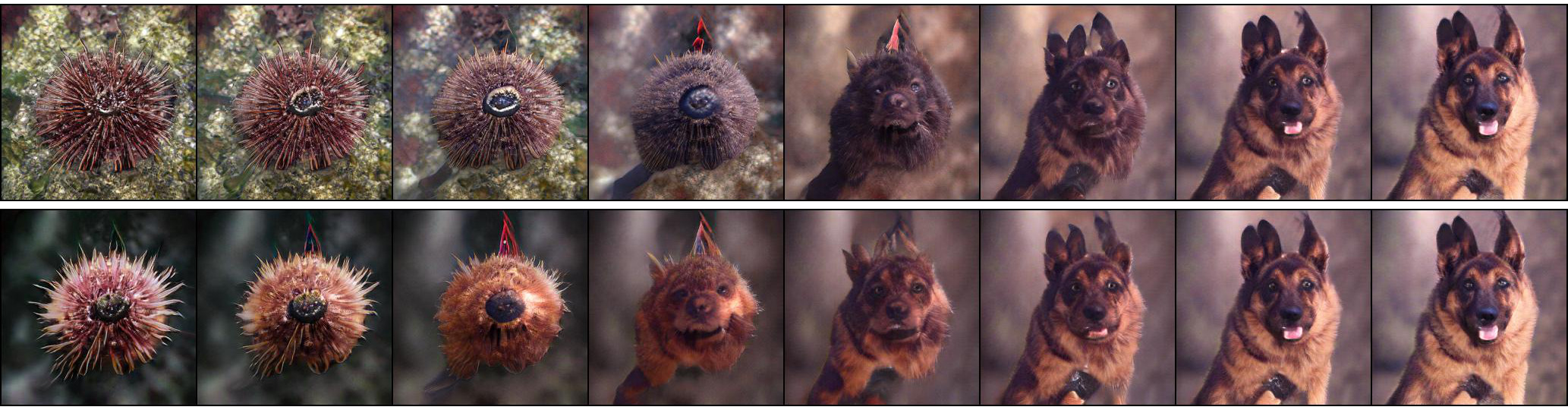}
		\caption{Interpolation in the embedding space between \class{sea urchin} (leftmost) and \class{German~shepherd} (rightmost).}
	\end{subfigure}
	\begin{subfigure}[b]{1.0\linewidth}
		\includegraphics[width=1.0\linewidth]{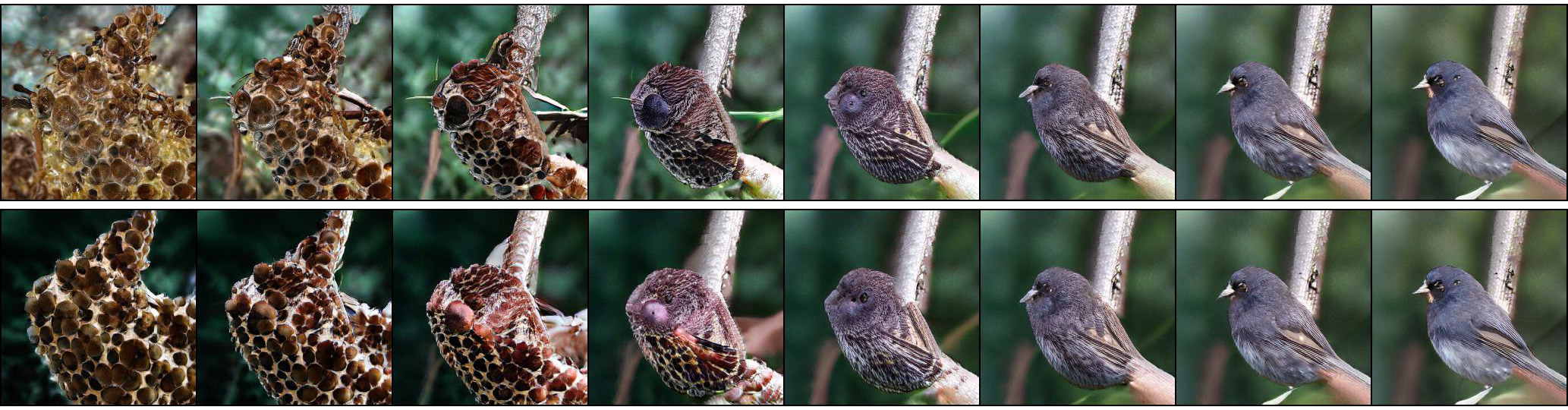}
		\caption{Interpolation in the embedding space between \class{honeycomb} (leftmost) and \class{junco~bird} (rightmost).}
		%		\caption{The most left side is class 599 (\class{honeycomb}), the most right class is 13 (\class{junco}) }
	\end{subfigure}
	\begin{subfigure}[b]{1.0\linewidth}
		\includegraphics[width=1.0\linewidth]{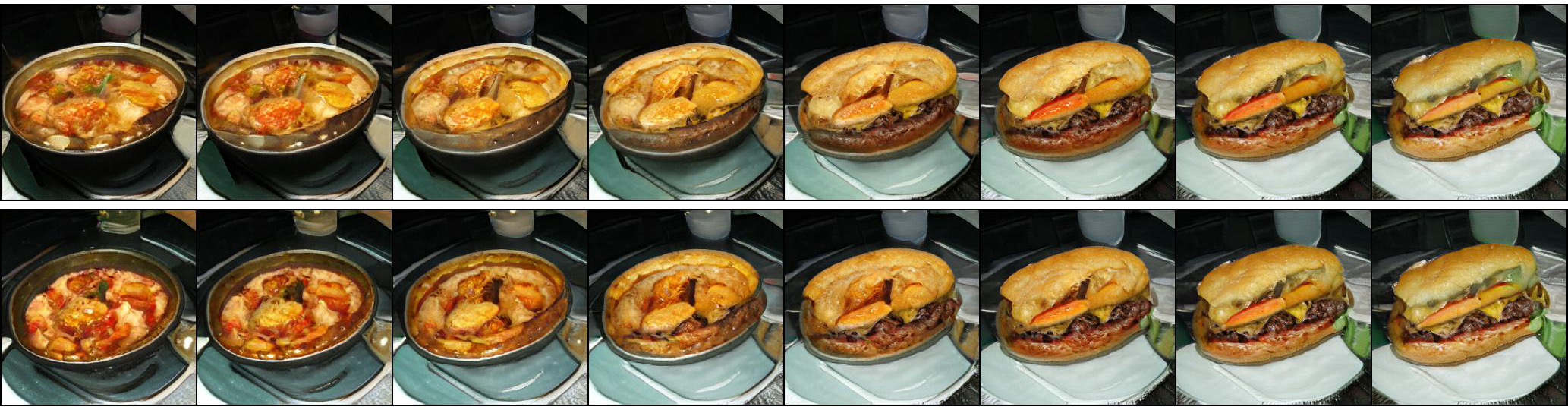}
		%		\caption{The most left side is class 926 (\class{hot pot}), the most right class is 933 (\class{cheeseburger}) }
		\caption{Interpolation in the embedding space between \class{hot~pot} (leftmost) and \class{cheeseburger} (rightmost).}		
	\end{subfigure}
	\caption{The interpolation samples between $\vc$ class-embedding pairs with latent vectors $\vz$ held constant. 
		In each panel, the top row shows the interpolation between two original $256\times256$ BigGAN embeddings while the bottom row shows the interpolation between an embedding found by AM (leftmost) and the original BigGAN embedding (right).
		%	The upper row are the samples from  BigGAN, the lower row are the samples from the embeddings found by our AM methods. 
		%	We optimized embedding weight of the most left side class and kept the most right embedding weight as same as BigGAN's. 
		In sum, the interpolation samples with the AM embeddings (bottom panels) appear to be similarly plausible as the original BigGAN interpolation samples (top panels).}
	\label{fig:inter_class_interpolation}
\end{figure*}

\begin{figure*}[h!]
	\centering
	\begin{subfigure}[b]{1.0\linewidth}
		\includegraphics[width=1.0\linewidth]{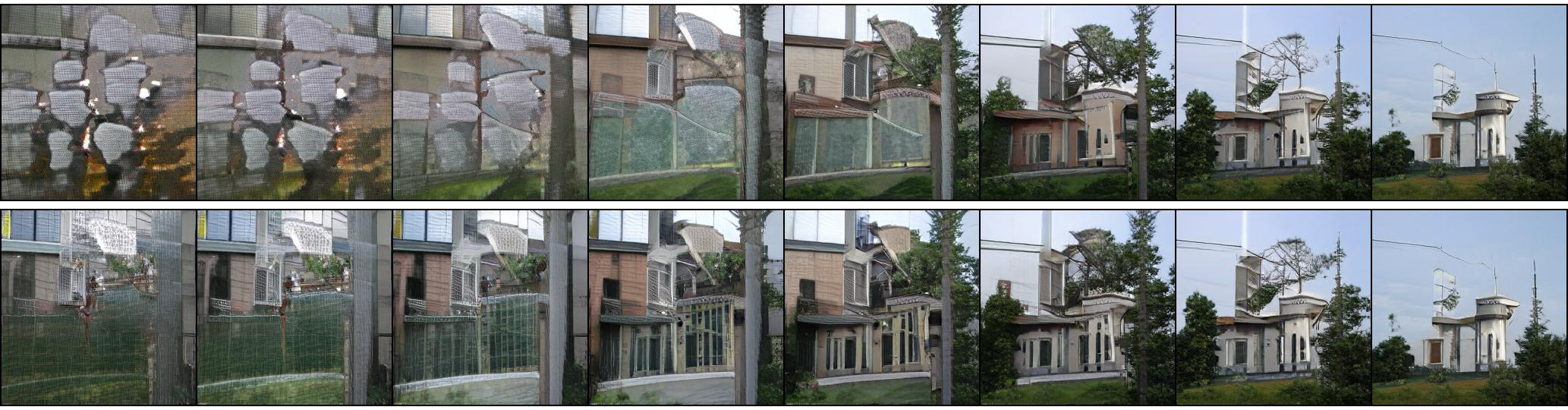}
		%		\caption{The most left side is class 904 (\class{window screen}), the most right class is 900 (\class{water tower})}
		\caption{Interpolation in the embedding space between \class{window~screen} (leftmost) and \class{water~tower} (rightmost).}
	\end{subfigure}
	\begin{subfigure}[b]{1.0\linewidth}
		\includegraphics[width=1.0\linewidth]{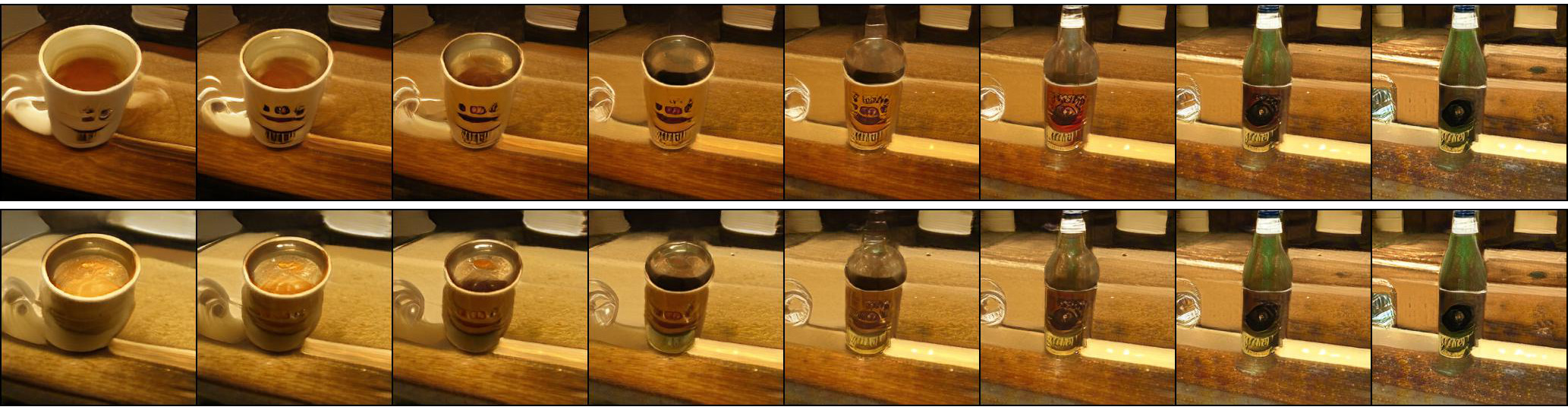}
		%		\caption{The most left side is class 967 (\class{espresso}), the most right class is 737 (\class{pop bottle}) }
		\caption{Interpolation in the embedding space between \class{espresso} (leftmost) and \class{pop~bottle} (rightmost).}
	\end{subfigure}
	\begin{subfigure}[b]{1.0\linewidth}
		\includegraphics[width=1.0\linewidth]{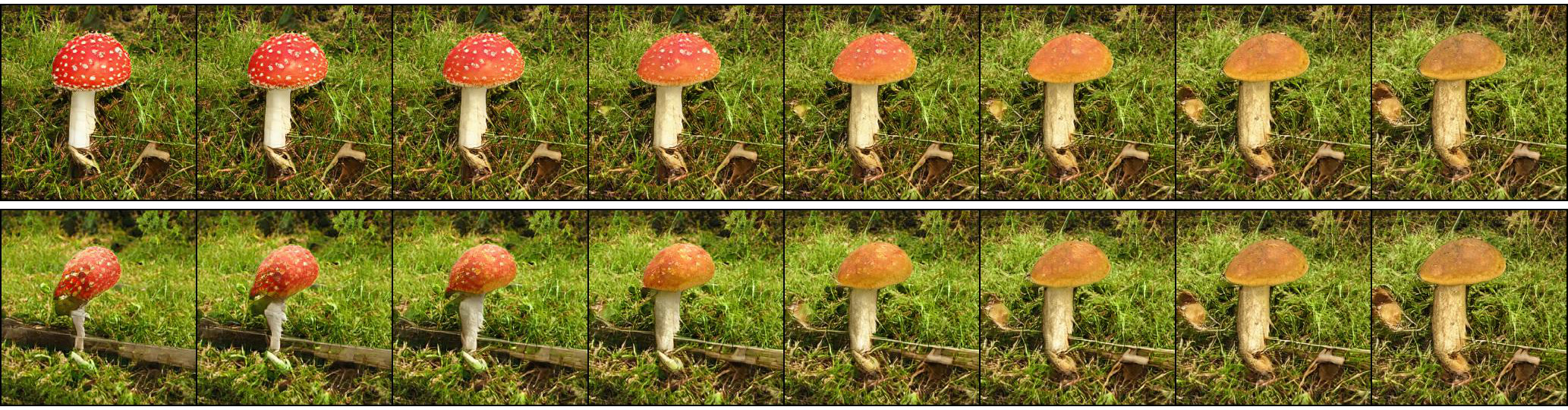}
		%		\caption{The most left side is class 992 (\class{agaric}), the most right class is 997 (\class{bolete}) }
		\caption{Interpolation in the embedding space between \class{agaric} (leftmost) and \class{bolete} (rightmost).}
	\end{subfigure}

	\caption{The interpolation samples between $\vc$ class-embedding pairs (from related ImageNet classes \eg \class{agaric} and \class{bolete} are both mushrooms) with latent vectors $\vz$ held constant. 
		In each panel, the top row shows the interpolation between two original $256\times256$ BigGAN embeddings while the bottom row shows the interpolation between an embedding found by AM (leftmost) and the original BigGAN embedding (right).
		In sum, the interpolation samples with the AM embeddings (bottom panels) appear to be similarly plausible as the original BigGAN interpolation samples (top panels).}
	\label{fig:intra_category_interpolation}
\end{figure*}

\begin{figure*}[h!]
	\centering
	\begin{subfigure}[b]{1.0\linewidth}
		\includegraphics[width=1.0\linewidth]{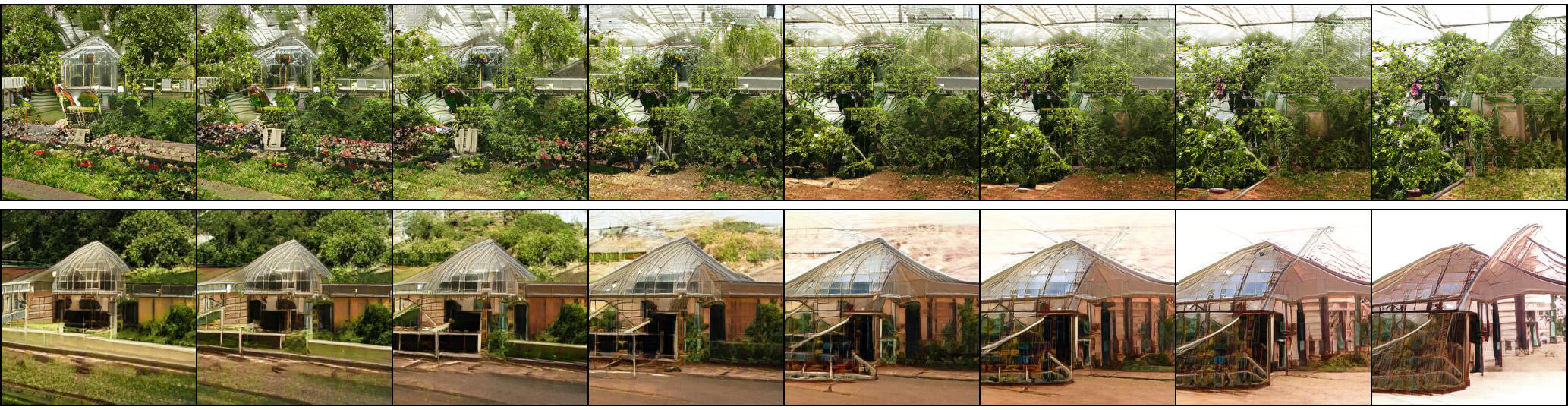}
		%		\caption{The class 580 (\class{greenhouse})}
		\caption{Interpolation in the latent space between two $\vz$ vectors with the same \class{greenhouse} class embedding.}
	\end{subfigure}
	\begin{subfigure}[b]{1.0\linewidth}
		\includegraphics[width=1.0\linewidth]{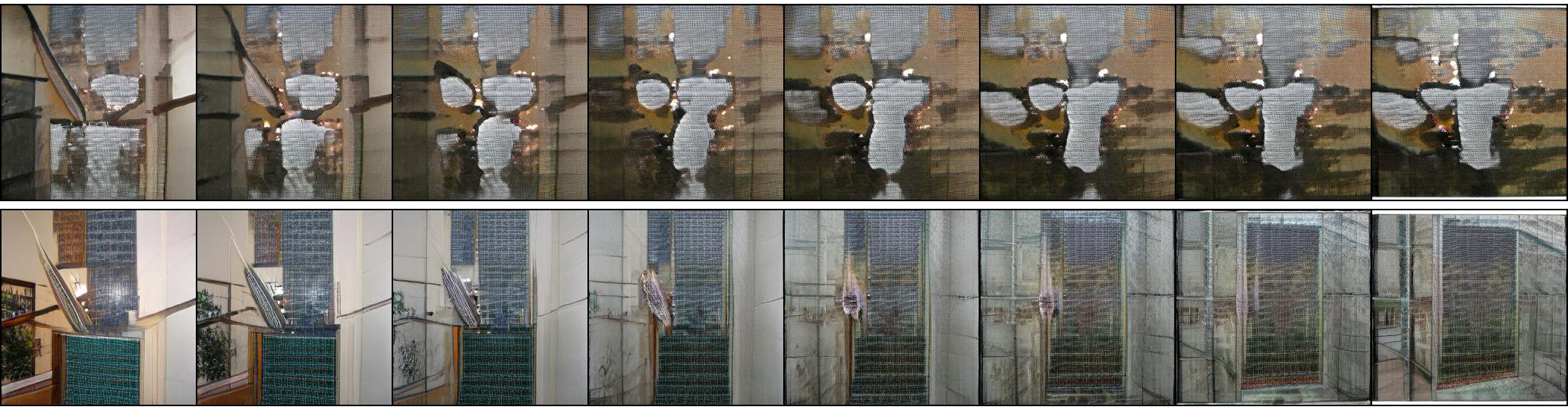}
		%		\caption{The class 904 (\class{window screen})}
		\caption{Interpolation in the latent space between two $\vz$ vectors with the same \class{window~screen} class embedding.}		
		\label{fig:window_screen_interpolation}
	\end{subfigure}
	\begin{subfigure}[b]{1.0\linewidth}
		\includegraphics[width=1.0\linewidth]{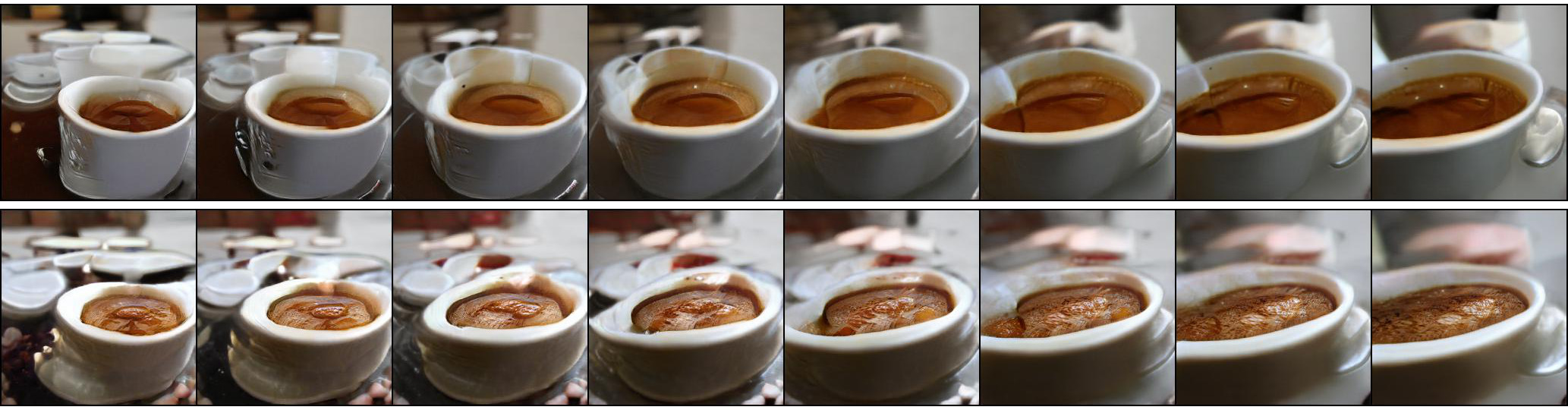}
		%		\caption{The class 967 (\class{espresso})}
		\caption{Interpolation in the latent space between two $\vz$ vectors with the same \class{espresso} class embedding.}				
	\end{subfigure}
	\begin{subfigure}[b]{1.0\linewidth}
		\includegraphics[width=1.0\linewidth]{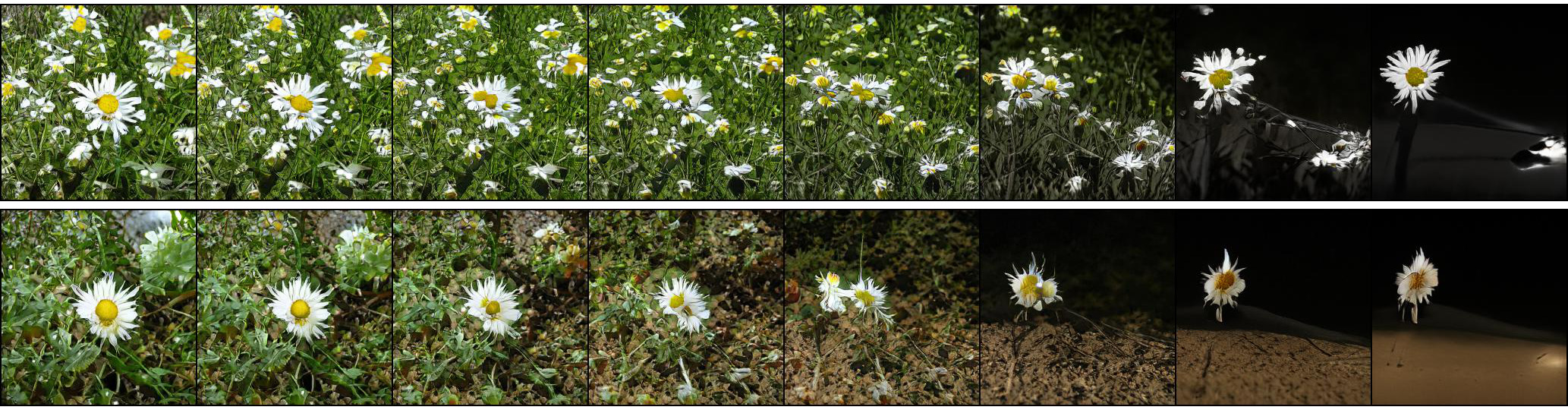}
		%		\caption{The class 985 (\class{daisy}) }
		\caption{Interpolation in the latent space between two $\vz$ vectors with the same \class{daisy~flower} class embedding.}				
	\end{subfigure}
	%	\caption{The interpolations between z pairs with c held constant. The upper row are the samples from BigGAN at 256x256 resolution, the lower row are the samples from AM at 256x256 resolution.}
	\caption{The interpolation samples between $\vz$ latent-vector pairs with the same class embeddings. 
		The $\vz$-interpolation samples with the AM embeddings (bottom panels) appear to be similarly plausible as the original BigGAN interpolation samples (top panels).
		For the \class{window~screen} class (b), AM recovered the human-unrecognizable BigGAN samples into a plausible interpolation between two scenes of windows.
	}	
	
	\label{fig:interpolate_z}
\end{figure*}

% MIT PLACES 365 BigGAN-AM final results

\begin{figure*}[h!]
	\centering
	{	
		\begin{flushleft}
			\hspace{1.5cm} (A) Places365
			\hspace{0.45cm} (B) BigGAN on ImageNet		
			\hspace{0.3cm} (C) AM (ours)
		\end{flushleft}
	}
	\begin{subfigure}[b]{1.0\linewidth}
		\centering
		\includegraphics[width=0.85\linewidth]{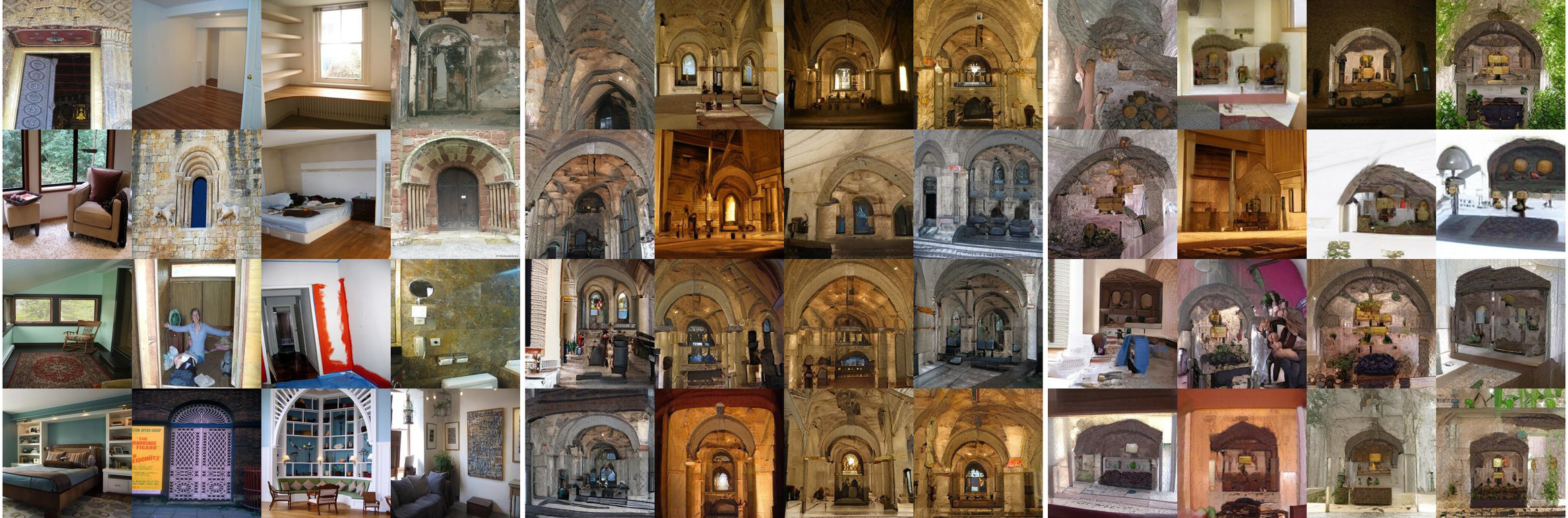}
		\vspace{-0.25cm}
		{
			\begin{flushleft}
				\hspace{2.0cm} \class{alcove}		
				\hspace{2.5cm}  \class{vault}
				\hspace{2.5cm} \class{alcove}
			\end{flushleft}
		}
		\vspace{0.1cm}
		%\caption{Class 003 (\class{alcove})}
	\end{subfigure}
	\begin{subfigure}[b]{1.0\linewidth}
		\centering
		\includegraphics[width=0.85\linewidth]{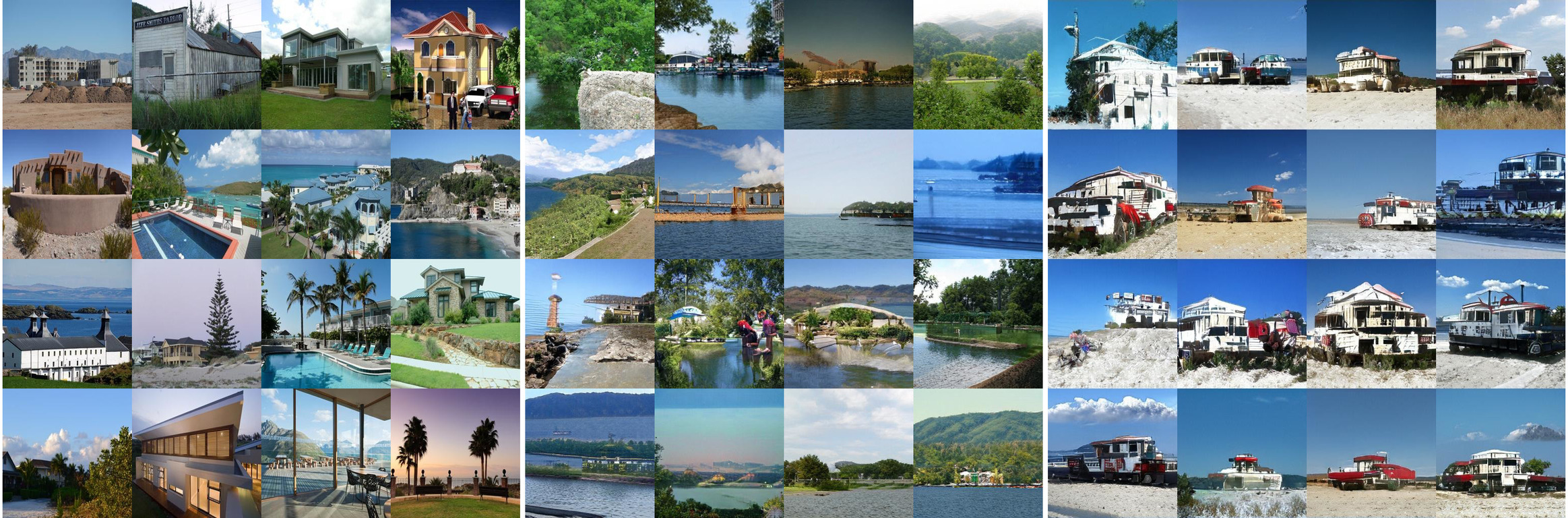}
		\vspace{-0.25cm}
		{
			\begin{flushleft}
				\hspace{1.7cm} \class{beach~house}		
				\hspace{1.8cm}  \class{lakeshore}
				\hspace{1.8cm} \class{beach~house}
			\end{flushleft}
		}
		\vspace{0.1cm}
		%\caption{Class 049 (\class{beach~house})}
	\end{subfigure}
	\begin{subfigure}[b]{1.0\linewidth}
		\centering
		\includegraphics[width=0.85\linewidth]{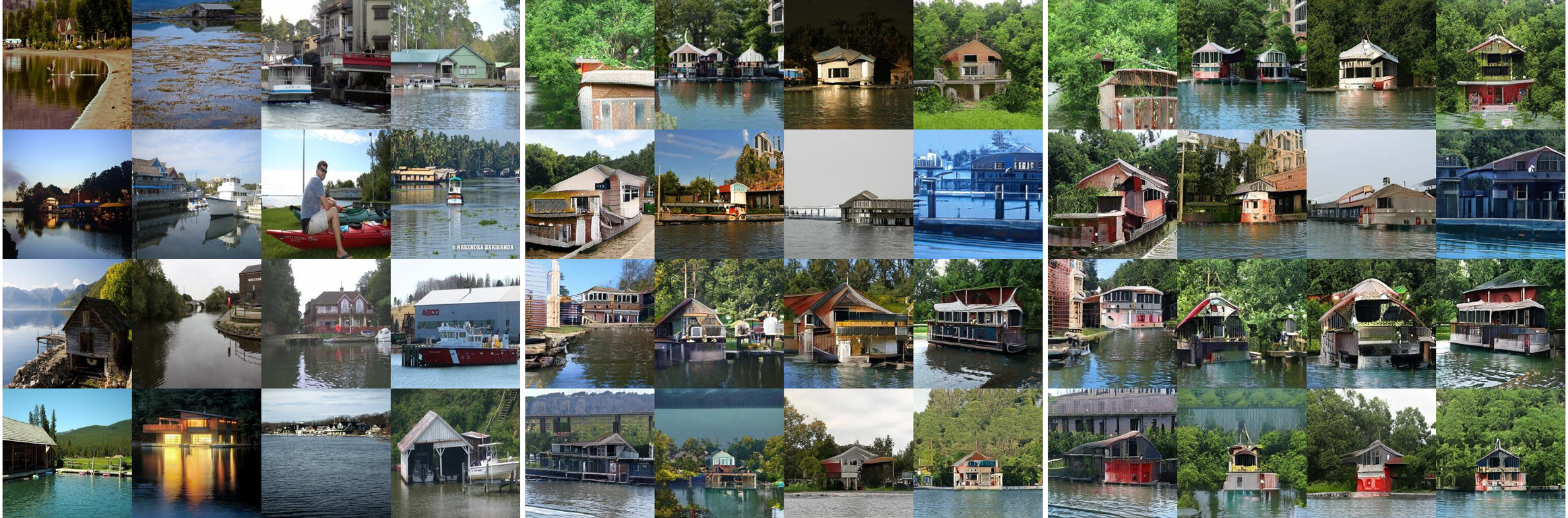}
		\vspace{-0.25cm}
		{
			\begin{flushleft}
				\hspace{1.8cm} \class{boathouse}		
				\hspace{1.8cm}  \class{boathouse}
				\hspace{1.8cm} \class{boathouse}
			\end{flushleft}
		}
		\vspace{0.1cm}
		%\caption{Class 059 (\class{boathouse})}
	\end{subfigure}
	\begin{subfigure}[b]{1.0\linewidth}
		\centering
		\includegraphics[width=0.85\linewidth]{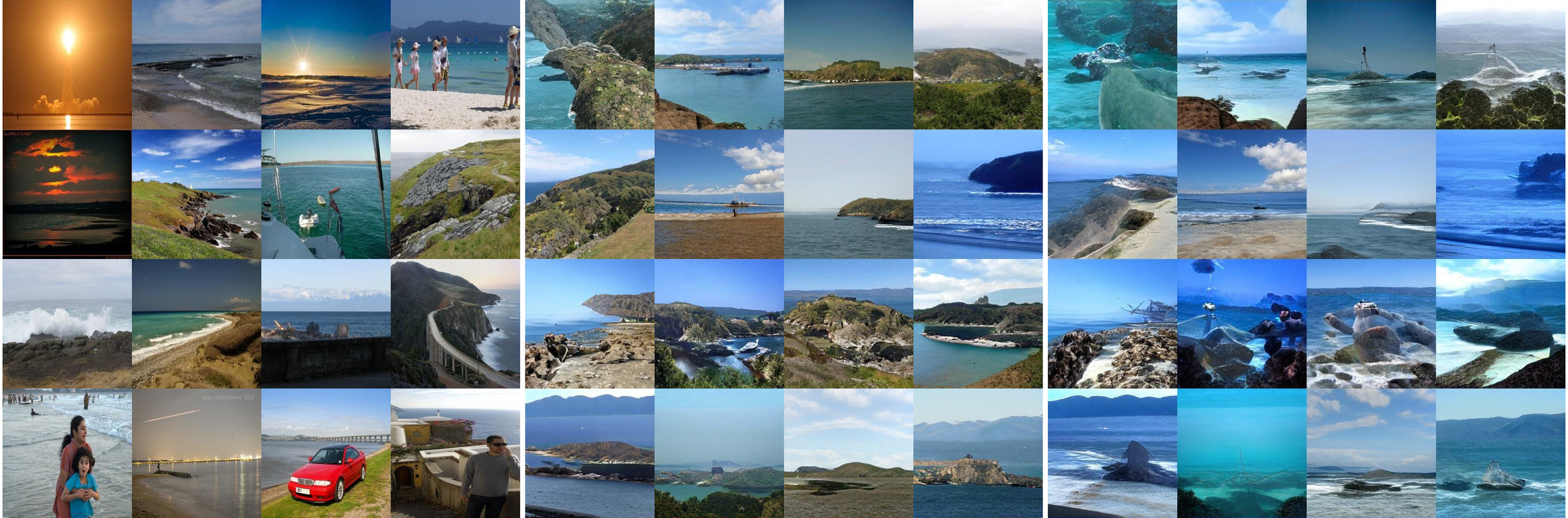}
		\vspace{-0.25cm}
		{
			\begin{flushleft}
				\hspace{2.2cm} \class{coast}		
				\hspace{2.1cm}  \class{promontory}
				\hspace{2.1cm} \class{coast}
			\end{flushleft}
		}
		%\vspace{0.1cm}
		%\caption{Class 097 (\class{coast})}
	\end{subfigure}
	\caption{A comparison between the $256\times256$ samples from the Places365 training set (A), the BigGAN samples generated for the ImageNet class whose 10 random samples were given the highest accuracy for the target class in Places365 (B), and our AM samples (C). AM samples (C) are of similar diversity but better quality than the original BigGAN samples (B). See \url{https://drive.google.com/drive/folders/1L-1ULPfOf_5-98I7emYW86OPDu3Fjxnx?usp=sharing} for a high-resolution version of this figure.}
	\label{fig:mit_places_final_01}
\end{figure*}

\begin{figure*}[h!]
	\centering
	{	
		\begin{flushleft}
			\hspace{1.5cm} (A) Places365
			\hspace{0.45cm} (B) BigGAN on ImageNet		
			\hspace{0.3cm} (C) AM (ours)
		\end{flushleft}
	}
	\begin{subfigure}[b]{1.0\linewidth}
		\centering
		\includegraphics[width=0.85\linewidth]{images/places365_biggan_biggan-am_182.jpg}
		\vspace{-0.25cm}
		{
			\begin{flushleft}
				\hspace{1.8cm}  \class{hotel~room}		
				\hspace{2.1cm}  \class{quilt}
				\hspace{2.2cm}  \class{hotel~room}
			\end{flushleft}
		}
		\vspace{0.1cm}
		%\caption{Class 182 (\class{hotel~room})}
	\end{subfigure}
	\begin{subfigure}[b]{1.0\linewidth}
		\centering
		\includegraphics[width=0.85\linewidth]{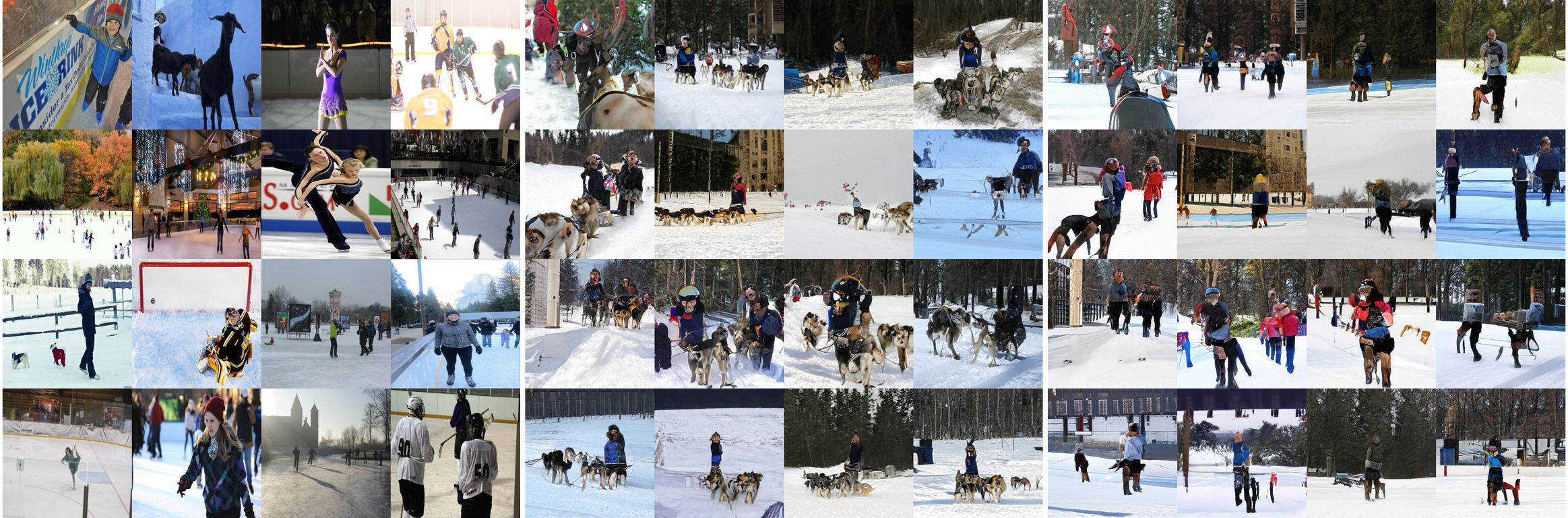}
		\vspace{-0.25cm}
		{
			\begin{flushleft}
				\hspace{0.9cm}  \class{ice~skating~rink~outdoor}		
				\hspace{1.1cm}  \class{dogsled}
				\hspace{1.1cm}  \class{ice~skating~rink~outdoor}
			\end{flushleft}
		}
		\vspace{0.1cm}
		%\caption{Class 189 (\class{ice~skating~rink~outdoor})}
	\end{subfigure}
	\begin{subfigure}[b]{1.0\linewidth}
		\centering
		\includegraphics[width=0.85\linewidth]{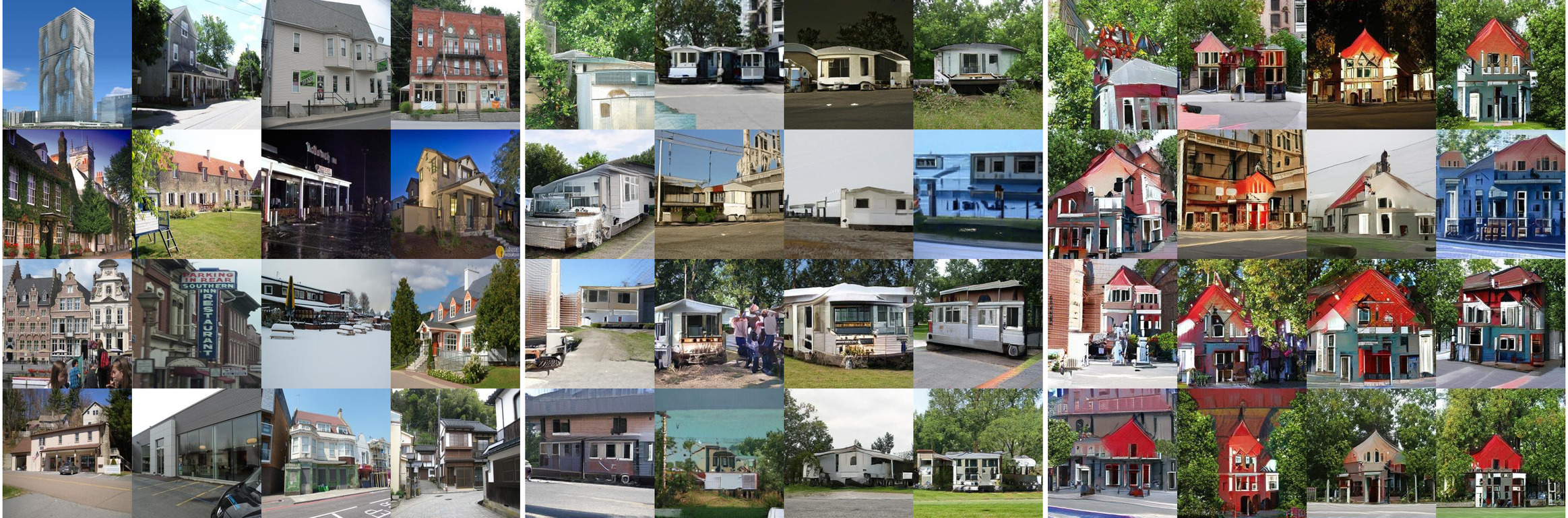}
		\vspace{-0.25cm}
		{
			\begin{flushleft}
				\hspace{1.7cm}  \class{inn~outdoor}		
				\hspace{1.6cm}  \class{mobile~home}
				\hspace{1.6cm}  \class{inn~outdoor}
			\end{flushleft}
		}
		\vspace{0.1cm}
		%\caption{Class 193 (\class{inn~outdoor})}
	\end{subfigure}
	\begin{subfigure}[b]{1.0\linewidth}
		\centering
		\includegraphics[width=0.85\linewidth]{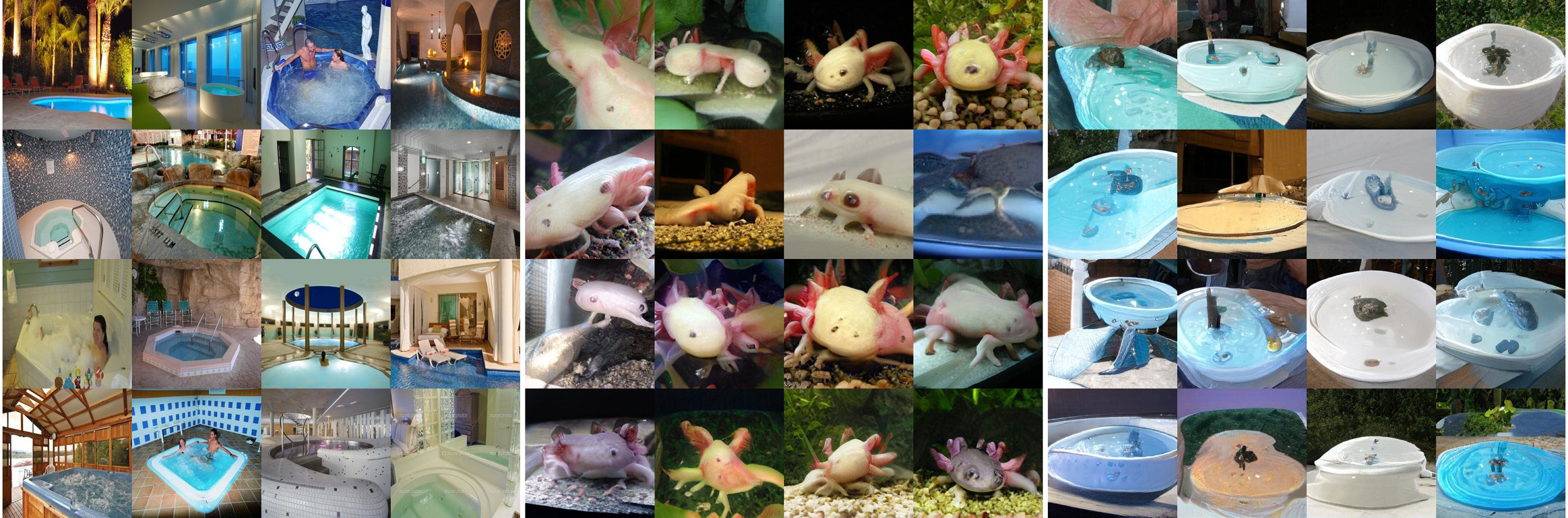}
		\vspace{-0.25cm}
		{
			\begin{flushleft}
				\hspace{1.6cm}  \class{jacuzzi~indoor}		
				\hspace{1.8cm}  \class{axolotl}
				\hspace{1.9cm}  \class{jacuzzi~indoor}
			\end{flushleft}
		}
		%\vspace{0.1cm}
		%\caption{Class 195 (\class{jacuzzi~indoor})}
	\end{subfigure}
	\caption{
		The same figure as Fig.~\ref{fig:mit_places_final_01} but for four different classes.
		While the ImageNet \class{axolotl} class samples were given the highest accuracy (bottom panel), they are qualitatively more different from the real \class{jacuzzi} images compared to the AM samples which shows the bathtubs. See \url{https://drive.google.com/drive/folders/1L-1ULPfOf_5-98I7emYW86OPDu3Fjxnx?usp=sharing} for a high-resolution version of this figure.
	}
	\label{fig:mit_places_final_02}
\end{figure*}

\begin{figure*}[h!]
	\centering
	{	
		\begin{flushleft}
			\hspace{1.5cm} (A) Places365
			\hspace{0.45cm} (B) BigGAN on ImageNet		
			\hspace{0.3cm} (C) AM (ours)
		\end{flushleft}
	}
	\begin{subfigure}[b]{1.0\linewidth}
		\centering
		\includegraphics[width=0.85\linewidth]{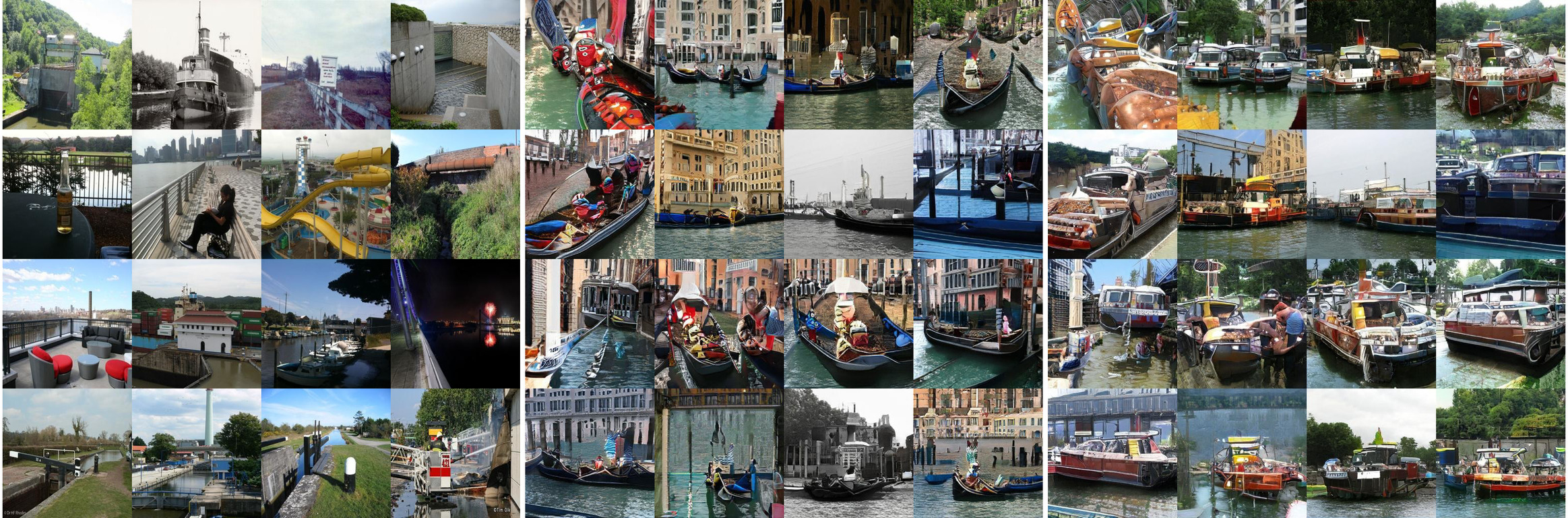}
		\vspace{-0.25cm}
		{
			\begin{flushleft}
				\hspace{1.6cm}  \class{lock~chamber}		
				\hspace{1.8cm}  \class{gondola}
				\hspace{1.7cm}  \class{lock~chamber}
			\end{flushleft}
		}
		\vspace{0.1cm}
		%\caption{Class 218 (\class{lock~chamber})}
	\end{subfigure}
	\begin{subfigure}[b]{1.0\linewidth}
		\centering
		\includegraphics[width=0.85\linewidth]{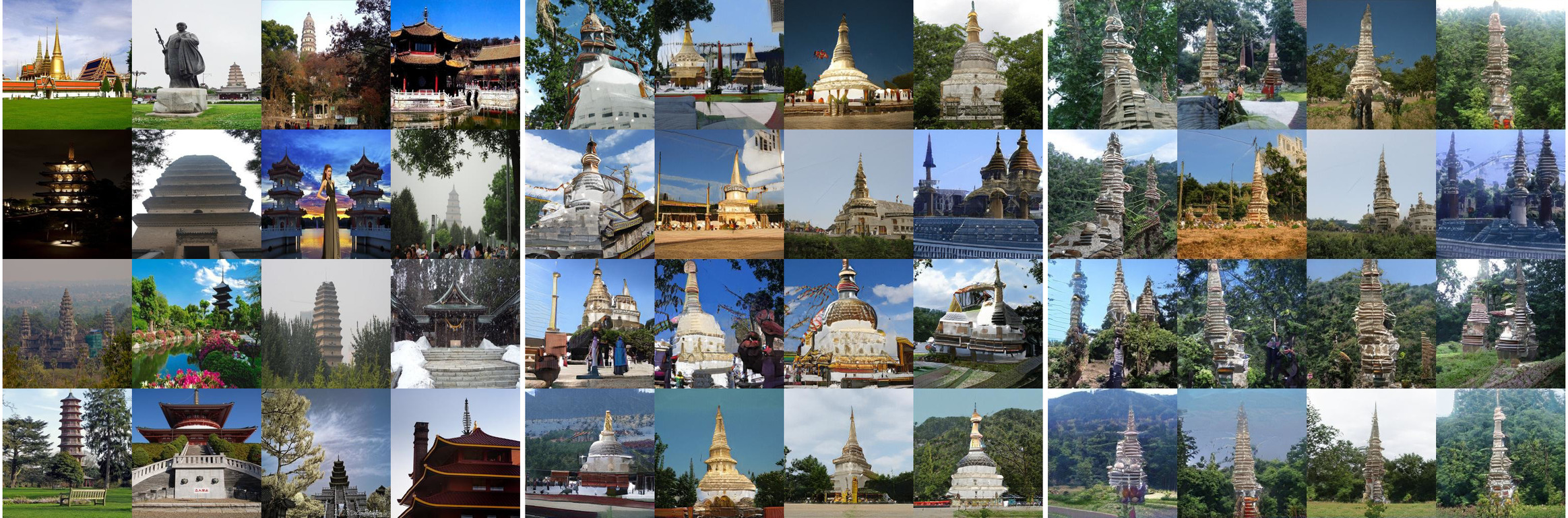}
		\vspace{-0.25cm}
		{
			\begin{flushleft}
				\hspace{2.0cm}  \class{pagoda}		
				\hspace{2.4cm}  \class{stupa}
				\hspace{2.3cm}  \class{pagoda}
			\end{flushleft}
		}
		\vspace{0.1cm}
		%\caption{Class 251 (\class{pagoda})}
	\end{subfigure}
	\begin{subfigure}[b]{1.0\linewidth}
	    \centering
		\includegraphics[width=0.85\linewidth]{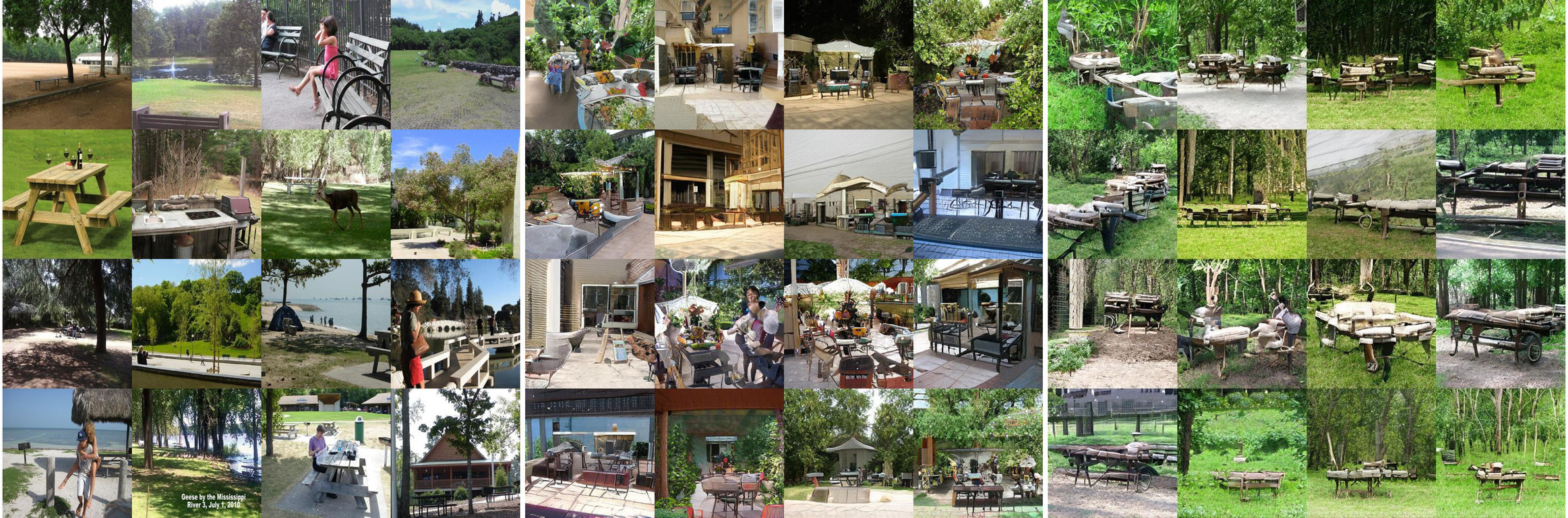}
		\vspace{-0.25cm}
		{
			\begin{flushleft}
				\hspace{1.8cm}  \class{picnic~area}	
				\hspace{2.2cm}  \class{patio}
				\hspace{2.1cm}  \class{picnic~area}
			\end{flushleft}
		}
		\vspace{0.1cm}
		%\caption{Class 265 (\class{picnic~area})}
	\end{subfigure}
	\begin{subfigure}[b]{1.0\linewidth}
		\centering
		\includegraphics[width=0.85\linewidth]{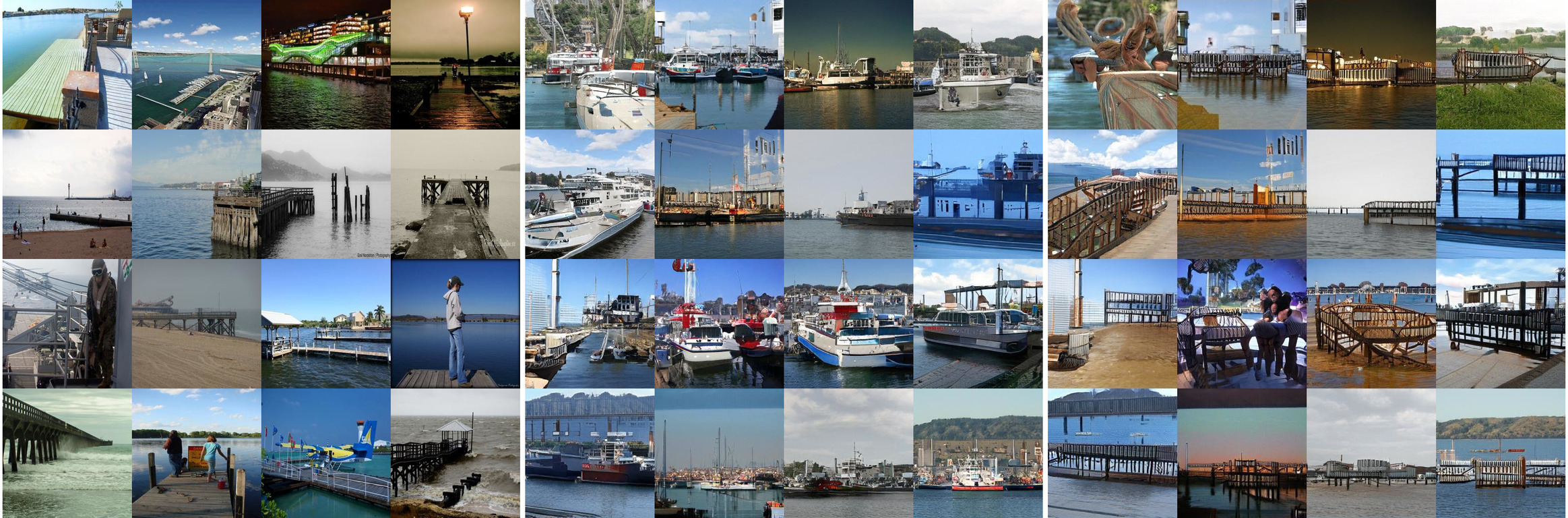}
		\vspace{-0.25cm}
		{
			\begin{flushleft}
				\hspace{2.3cm}  \class{pier}	
				\hspace{2.6cm}  \class{dock}
				\hspace{2.7cm}  \class{pier}
			\end{flushleft}
		}
		%\vspace{0.1cm}
		%\caption{Class 266 (\class{pier})}
	\end{subfigure}
	\caption{
		The same figure as Fig.~\ref{fig:mit_places_final_01} but for four different classes.
		In the bottom panel, while the BigGAN samples are \class{dock} images that contain mostly ships whereas AM samples show more bridges that resemble the real \class{pier} samples in Places365. See \url{https://drive.google.com/drive/folders/1L-1ULPfOf_5-98I7emYW86OPDu3Fjxnx?usp=sharing} for a high-resolution version of this figure.
	}
	\label{fig:mit_places_final_03}
\end{figure*}

\begin{figure*}[h!]
	\centering
	{	
		\begin{flushleft}
			\hspace{1.5cm} (A) Places365
			\hspace{0.45cm} (B) BigGAN on ImageNet		
			\hspace{0.3cm} (C) AM (ours)
		\end{flushleft}
	}
	\begin{subfigure}[b]{1.0\linewidth}
		\centering
		\includegraphics[width=0.85\linewidth]{images/places365_biggan_biggan-am_270.jpg}
		\vspace{-0.25cm}
		{
			\begin{flushleft}
				\hspace{2.1cm}  \class{plaza}	
				\hspace{2.0cm}  \class{parking~meter}
				\hspace{1.9cm}  \class{plaza}
			\end{flushleft}
		}
		\vspace{0.1cm}
		%\caption{Class 270 (\class{plaza})}
	\end{subfigure}
	\begin{subfigure}[b]{1.0\linewidth}
		\centering
		\includegraphics[width=0.85\linewidth]{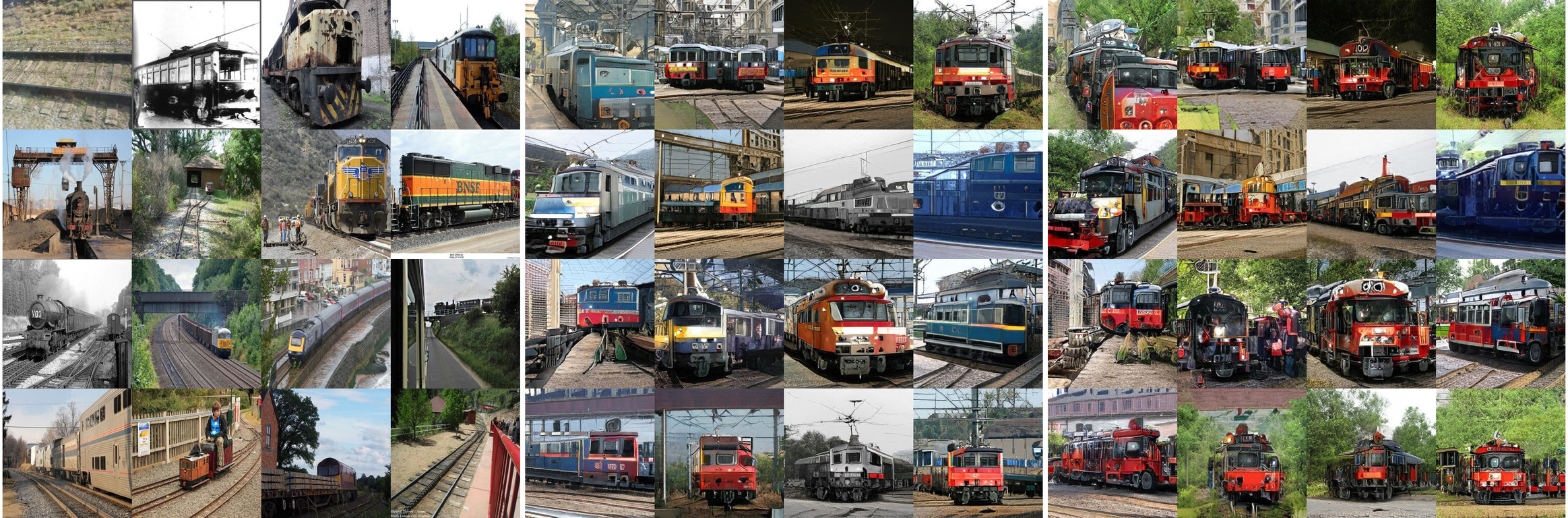}
		\vspace{-0.25cm}
		{
			\begin{flushleft}
				\hspace{1.6cm}  \class{railroad~track}	
				\hspace{1.1cm}  \class{electric~locomotive}
				\hspace{1.0cm}  \class{railroad~track}
			\end{flushleft}
		}
		\vspace{0.1cm}
		%\caption{Class 278 (\class{railroad~track})}
	\end{subfigure}
	\begin{subfigure}[b]{1.0\linewidth}
		\centering
		\includegraphics[width=0.85\linewidth]{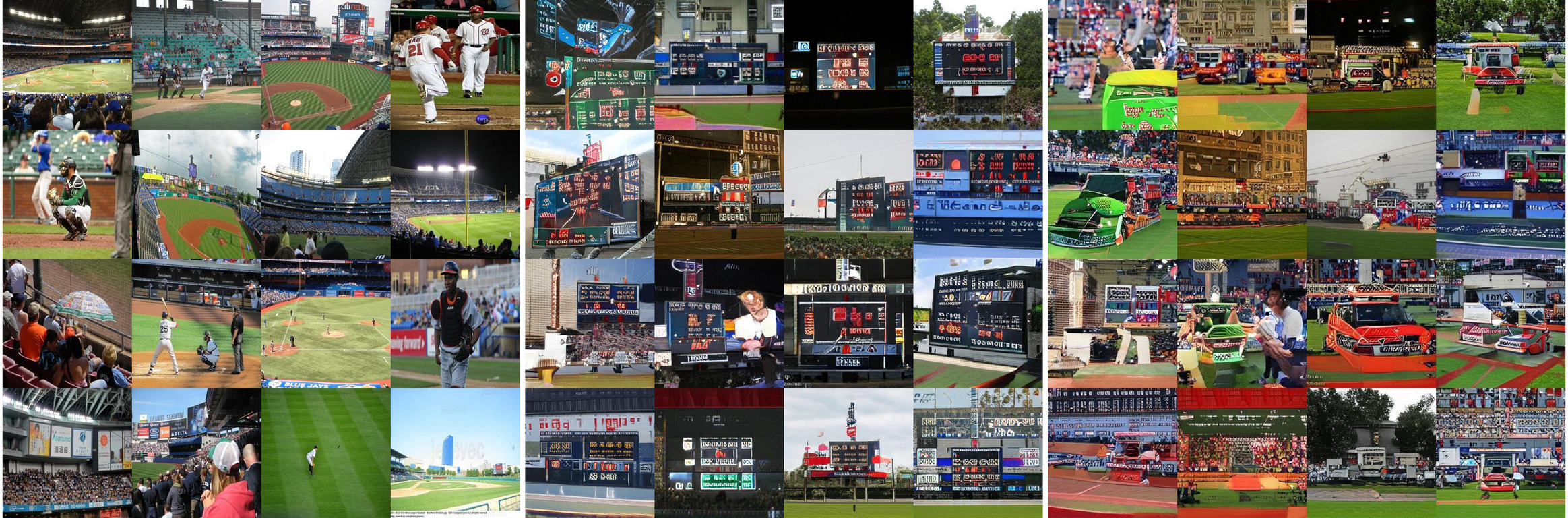}
		\vspace{-0.25cm}
		{
			\begin{flushleft}
				\hspace{1.4cm}  \class{baseball~stadium}	
				\hspace{1.4cm}  \class{scoreboard}
				\hspace{1.5cm}  \class{baseball~stadium}
			\end{flushleft}
		}
		\vspace{0.1cm}
		%\caption{Class 312 (\class{baseball~stadium})}
	\end{subfigure}
	\begin{subfigure}[b]{1.0\linewidth}
		\centering
		\includegraphics[width=0.85\linewidth]{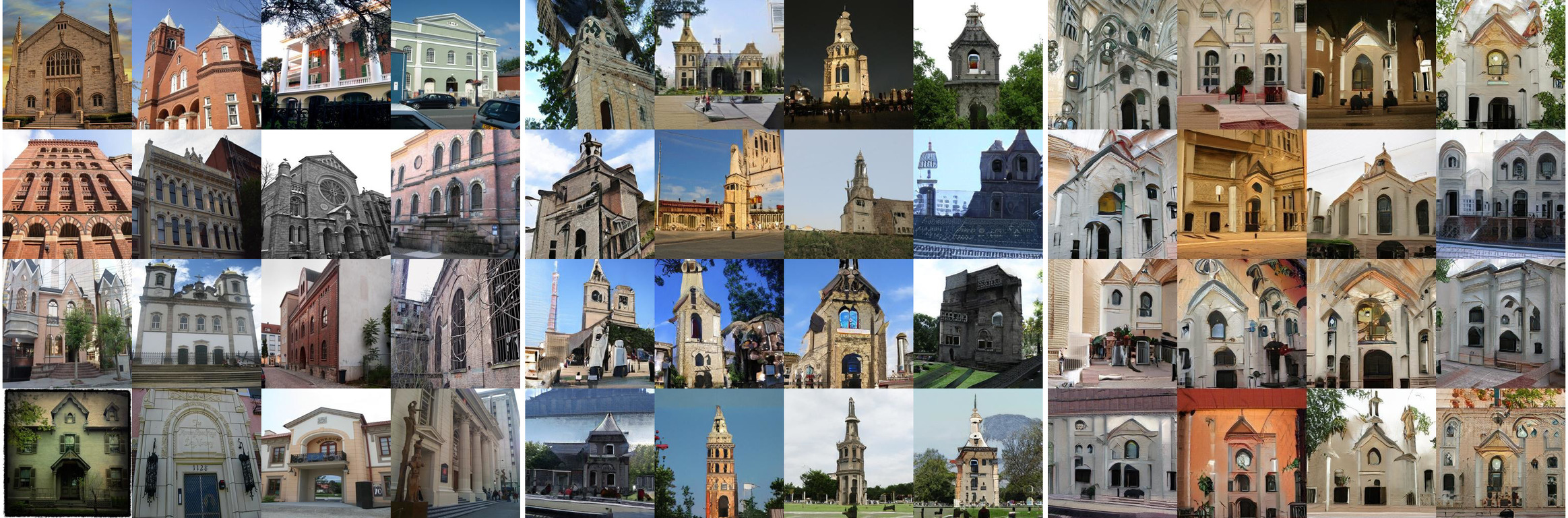}
		\vspace{-0.25cm}
		{
			\begin{flushleft}
				\hspace{1.3cm}  \class{synagogue~outdoor}	
				\hspace{1.4cm}  \class{bell~cote}
				\hspace{1.3cm}  \class{synagogue~outdoor}
			\end{flushleft}
		}
		%\vspace{0.1cm}
		%\caption{Class 327 (\class{synagogue~outdoor})}
	\end{subfigure}
	\caption{
		The same figure as Fig.~\ref{fig:mit_places_final_01} but for four different classes.
		For the \class{baseball~stadium}, the top-1 ImageNet class is \class{scoreboard} (B), an object commonly found in stadiums.
		However, the AM samples are more similar to the images from Places365, which often do not contain scoreboards (A~vs.~C). See \url{https://drive.google.com/drive/folders/1L-1ULPfOf_5-98I7emYW86OPDu3Fjxnx?usp=sharing} for a high-resolution version of this figure.}
	\label{fig:mit_places_final_04}
\end{figure*}

\begin{figure*}[h!]
	\centering

	\begin{subfigure}[b]{0.3\linewidth}
		\centering
		\includegraphics[width=1.0\linewidth]{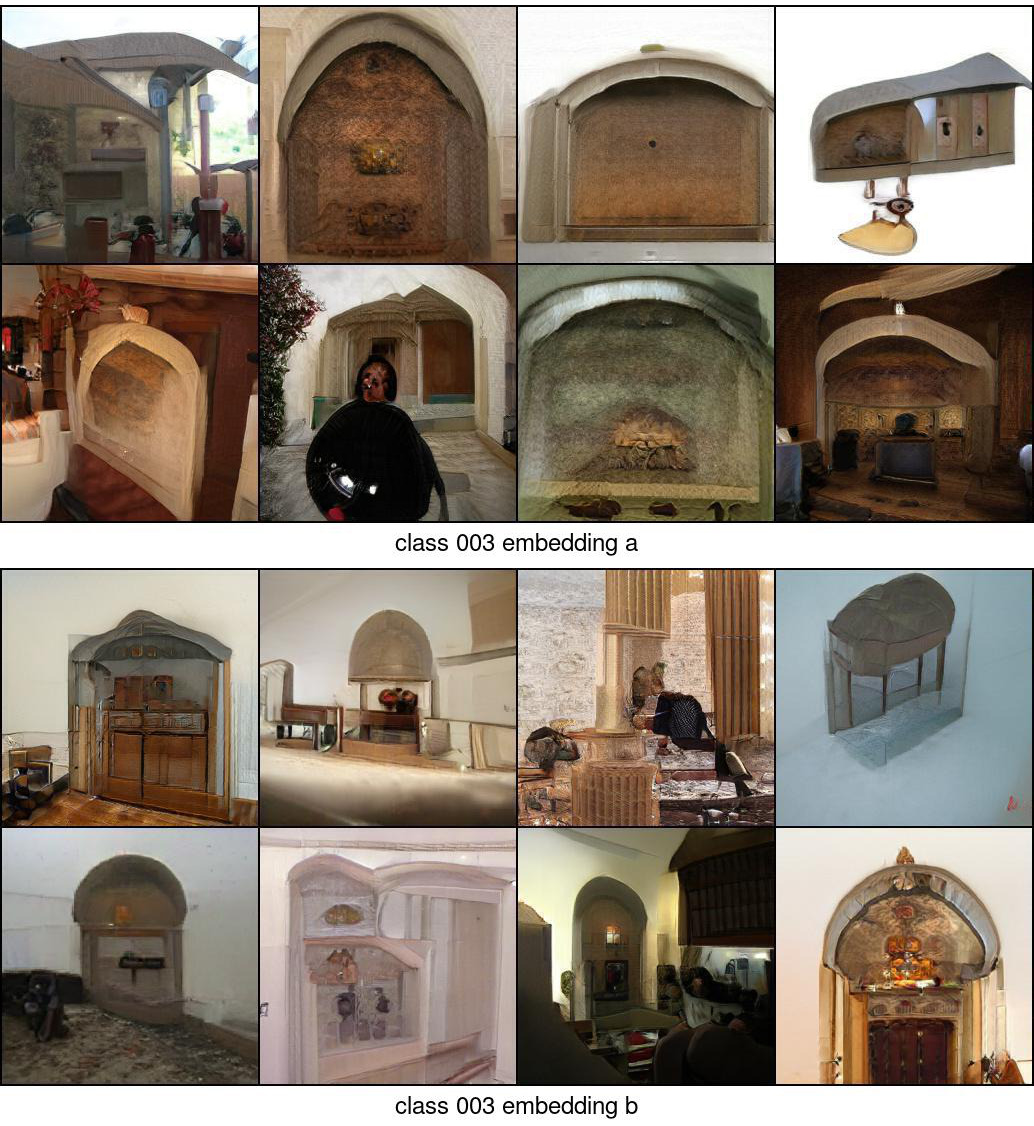}
		\caption{\class{alcove}}
	\end{subfigure}
	\begin{subfigure}[b]{0.3 \linewidth}
		\centering
		\includegraphics[width=1.0\linewidth]{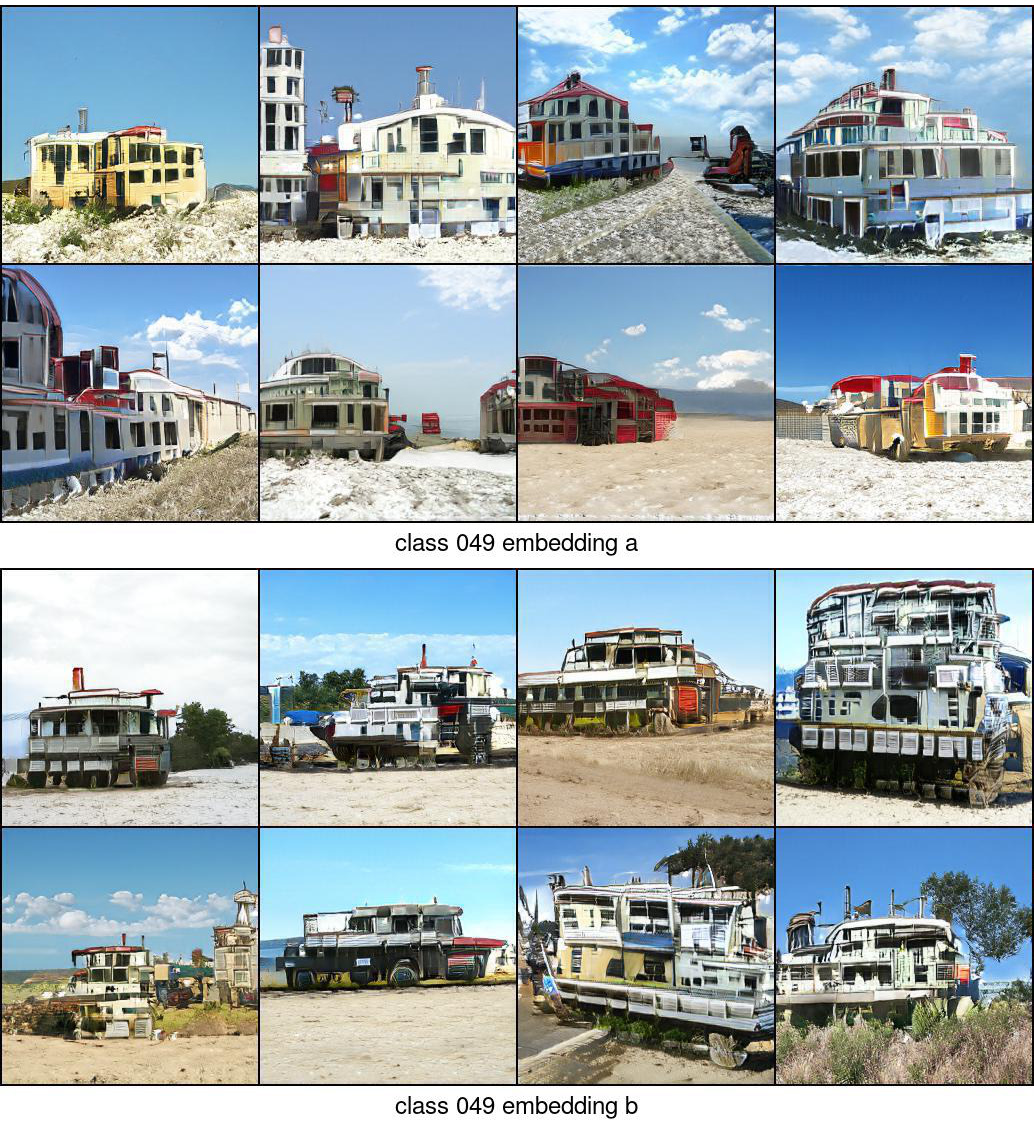}
        \caption{ \class{beach~house}}
	\end{subfigure}
	\begin{subfigure}[b]{0.3\linewidth}
		\centering
		\includegraphics[width=1.0\linewidth]{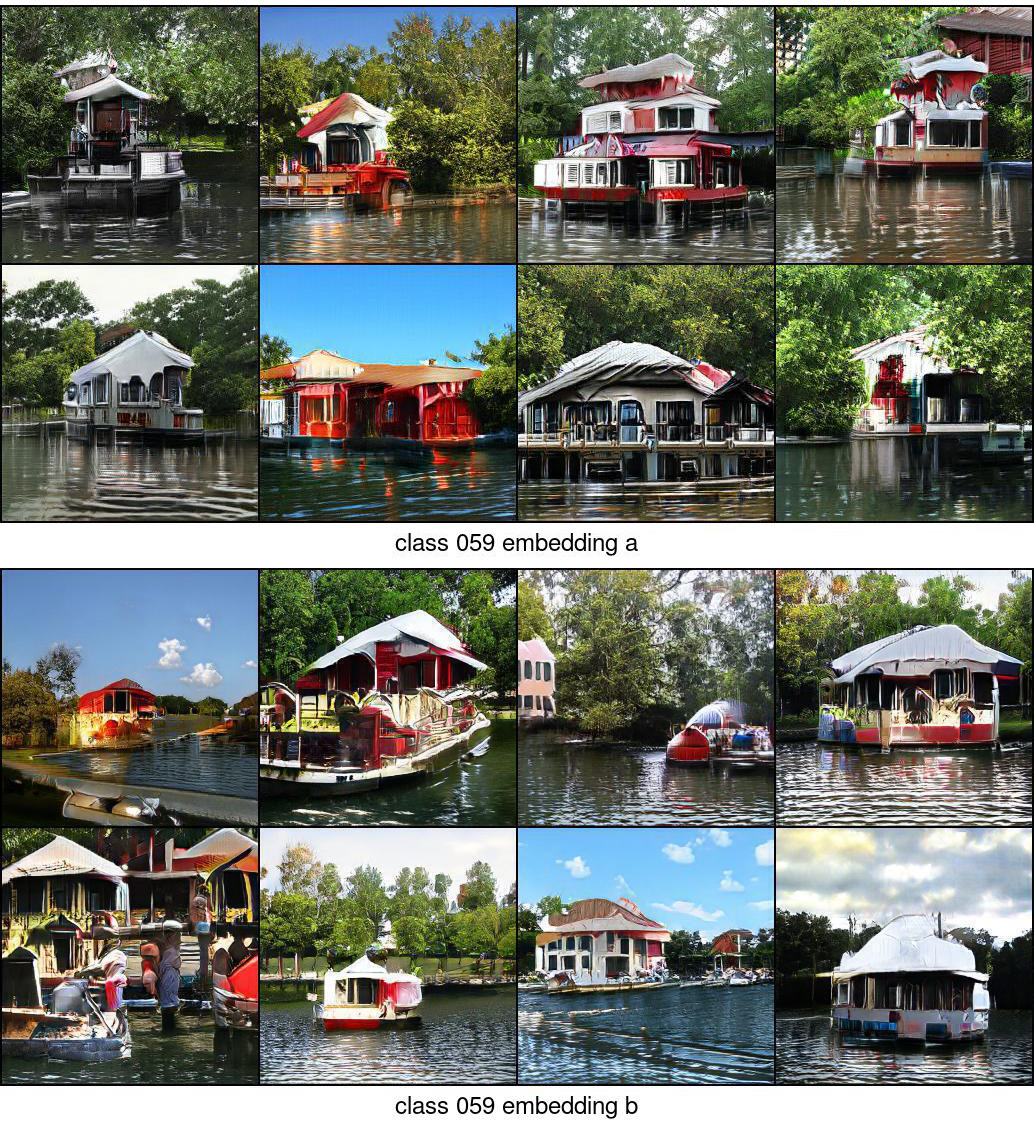}
        \caption{\class{boathouse}}
	\end{subfigure}
	
% 	\begin{subfigure}[b]{0.3\linewidth}
% 		\centering
% 		\includegraphics[width=1.0\linewidth]{images/2embeddings_for_mit_places_189.jpg}
% 		\caption{\class{ice~skating~rink~outdoor}}
% 	\end{subfigure}
	
	\begin{subfigure}[b]{0.3\linewidth}
		\centering
		\includegraphics[width=1.0\linewidth]{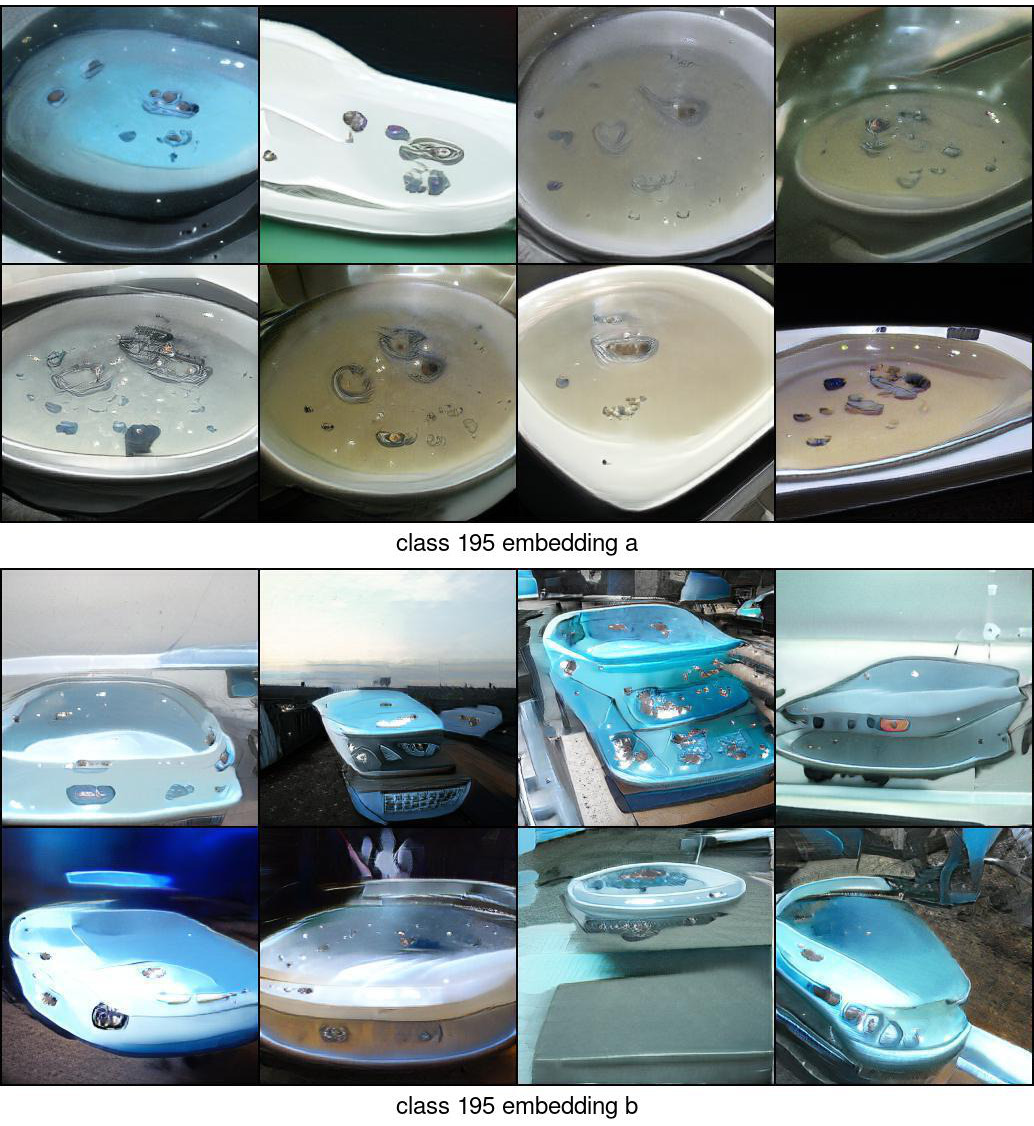}
		\caption{\class{jacuzzi~indoor}}
	\end{subfigure}
	\begin{subfigure}[b]{0.3\linewidth}
		\centering
		\includegraphics[width=1.0\linewidth]{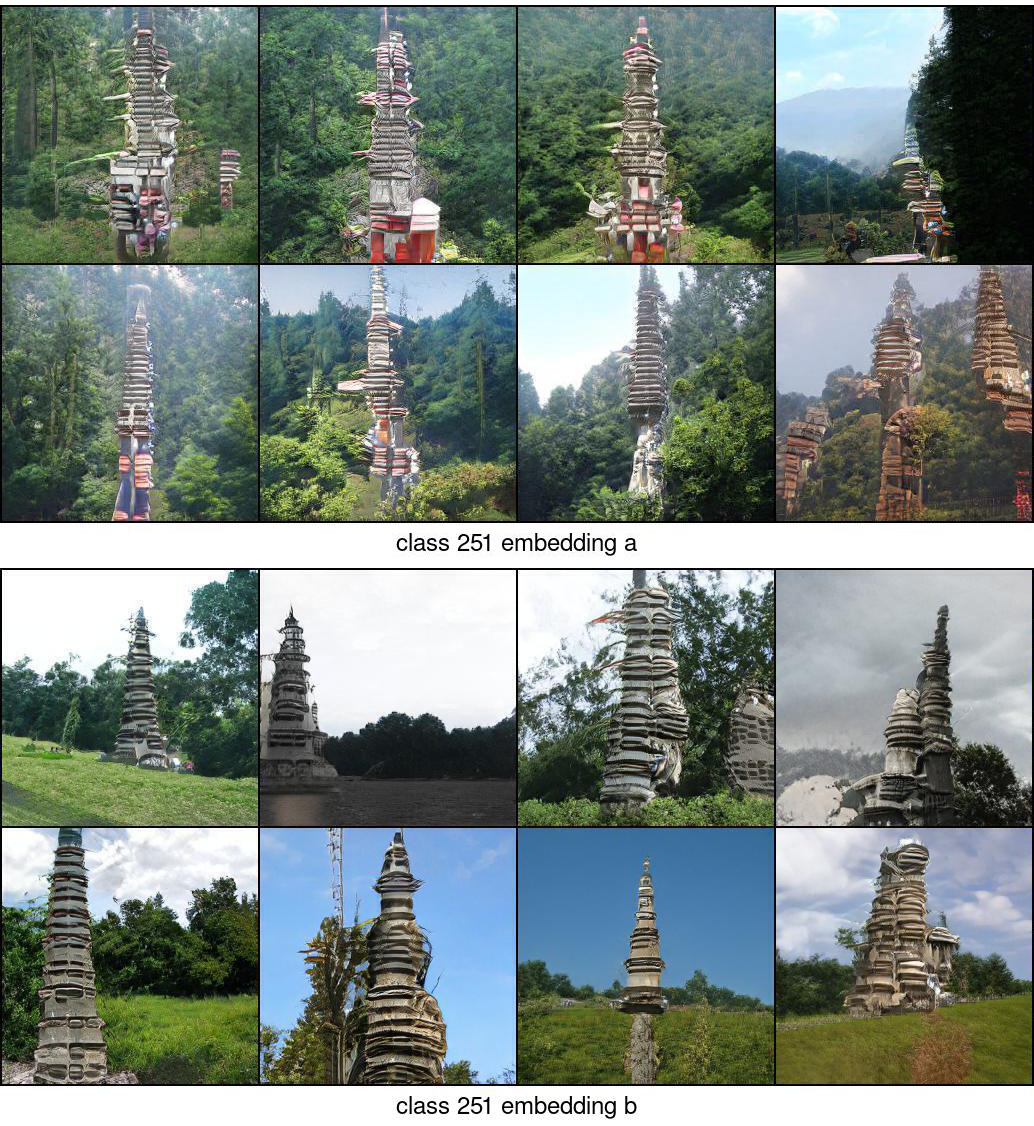}
		\caption{\class{pagoda}}
	\end{subfigure}
	\begin{subfigure}[b]{0.3\linewidth}
		\centering
		\includegraphics[width=1.0\linewidth]{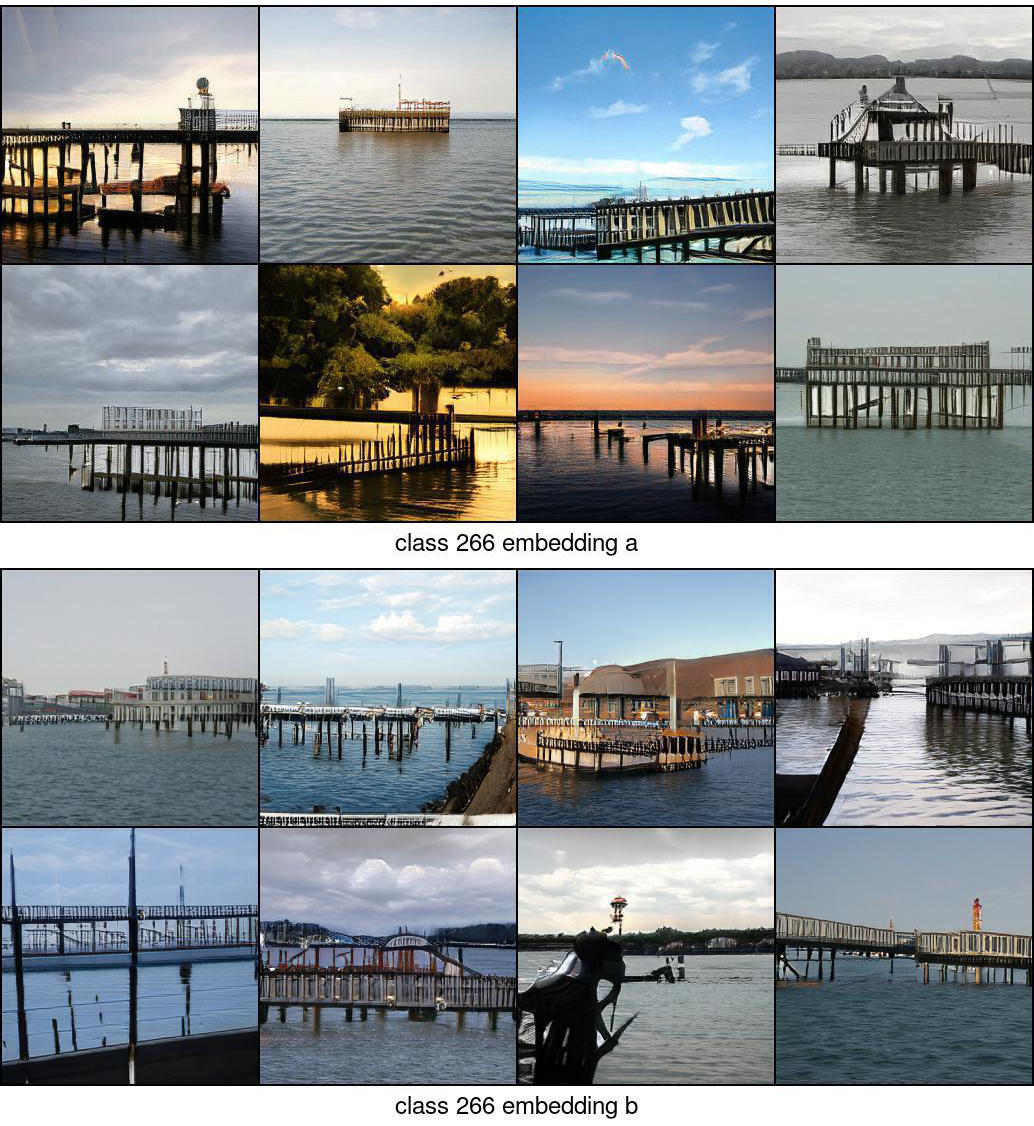}
		\caption{\class{pier}}
	\end{subfigure}
	\begin{subfigure}[b]{0.3\linewidth}
		\centering
		\includegraphics[width=1.0\linewidth]{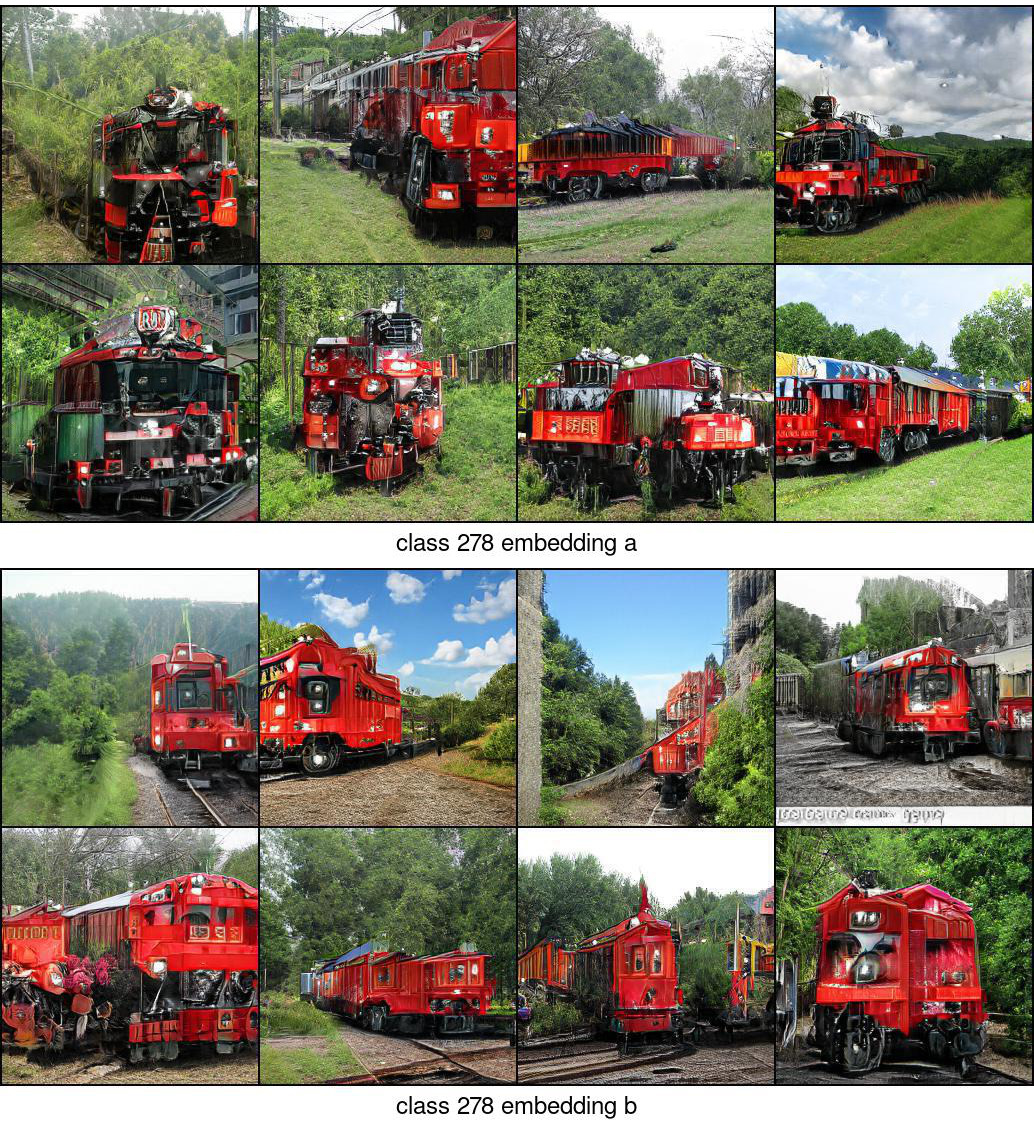}
		\caption{\class{railroad~track}}
	\end{subfigure}
	\begin{subfigure}[b]{0.3\linewidth}
		\centering
		\includegraphics[width=1.0\linewidth]{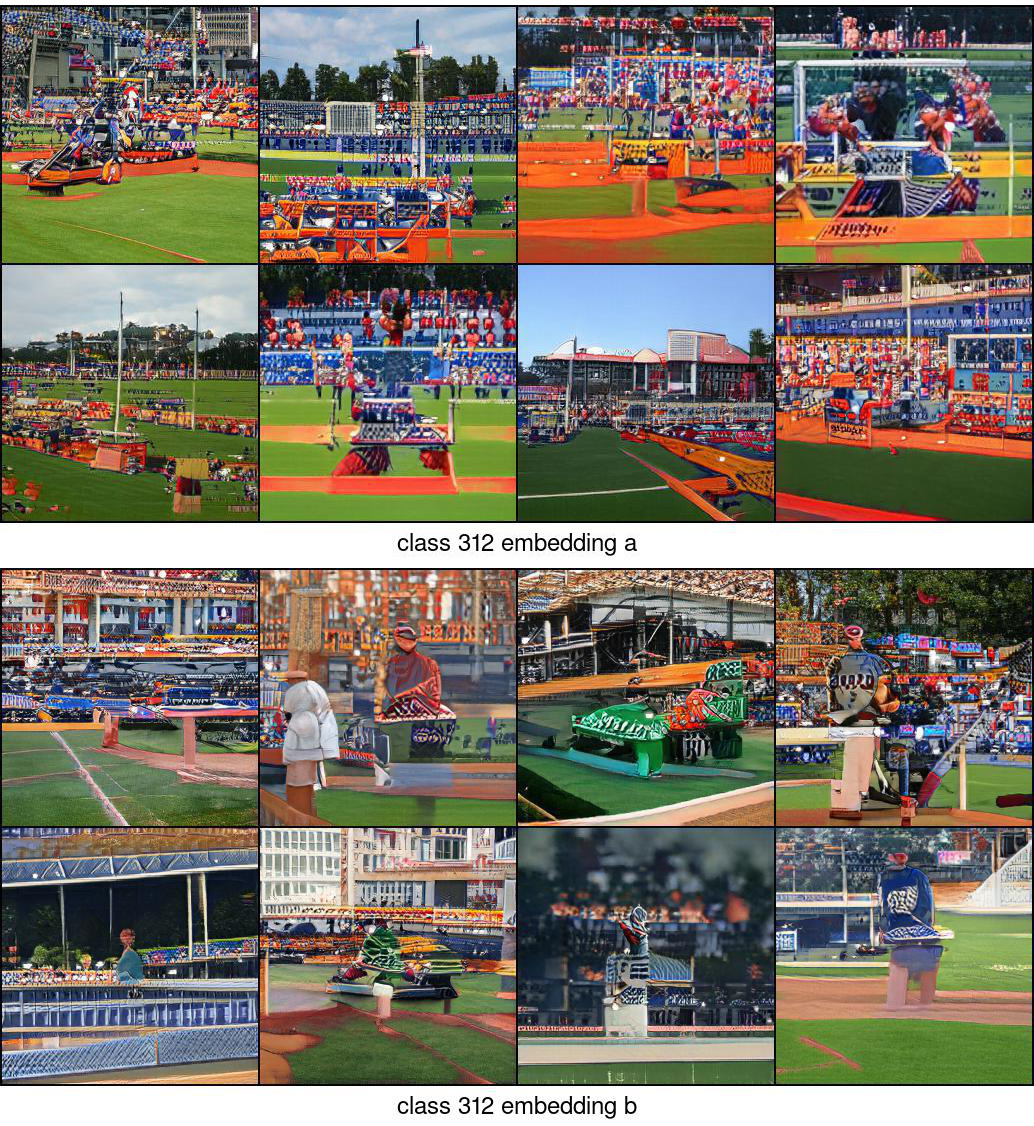}
		\caption{\class{baseball~stadium}}
	\end{subfigure}
	\begin{subfigure}[b]{0.3\linewidth}
		\centering
		\includegraphics[width=1.0\linewidth]{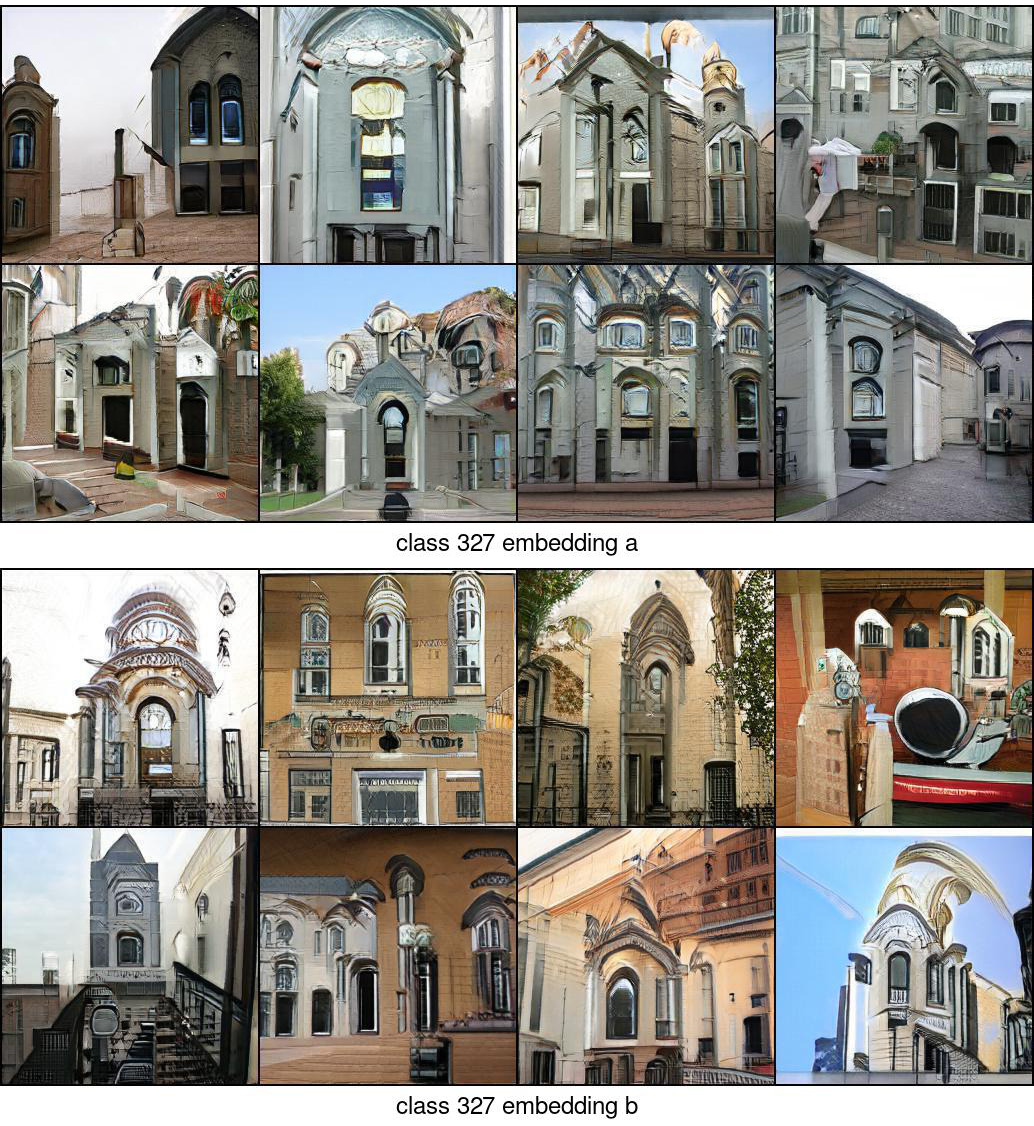}
		\caption{\class{synagogue~outdoor}}
	\end{subfigure}
	
	\caption{ \qi{For each class, we find 2 class embeddings by using AM and generate a set of images by using the same $\vz$. The samples from each class have different style corresponding to different class embeddings.}
}
	\label{fig:mit_places_2_embeddings_01}
\end{figure*}

\end{document}